\documentclass{article}

\usepackage[nonatbib,final]{neurips_2021}

\usepackage[utf8]{inputenc} %
\usepackage[T1]{fontenc}    %
\usepackage{hyperref}       %
\usepackage{url}            %
\usepackage{booktabs}       %
\usepackage{amsfonts}       %
\usepackage{nicefrac}       %
\usepackage{microtype}      %
\usepackage{xcolor}         %
\definecolor{codegreen}{rgb}{0,0.5,0}
\definecolor{codeblue}{rgb}{0.25,0.5,0.5}
\definecolor{codegray}{rgb}{0.6,0.6,0.6}
\definecolor{comments}{RGB}{0,0,113}
\definecolor{red}{RGB}{160,0,0}
\definecolor{green}{RGB}{0,100,0}
\usepackage{listings}
\lstdefinestyle{codestyle}{
  backgroundcolor=\color{white},
  basicstyle=\fontsize{8.5pt}{9.5pt}\fontfamily{lmtt}\selectfont,
  columns=fullflexible,
  breaklines=true,
  captionpos=b,
  commentstyle=\fontsize{8pt}{9pt}\color{codegreen},
  keywordstyle=\fontsize{8pt}{9pt}\color{comments},
  stringstyle=\fontsize{8pt}{9pt}\color{red},
  showstringspaces=false,
  frame=tb,
  otherkeywords = {self},
  escapeinside=@@,
}
\usepackage{enumitem}
\usepackage{amsmath}
\usepackage{lipsum}
\usepackage[numbers, compress, sort]{natbib}
\usepackage{caption}
\usepackage{subcaption}
\usepackage{graphicx}
\usepackage{wrapfig}
\usepackage{pifont}
\usepackage{algorithm}
\usepackage{algorithmic}

\newcommand\blfootnote[1]{%
  \begingroup
  \renewcommand\thefootnote{}\footnote{#1}%
  \addtocounter{footnote}{-1}%
  \endgroup
}

\title{Dynamic Inference with Neural Interpreters}

\author{%
  Muhammad Waleed Gondal$^{*,1}$
  \And
  Nasim Rahaman$^{*,1,2,3,\ddag}$
  \And
  Shruti Joshi$^{1}$
  \And
  Peter Gehler$^{3}$
  \And
  Yoshua Bengio$^{\dagger,2}$
  \And
  Francesco Locatello$^{\dagger,3}$
  \And
  Bernhard Sch\"olkopf$^{\dagger,1}$
}

\begin{document}

\maketitle

\begin{abstract}

Modern neural network architectures can leverage large amounts of data to generalize well within the training distribution. However, they are less capable of systematic generalization to data drawn from unseen but related distributions, a feat that is hypothesized to require compositional reasoning and reuse of knowledge. In this work, we present Neural Interpreters, an architecture that factorizes inference in a self-attention network as a system of modules, which we call \emph{functions}. Inputs to the model are routed through a sequence of functions in a way that is end-to-end learned. 
The proposed architecture can flexibly compose computation along width and depth, and lends itself well to capacity extension after training. 
To demonstrate the versatility of Neural Interpreters, we evaluate it in two distinct settings: image classification and visual abstract reasoning on Raven Progressive Matrices. In the former, we show that Neural Interpreters perform on par with the vision transformer using fewer parameters, while being transferrable to a new task in a sample efficient manner. In the latter, we find that Neural Interpreters are competitive with respect to the state-of-the-art in terms of systematic generalization.\blfootnote{$^{*,\dagger}$Equal contribution, ordered alphabetically by last name. $^1$Max-Planck-Institute for Intelligent Systems, T\"ubingen. $^2$Mila, Qu\'ebec. $^3$Amazon Web Services. $^{\ddag}$Work partially done during an internship at Amazon Web Services.}

\end{abstract}

\section{Introduction}

Rule-based programming is the basis of computer science, and builds the foundation of symbolic-AI based expert systems that attempt to emulate human decision-making to solve real-world problems. The process of inference entails channeling information through a chain of computational units (e.g., logical primitives) and culminates in a conclusion that answers a given query. 
Such systems have the advantage that they permit efficient reuse of computational units and enable iterative reasoning over multiple cycles of computation. As a simple example, consider the relation \texttt{parent\_child(A, B)}, which can be used to construct a new relation \texttt{sibling(U, V) = parent\_child(A, U) AND parent\_child(A, V)}, which in turn can be used to construct yet another relation \texttt{cousin(U, V) = parent\_child(X, U) AND parent\_child(Y, V) AND sibling(X, Y)}, and so on. However, such systems are known to suffer from the knowledge acquisition problem, which is the inability to leverage unstructured data to derive new computational units and improve existing ones \citep{kendal2007introduction}. 

\looseness=-1
In stark contrast to the symbolic paradigm, modern machine learning models excel at absorbing large amounts of unstructured data and yield strong performance in many challenging domains, ranging from large-scale image classification to language modelling. However, they are relatively rigid in how they share and reuse computation in order to process information: convolutional neural networks, for instance, process the content of an image at every location to arrive at the class label of a given image. 
In doing so, they only reuse computational units (here, convolutional filters) \emph{laterally} at a constant depth i.e., amongst information being processed simultaneously. In the same spirit, recurrent neural networks only reuse computational units (here, RNN cells) \emph{vertically}, i.e., in depth. Such rigidity in how computation is reused is believed to be one of the reasons current deep neural networks are less capable of systematically generalizing to problems not encountered during training \citep{bahdanau2018systematic, lake2018generalization,loula2018rearranging}.

\looseness=-1
In the present work, we draw inspiration from typed programming languages to develop a self-attention based neural architecture that relaxes this rigidity in computation reuse. The resulting class of models, which we call Neural Interpreters, \emph{learns} to flexibly use and compose computational units directly from data without additional supervision. 
Neural Interpreters factorize a self-attention network \citep{vaswani2017attention} as a system of computational units that we call \emph{functions}.
The input to the network is set-valued and processed in the network by a dynamically inferred sequence of functions -- potentially by the same function multiple times, enabling vertical sharing of computational units. This aligns with earlier ideas on
independent mechanisms~\citep{peters2017elements,ParKilRojSch18,goyal2019recurrent,goyal2020object,goyal2020inductive, Scholkopfetal21, rahaman2020s2rms,goyal2021neural}: a set of mechanisms can be combined and reused in many different ways depending on context (the current input or task), thus factorizing knowledge in independent pieces which can lead to better systematic generalization.

\looseness=-1
Neural Interpreters have two key benefits over vanilla self-attention networks. 
First, the modular inductive biases facilitate %
generalization beyond the training distribution and adaptation to new tasks in a sample-efficient manner. 
Second, but consistent with the notion of factorizing knowledge into approximately independent and composable pieces, 
the proposed parameterization is (by construction) not only agnostic to the cardinality of the input set, but also to the number of functions. The latter implies that given a new task, additional functions can be non-disruptively added and fine-tuned. In other words, knowledge acquired from the prior training tasks can be effectively repurposed for new tasks.

\looseness=-1
\textbf{Our primary contributions are as follows.} \textbf{(a)} We introduce the Neural Interpreter, an attention-based architecture that can be applied on arbitrary set-valued inputs or representations. \textbf{(b)} We quantitatively evaluate the proposed architecture in two distinct problem settings: multi-task image classification and abstract reasoning. In the former setting, we show that Neural Interpreters are capable of sample-efficient adaptation to new tasks and can exploit additional capacity added after pre-training. In the latter setting, we demonstrate that Neural Interpreters are capable of out-of-distribution generalization. In particular, we find that Neural Interpreters outperform the Vision Transformer baseline \citep{cordonnier2019relationship, dosovitskiy2020image}, which in turn is competitive in-distribution with the prior state-of-the-art. In both settings, we find that Neural Interpreters develop the ability to gracefully trade-off performance with compute at inference time \citep{zilberstein1996using}. In addition, we include results on a toy problem, where we explicitly probe the ability of Neural Interpreters to learn recomposable computational primitives. 
\textbf{(c)} We ablate over the architectural parameters of Neural Interpreters and qualitatively visualize what the model learns. We find patterns in how the input is routed through the network and that a wide range of hyperparameters yield promising results.

\section{Neural Interpreters} \label{sec:ni}

\begin{figure}[htp]
\vspace{-11pt}
\centering
\begin{subfigure}{\textwidth}
  \centering
  \includegraphics[width=1\linewidth]{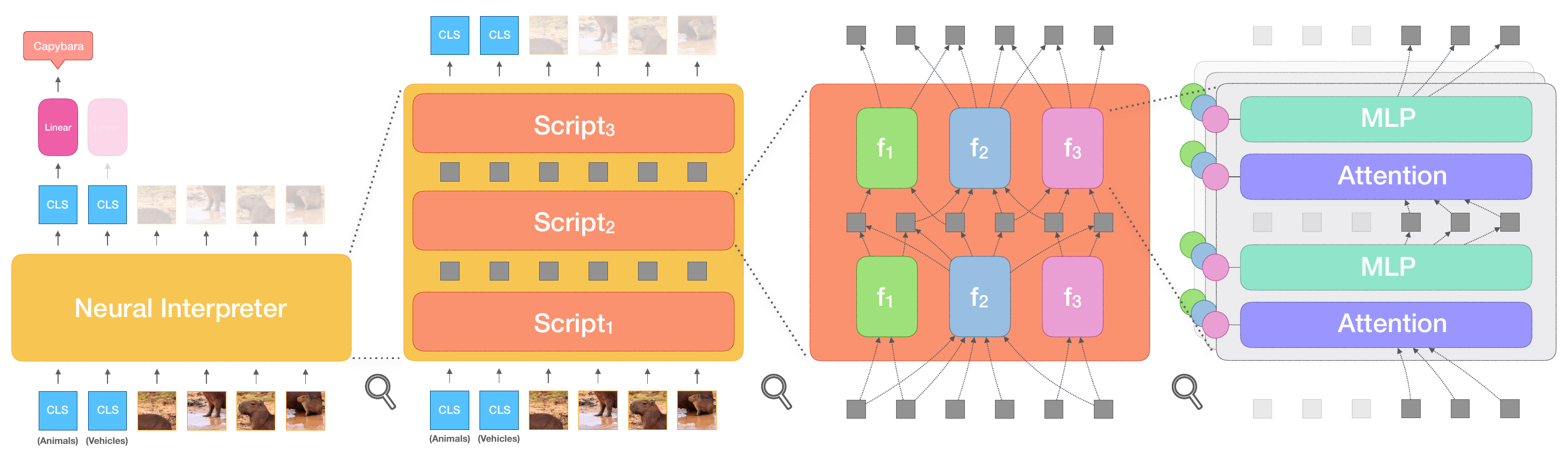}
\end{subfigure}
\caption{\textbf{Leftmost:} Overview of the architecture, shown here with image patches as inputs and two CLS tokens with corresponding classification heads.
\textbf{Center Left:} The Neural Interpreter, shown here as a stack of three \textit{scripts}. \textbf{Center Right:} A script, shown here as a collection of three \textit{functions} applied over two \textit{function iterations}. \textbf{Rightmost:} A stack of two \textit{Lines of Code} (LOCs), spread over three parallel computational streams (one per function) and conditioned by the respective function codes (colored circles). 
Residual connections are present but not shown in this figure.}
\label{fig:ni_overview}
\vspace{-8pt}
\end{figure}

In this section, we describe in detail the components of a Neural Interpreter; see Figure~\ref{fig:ni_overview} for an overview. This architecture can be used as a drop-in replacement for a self-attention network, e.g., Transformers \citep{vaswani2017attention}. In line with prior work \citep{cordonnier2019relationship, dosovitskiy2020image, touvron2021training}, we focus on applications to visual data.

\looseness=-1
\textbf{Input and Output.} The input to the Neural Interpreter is any set with vector-valued elements $\{\mathbf{x}_i\}_i, \mathbf{x}_i\in\mathbb{R}^{d_{\text{in}}}$ and the output is another set of vectors $\{\mathbf{y}_j\}_j, \mathbf{y}_j \in \mathbb{R}^{d_{\text{out}}}$ with the same cardinality as the input set. We assume in this work that the input set contains vector embeddings of image patches \citep{cordonnier2019relationship} or of entire images \citep{barrett2018measuring}. The input set additionally includes one or more learned vectors, called CLS Tokens \citep{devlin2019bert}, for which the corresponding outputs interface with their respective classifiers \citep{logeswaran2020few}.

\looseness=-1
\textbf{Scripts.} 
A Neural Interpreter can be expressed as a stack of $n_s$ \textit{scripts}, where a script maps one set of vectors to another with the same number of elements:
\begin{equation} \label{eq:script}
    \{\mathbf{y}_j\}_{j} = \text{NeuralInterpreter}(\{\mathbf{x}_i\}_{i}) = \left[\text{Script}_{n_s} \circ ... \circ (n_s \text{ times}) \circ ... \circ \text{Script}_{1}\right](\{\mathbf{x}_i\}_{i})
\end{equation}
A script has four components: a type inference module, a type matching mechanism, a set of functions and an interpreter, as explained below. The parameters of these components are not shared between scripts. 
\textit{\underline{Role}:} Scripts function as independent building blocks that can be dropped in any set-to-set architecture, and Neural Interpreters with a single script can perform well in practice. 

\looseness=-1
\textbf{Functions.} 
\textit{Functions} make the computational units in Neural Interpreters; they can be represented as a tuple of vector valued parameters, i.e., $f_u = (\mathbf{s}_u, \mathbf{c}_u)$ where $u$ indexes functions. Here, $\mathbf{s}_u$ is referred to as the signature of the function $f_u$ (with a meaning similar to that in programming languages), and it is a normalized vector of $d_{\text{type}}$ dimensions. The signature vector specifies to the type matching mechanism (see below) what inputs are to be routed to $f_u$. We refer to $\mathbf{c}_u$ as the code vector of $f_u$, as it instructs an interpreter module (shared between functions, see below) how to process the inputs to $f_u$. 
\textit{\underline{Role}:} Functions are vector-valued \emph{instructions} to other components in the script. They implement the computational units that can be reused in the computational graph.

\begin{wrapfigure}{r}{0.47\textwidth}
\vspace{-15pt}
\centering
\includegraphics[width=0.47\textwidth]{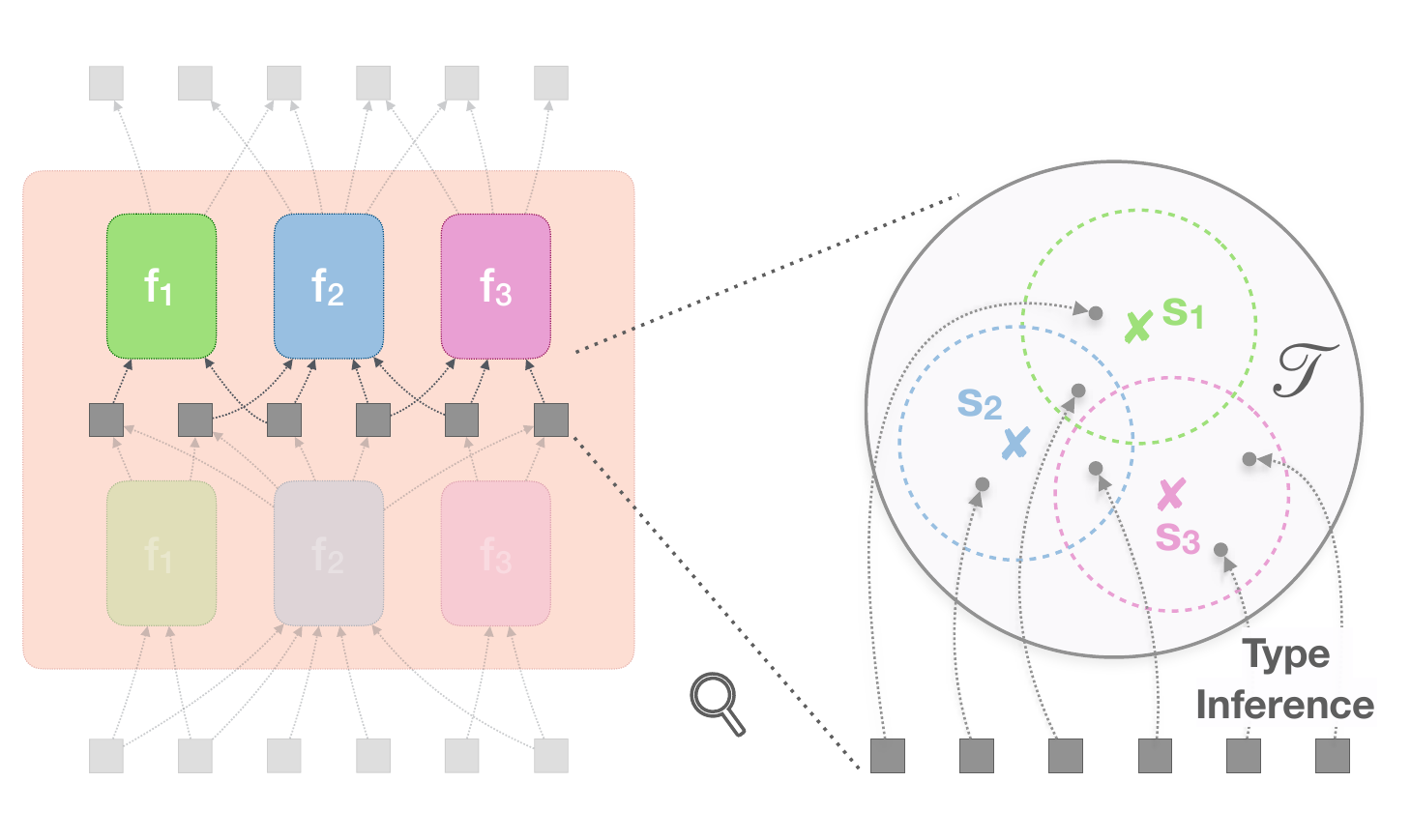}
\caption{\small Illustration of the type matching mechanism. Functions only access set elements whose types lie in the vicinity of their signatures. \label{fig:ni_typeinference}}
\vspace{-8pt}
\end{wrapfigure}

\looseness=-1
\textbf{Type Matching and Inference.} The type matching mechanism (Figure~\ref{fig:ni_typeinference})
enables the learned routing of set elements through functions, and it proceeds in three steps. First, given a set element $\mathbf{x}_i$, it is processed by an MLP module that outputs its \emph{type vector} $\mathbf{t}_i$. This module is called the \emph{type inference} module, and the resulting type vector is an element of the same topological space as function signatures, i.e., a $d_{\text{type}}$-dimensional hypersphere $\mathcal{T}$.
Next, given a function $f_u$ and its signature vector $\mathbf{s}_u \in \mathcal{T}$, the compatibility $C_{ui}$ between the function $f_u$ and a set element $\mathbf{x}_i$ is determined by the cosine similarity between $\mathbf{s}_u$ and $\mathbf{t}_i$ in $\mathcal{T}$ (larger $\implies$ more compatible).
Finally, if this compatibility is larger than a threshold ($\tau$), $f_u$ is permitted access to $\mathbf{x}_i$. Formally, let $\{\mathbf{x}_i\}_{i}$ be a set of intermediate representation vectors (indexed by $i$). With a learnable parameter $\sigma$ and a hyper-parameter $\tau$, we have: 
\begin{align} \label{eq:type_inference}
    \mathbf{t}_i = \text{TypeInference}(\mathbf{x}_i) \in \mathcal{T}; \quad & \quad d_{\mathcal{T}}(\mathbf{s}_u, \mathbf{t}_i) = (1 - \mathbf{s}_u \cdot \mathbf{t}_i)\\
    C_{ui} = \frac{\tilde{C}_{ui}}{\epsilon + \sum_u \tilde{C}_{ui}} \quad \text{where}\quad \tilde{C}_{ui} = \exp &\left[ -\frac{d_{\mathcal{T}}(\mathbf{s}_u, \mathbf{t}_i)}{\sigma}\right] \text{~~if~~} d_{\mathcal{T}}(\mathbf{s}_u, \mathbf{t}_i) > \tau \text{,~~else~~} 0.
\end{align}
$\tau$ is called the truncation parameter of the kernel and $\epsilon$ is a small scalar for numerical stability. The compatibility matrix $C_{ui} \in [0, 1]$ will serve as a modulation mask \citep{rahaman2020s2rms} for the self-attention mechanism in the interpreter (see below).
\textit{\underline{Role}:} The type matching mechanism is responsible for routing information through functions. The truncation parameter controls the amount of sparsity in routing. 

\looseness=-1
\textbf{ModLin Layers and ModMLP.} The components described below make use of linear layers conditioned by some code $\mathbf{c}$. Consider a linear layer with weight matrix $\mathbf{W} \in \mathbb{R}^{d_{\text{out}}} \times \mathbb{R}^{d_{\text{in}}}$ and a bias vector $\mathbf{b} \in \mathbb{R}^{d_{\text{out}}}$. Let $\mathbf{x} \in \mathbb{R}^{d_{\text{in}}}$ denote the input vector to the layer, and $\mathbf{c} \in \mathbb{R}^{d_{\text{cond}}}$ a condition vector, and $\mathbf{W}_c \in \mathbb{R}^{d_{\text{in}}} \times \mathbb{R}^{d_{\text{cond.}}}$ a learnable matrix. The output $\mathbf{y} \in \mathbb{R}^{d_{\text{out}}}$ is given as: 
\begin{equation} \label{eq:modlin}
    \mathbf{y} = \text{ModLin}(\mathbf{x}; \mathbf{c}) = \mathbf{W} (\mathbf{x} \otimes \text{LayerNorm}(W_c \mathbf{c})) + \mathbf{b}
\end{equation}
where $\otimes$ denotes element-wise product and we call the resulting layer a modulated linear layer, or a $\text{ModLin}$ layer \citep{anokhin2020image}. Further, one may stack (say) $L$ such ModLin layers (sharing the same condition or code vector $\mathbf{c}$) interspersed with an activation function (we use GELUs \citep{hendrycks2020gaussian}) to obtain a ModMLP: 
\begin{equation}
\mathbf{y} = \text{ModMLP}(\mathbf{x}; \mathbf{c}) = (\text{ModLin}_L(\,\cdot\,; \mathbf{c}) \circ \text{Activation} \circ ... \circ \text{ModLin}_1(\,\cdot\,; \mathbf{c}))(\mathbf{x})
\end{equation}
\textit{\underline{Role}:} ModLin layers and the ModMLP can be interpreted as \emph{programmable} neural modules, where the \emph{program} is specified by the condition or code vector $\mathbf{c}$. 

\textbf{ModAttn.} ModAttn is a conditional variant of the kernel modulated dot product attention (KMDPA) \citep{rahaman2020s2rms}, where the key, query and value vectors are obtained from ModLin layers conditioned by a vector. In our case, this vector is the code vector $\mathbf{c}_u$ of function $f_u$, and the corresponding key, query and value vectors are computed as follows (with $\{\mathbf{x}_i\}_i$ as input and $h$ indexing attention heads): 
\begin{equation} \label{eq:mod_qkv}
    \mathbf{k}_{uhi} = \text{ModLin}_{\text{key}}^h(\mathbf{x}_i, \mathbf{c}_u) \qquad \mathbf{q}_{uhi} = \text{ModLin}_{\text{query}}^h(\mathbf{x}_i, \mathbf{c}_u) \qquad \mathbf{v}_{uhi} = \text{ModLin}_{\text{value}}^h(\mathbf{x}_i, \mathbf{c}_u) 
\end{equation}
Note that further below (e.g., in Equation~\ref{eq:loc}), we will encounter $\mathbf{x}_{ui}$, where the extra $u$ in the subscript denotes that the set element at index $i$ is specific to the function $u$; in this case, $\mathbf{x}_i$ is substituted with $\mathbf{x}_{ui}$ in Equation~\ref{eq:mod_qkv}. 
Next, given the keys, queries and the function-variable compatibility matrix $C_{ui}$, the modulated self-attention weights $W_{uhij}$ are given by: 
\begin{align} \label{eq:mod_attn_weights}
W_{uhij} = \frac{\tilde W_{uhij}}{\epsilon + \tilde W_{uhij}} \quad \text{where} \quad
\tilde W_{uhij} = C_{ui} C_{uj} \left[\text{softmax}_{j}\left(\frac{\mathbf{q}_{uhi} \cdot \mathbf{k}_{uhj}}{\sqrt{d_{\text{key}}}}\right)\right]
\end{align}
Here, the quantity $W_{uhij}$ denotes the attention weights in function $f_u$ between elements $\mathbf{x}_i$ and $\mathbf{x}_j$ at head $h$ and the softmax operation normalizes along $j$; intuitively, information about $\mathbf{x}_i$ and $\mathbf{x}_j$ is mixed by $f_u$ at head $h$ only if $W_{uhij} \ne 0$. Now, on the one hand, this can be the case only if both $C_{ui}$ and $C_{uj}$ are non-zero, i.e., $f_u$ is granted access to both variables $\mathbf{x}_i$ and $\mathbf{x}_j$ by the typing mechanism. But on the other hand, $f_u$ does not necessarily mix $\mathbf{x}_i$ and $\mathbf{x}_j$ even if both $C_{ui}$ and $C_{uj}$ are non-zero, for the self-attention weights (square brackets) may still be close to zero depending on the context (i.e., the content of $\mathbf{x}_i$ and $\mathbf{x}_j$). Next, the values are linearly mixed using the computed attention weights, which is then processed by a final ModLin layer to yield the output $\mathbf{y}_{ui}$:
\begin{equation} \label{eq:modlin_mixheads}
\mathbf{y}_{ui} = \text{ModLin}(\mathbf{\tilde y}_{ui;h}; \mathbf{c}_u) \quad \text{where} \quad \mathbf{\tilde y}_{uhi} = \textstyle\sum_j W_{uhij} \mathbf{v}_{uhj}
\end{equation}
Here, $\mathbf{\tilde y}_{ui;h}$ means the head-axis is folded into channels. \textit{\underline{Role}:} ModAttn enables interaction between the elements of its input set in multiple parallel streams, one for each function. The query, key, value, and output projectors of each stream are conditioned on the corresponding code vectors, and the interaction between elements in each stream is weighted by their compatibility with the said function.

\looseness=-1
\textbf{Line of Code (LOC).} An LOC layer is a ModAttn layer followed by a ModMLP layer (Figure~\ref{fig:ni_overview}, rightmost),
where both layers share the same condition vector $\mathbf{c}_u$, and there are weighted residual connections between the layers. Assuming inputs $\{\mathbf{x}_{ui}\}_{u,i}$ to the LOC, we have:
\begin{align} \label{eq:loc}
\mathbf{\tilde a}_{ui} = \text{ModAttn}(\{\text{LayerNorm}(\mathbf{x}_{uj})\}_j; \mathbf{c}_{u}, \{C_{uj}\}_j) \quad & \quad \mathbf{a}_{ui} = \mathbf{x}_{ui} + C_{ui} \mathbf{\tilde a}_{ui} \\
\label{eq:loc_2}
\mathbf{\tilde b}_{ui} = \text{ModMLP}(\text{LayerNorm}(\mathbf{a}_{ui}); \mathbf{c}_u) \quad & \quad \textbf{y}_{ui} = \mathbf{a}_{ui} + C_{ui} \mathbf{\tilde b}_{ui}
\end{align}
This parameterization ensures that $\mathbf{y}_{ui} = \mathbf{x}_{ui}$ if $C_{ui} = 0$. In words, if a function ($f_u$) is not granted access to a variable ($\mathbf{x}_i$) by the typing mechanism, it acts as an identity function for this variable. Further, note that we allow the input set to be indexed only by $i$; in this case, we assume $\mathbf{x}_{ui} = \mathbf{x}_i$ for all $u$. 
\textit{\underline{Role}:} A LOC can be thought of as multiple instances of a layer in the original transformer architecture (comprising a self-attention and a MLP module with residual connections), applied in parallel streams, one per function. Computations therein are conditioned on
the respective code and signature vectors. 

\textbf{Interpreter.} The interpreter layer
is a stack of $n_l$ LOCs sharing the same function codes $\mathbf{c}_u$ and function-variable compatibilities $C_{ui}$. Assuming the input to the interpreter is a set $\{\mathbf{x}_i\}_{i}$, we have: 
\begin{equation} \label{eq:interpreter}
\mathbf{y}_i = \mathbf{x}_i + \textstyle\sum_u C_{ui} \, (\text{LOC}_{n_l} \circ ... (n_l \text{~times}) ... \circ \text{LOC}_{1})(\{\mathbf{x}_j\}_j; \mathbf{c}_u, \{C_{uj}\}_j)
\end{equation}
In words: the interpreter broadcasts a given set element to multiple parallel computational streams, one for each function. After the streams have processed their copy of the input element, the results are aggregated by a weighted sum over the streams, where the weights correspond to the compatibility of the input element with the respective function. 
Equation~\ref{eq:interpreter} can be justified by making two observations. First, if a variable $\mathbf{x}_i$ is not matched with any function by the typing mechanism, it is left unmodified by the interpreter; i.e., if $C_{ui} = 0$ for all $u$, then $\mathbf{y}_i = \mathbf{x}_i$. This allows signals to be propagated through the interpreter without interference from existing functions, if so determined by the type inference module. Second, the additive aggregation over the function index $u$ implies that the overall parameterization of Neural Interpreters does not depend on the number of functions. This allows one to add a new function $f_v$ simply by including its signature and code $(\mathbf{s}_v, \mathbf{c}_v)$ as learnable parameters and finetuning these on (say) a new problem. 
\textit{\underline{Role}:} The interpreter serves as a general-purpose instruction executor (one that is shared between functions). Given a set of inputs and an instruction (here, the function code), it executes said instruction to compute the output. 

\looseness=-1
\textbf{Function Iterations in Script.} Having expressed the overall model as a stack of multiple scripts, we are now equipped to describe the computational graph of a single script. A script can be expressed as a recurrent application of the type matching mechanism and the interpreter, where we refer to the composition of the latter two as a Function Iteration (FnIter):
\begin{align}\label{eq:ftn_iter}
    \{\mathbf{y}_i\}_i = \text{FnIter}(\{\mathbf{x}_j\}_j) &= \text{Interpreter}(\{\mathbf{x}_{j}\}_{j}, \{\mathbf{c}_{u}\}_u, \{C_{uj}\}_{u,j}) \\
    \text{where~} C_{uj} &= \text{TypeMatching}(\mathbf{s}_u, \mathbf{x}_j)
\end{align}
Here, the TypeMatching component encapsulates both type inference and kernel matching, as detailed in Equation~\ref{eq:type_inference}. 
A script (cf. Equation~\ref{eq:script}) can now be expressed as a recurrent application of FnIter: 
\begin{align}\label{eq:recurrence}
\{\mathbf{y}_j\}_j = (\text{FnIter} \circ ... \circ (n_i \text{~times}) \circ ... \circ \text{FnIter})(\{\mathbf{x}_i\}_i)
\end{align}
\textit{\underline{Role}:} Inside a script, function iterations enable sharing of computational units in depth. Increasing the number of function iterations can increase depth without increasing the number of parameters. 

\looseness=-1
\textbf{Preventing function signatures and variable types from collapsing on each other.} One might obtain a scenario where the signatures of all functions and the types of all possible set elements collapse to a single point in type-space. This causes all set elements to be routed to all functions, thereby undermining the inductive bias of modularity. One effective way of preventing this degeneracy is to keep the function signatures fixed at initialization (i.e. a high-entropy distribution). This effectively encourages the (learnable) type-inference module to produce diverse types, in order to use all available functions. 

\looseness=-1
\textbf{In summary,} we observe that any input element $\mathbf{x}_i$ may trace a large number of \emph{computational paths} as it progresses through the model, where a computational path is a partially ordered set (poset) of functions (refer to Figure~\ref{fig:digits_compgraphs} for a visualization of such computational paths). In particular, owing to the fact that we recurrently apply function iterations in scripts, this poset may contain repeated functions, enabling weight sharing in depth. Further, the use of an interpreter that is shared between functions but explicitly conditioned (or \emph{programmed}) by their code vector allows us to remain independent of the number of functions and retain the ability to add more functions after the model has been trained. Future work may investigate manipulating the code vectors themselves, thereby enabling higher-order functions (i.e., functions that manipulate other functions).

\vspace{-3pt}
\section{Related Work}
\vspace{-3pt}

\looseness=-1
\textbf{Modularity.} Despite recent successes of deep neural networks, their ability to recombine and reuse meaningful units for systematic compositional generalization is shown to be limited \citep{lake2018generalization, loula2018rearranging,keysers2019measuring}. 
To mitigate this issue, work has been done to modularize deep architectures to enable the reuse and recombination of learned modules \citep{andreas2016neural,chang2018automatically,hudson2018compositional, rosenbaum2017routing,shazeer2017outrageously,fernando2017pathnet,kirsch2018modular}, and connections have been drawn to the principle of Independent Causal Mechanisms \citep{ scholkopf2012causal,peters2017elements, ParKilRojSch18, goyal2019recurrent,besserve2020theory, rahaman2020s2rms,goyal2020object,lamb2021transformers,goyal2021neural}.
Modular architectures vary in how the modules are learned and laid out during inference. For instance, Neural Module Networks \citep{andreas2016neural} and Neural Event Semantics \citep{buch2021nes} dynamically compose the modules using a natural language parser, whereas neuro-symbolic architectures \citep{van2018relational,goyal2019recurrent, mao2019neuro, yi2019clevrer,goyal2020inductive,locatello2020objectcentric, goyal2020object,wu2020scattering,goyal2021neural} typically require entity-centric representations \citep{van2018relational, yi2019clevrer, locatello2020objectcentric} but use neural components like attention for compositional reasoning. 
Unlike some of these methods, Neural Interpreters (NI) neither require domain knowledge nor object-centric representations. To obtain a strong compositional inductive bias, NIs aim at factorizing the computational graph in terms of functional units which are dynamically recombined in order to solve a particular task. In comparison with SCOFF~\citep{goyal2020object} and NPS~\citep{goyal2021neural} which also separate functions, arguments and values and enable a similar dynamic and composed application of functions, NIs introduce the notion of interpreter (which enables parameter sharing between functions and extending the set of functions easily) and sophisticated signature and typing mechanisms described in the previous section.

\looseness=-1
\textbf{Dynamic Routing of Information.} The systematic generalization of modular architectures is shown to depend on the layout of modules \citep{bahdanau2018systematic,rosenbaum2019routing}.
There exist different frameworks to facilitate the joint learning of module layout and parameters \citep{sabour2017dynamic, rosenbaum2017routing, rahaman2020s2rms, goyal2019recurrent,fernando2017pathnet, shazeer2017outrageously, buch2021nes}. For example, \citep{rosenbaum2017routing} uses collaborative multi-agent reinforcement learned routing, \citep{shazeer2017outrageously} uses a trainable gating network. In this work, we use kernel modulated dot product attention \citep{rahaman2020s2rms} between function signatures and the input types to assign inputs to functions at each step.

\looseness=-1
\textbf{Transformers.} Modularity in Transformers \citep{vaswani2017attention} is relatively less studied. Repulsive attention \citep{an2020repulsive} is based on repelling transformer heads to encourage specialization, whereas TIM \citep{lamb2021transformers} draws on the principle of independent mechanisms by dividing computation over parallel transformers that interact via competitive top-k attention \citep{goyal2019recurrent}. 
In contrast, functions in NIs iteratively acquire set elements with kernel dot-product attention, which, if need be, can uniformly activate all the functions for a given token, encouraging efficient use of computation. 
Closer to Neural Interpreters, Switch Transformers (STs) \citep{fedus2021switch} employ a mixture of experts framework where experts specialize across tokens, guided by a routing scheme which activates only one expert per token. 
However, STs focus more on dynamically \textit{selecting} experts, whereas Neural Interpreters focus on dynamically \textit{composing} available experts. 
The starting point of this work is the vision transformer (ViT) \citep{cordonnier2019relationship, dosovitskiy2020image},
which applies a standard Transformer directly to image patches. We use the same architectural scaffolding as ViTs, but implement additional inductive biases that facilitate fast adaptation and systematic generalization.

\section{Experiments}
\vspace{-4pt}
In this section, we empirically evaluate Neural Interpreters in a variety of problem settings. The questions we seek to answer are the following. \textbf{(a)} Can Neural Interpreters learn reusable computational units, given a number of training tasks that can in principle be solved with a fixed set of primitives? (Section~\ref{sec:fuzzybool}) \textbf{(b)} Do Neural Interpreters modularize information in a way that helps fast adaptation \citep{bengio2019metatransfer}? (Section~\ref{sec:digits}) \textbf{(c)} Can Neural Interpreters gracefully handle a larger number of functions (computational units) than they were trained with? (Section~\ref{sec:digits}) \textbf{(d)} Do the inductive biases help with systematic generalization required in abstract reasoning problems? (Section~\ref{sec:pgm}).

\vspace{-3pt}
\subsection{Learning Fuzzy Boolean Expressions} \label{sec:fuzzybool}
\vspace{-2pt}
In this section, we construct a toy problem to investigate whether neural interpreters can learn reusable functions when trained on a large number of tasks that share the same underlying building blocks. 

\linepenalty=100
\textbf{Task Definition.} Consider a set of $N$ scalars $\{x_1, ..., x_N\}$, where $x_i \in [0, 1]$ is a real number between $0$ and $1$ (inclusive). We now define the following operations on the elements of this set: 
\begin{align} \label{eq:fuzzbool}
\texttt{and}(x_i, x_j) = x_i x_j; \quad \texttt{not}(x_i) = \bar{x}_i = 1 - x_i; \quad \texttt{or}(x_i, x_j) = x_i \oplus x_j = \overline{\bar{x}_i \bar{x}_j}
\end{align}
Note that the above operations map from $[0, 1]^2$ to $[0, 1]$; if $x_i, x_j \in \{0, 1\}$, they reduce to their Boolean namesakes. By combining these primitives, it is possible to construct and sample from a family of $2^{2N}$ \emph{fuzzy} Boolean functions mapping from $[0, 1]^N$ to $[0, 1]$ (Appendix~\ref{app:fuzzy-boolean}). 
The problem is now set-up as follows. We sample 30 random fuzzy Boolean functions of $N = 5$ variables, $\{f_i\}_{i=1}^{30}$ where $f_i: [0, 1]^5 \to [0, 1]$, of which we use 20 for the pre-training phase and reserve 10 for the later adaptation phase. For each function, we sample 163840 points from a uniform distribution over $[0, 1]^5$, of which we use 80\% for training, and the remaining for validation. This yields two multi-task regression datasets, one for pre-training and the other for adaptation. A sample from the pre-training dataset comprises a 5D input vector $\mathbf{x} \in [0, 1]^5$ and the scalar regression targets $f_1(\mathbf{x}), ..., f_{20}(\mathbf{x})$. 

\begin{wraptable}{r}{0.52\textwidth}
\vspace{-13.5pt}
\caption{Mean Coefficient of Determination ($R^2$) and StdDev over 10 tasks after training various sets of parameters. \textbf{Gist:} By just allowing the functions to \textit{rewire} themselves by training only the type-matching parameters, one approaches the performance of tuning all remaining parameters, including those of the functions themselves. This suggests that the functions have learned recomposable primitives.}
\label{tab:logic}
\begin{tabular}{l|l}
\toprule
\multicolumn{1}{c}{\textbf{Parameter Set}} & \multicolumn{1}{c}{$\mathbf{R^2}$}  \\ \midrule
All Parameters (Pretraining)                & $0.9983 \pm 0.0005$                 \\ \midrule
Finetuning CLS Tokens                       & $0.9202 \pm 0.0198$                 \\ 
\hspace{5pt} + Type Inference Parameters    & $0.9857 \pm 0.0034$                 \\
\hspace{5pt} + Remaining Parameters         & $0.9953 \pm 0.0013$                 \\
\bottomrule
\end{tabular}
\vspace{-10pt}
\end{wraptable}
\looseness=-1
\textbf{Method.} We pre-train a Neural Interpreter to regress all $20$ functions simultaneously. To do this, we feed it a set of 25 elements as input: the first 5 are the components of $\mathbf{x}$ (with learnable positional embeddings \citep{radford2019language} added in) and the remaining 20 are learned CLS tokens (vectors), one for each function. 
We attach a (shared) regression head to the output elements corresponding to the CLS tokens, and train their outputs to regress the respective function. Having pre-trained the model for 20 epochs, we finetune it for an additional 3 epochs on the 10 reserved functions $\{f_{21}, ..., f_{30}\}$. For the latter, we always instantiate and train 10 new CLS tokens, corresponding to the new functions; however, we investigate three settings in which three different sets of parameters remain frozen. In the first setting, all parameters remain frozen, implying that the CLS tokens are the only trainable parameters. In the second setting, we unfreeze the parameters of the type matching mechanism (function signatures and parameters of the type inference MLP, in addition the CLS tokens). In the third setting, we unfreeze all parameters and finetune the entire model. Additional details in Appendix~\ref{app:fuzzy-boolean}. 

\looseness=-1
\textbf{Hypothesis.} By finetuning just the type-matching parameters, we only permit adaptation in how information is routed through the network. In other words, we only allow the computational units to \emph{rewire} themselves in order to adapt to the new task at hand, while preserving their functionality. Now if the computational primitives that are learned during pre-training are recomposable, one would expect the performance of Neural Interpreters having finetuned just the type-matching parameters to approach that obtained by finetuning all the parameters, the latter including ones that determine the functionality of the computational primitives. 

\linepenalty=2000
\textbf{Results.} Table~\ref{tab:logic} compares the coefficients of determination\footnote{$R^2 = 1$ implies perfect fit; a model that regresses to the mean has $R^2 = 0$.} ($R^2$) obtained in each of the investigated finetuning settings. We find that relative to finetuning just the CLS tokens, the performance difference between finetuning all parameters and just the type matching parameters is small. This is in line with expectation, suggesting that Neural Interpreters are indeed capable of learning recomposable computational primitives.

\begin{figure}[t]
\begin{subfigure}[t]{0.32\textwidth} 
\centering
\includegraphics[width=1\textwidth]{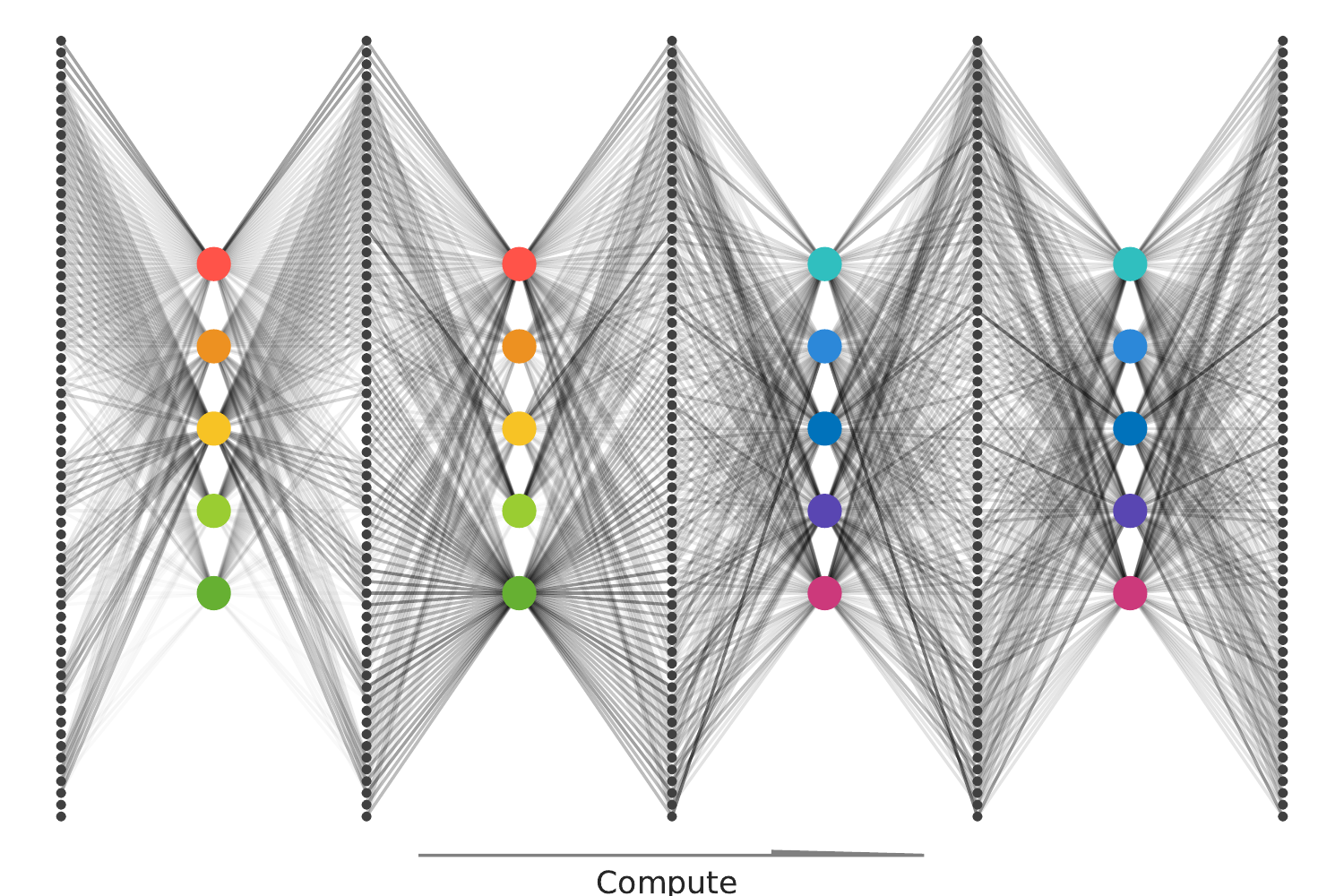}
\end{subfigure}
\hfill
\begin{subfigure}[t]{0.32\textwidth}
\centering
\includegraphics[width=1\textwidth]{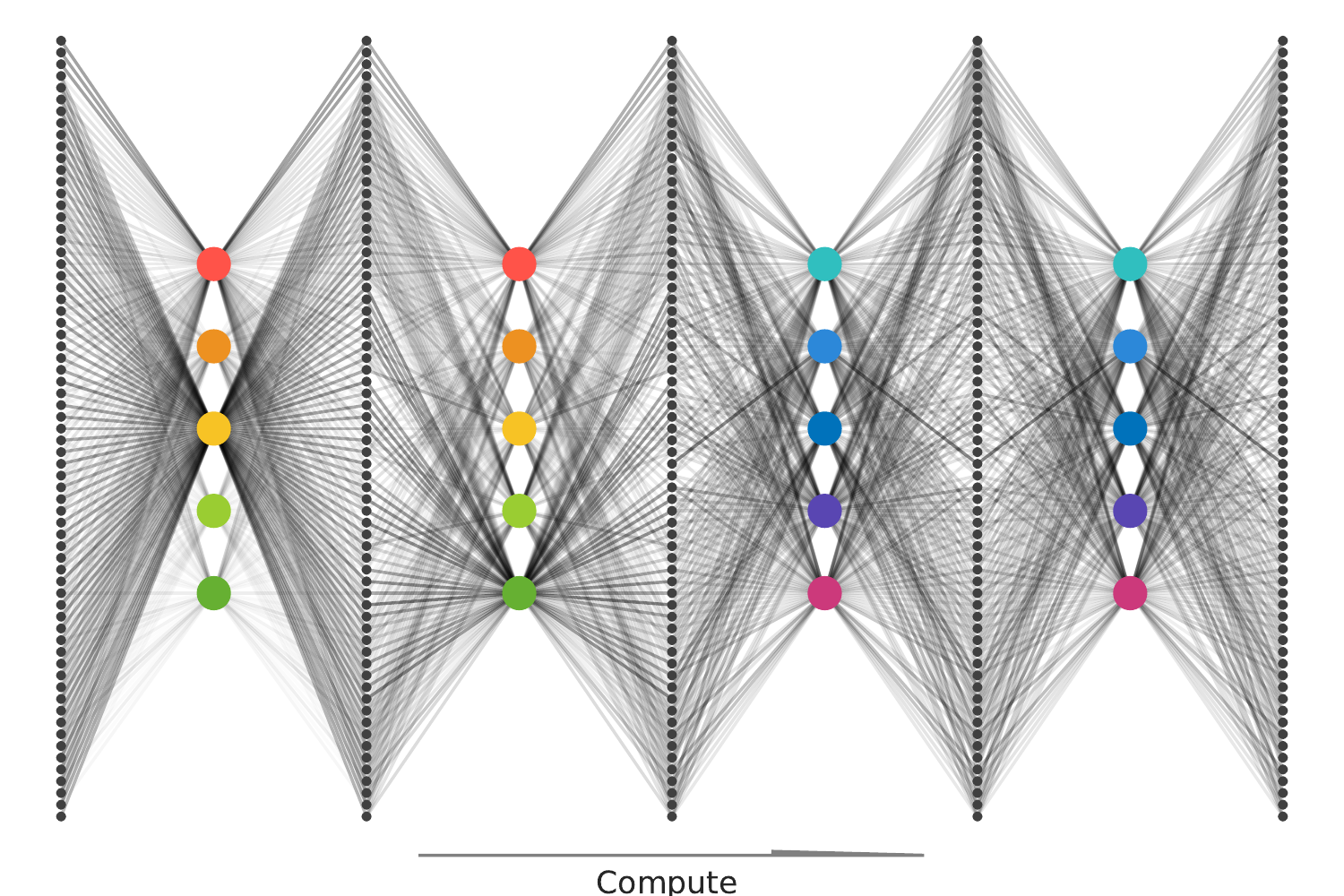}
\end{subfigure} 
\hfill
\begin{subfigure}[t]{0.32\textwidth}
\centering
\includegraphics[width=1\textwidth]{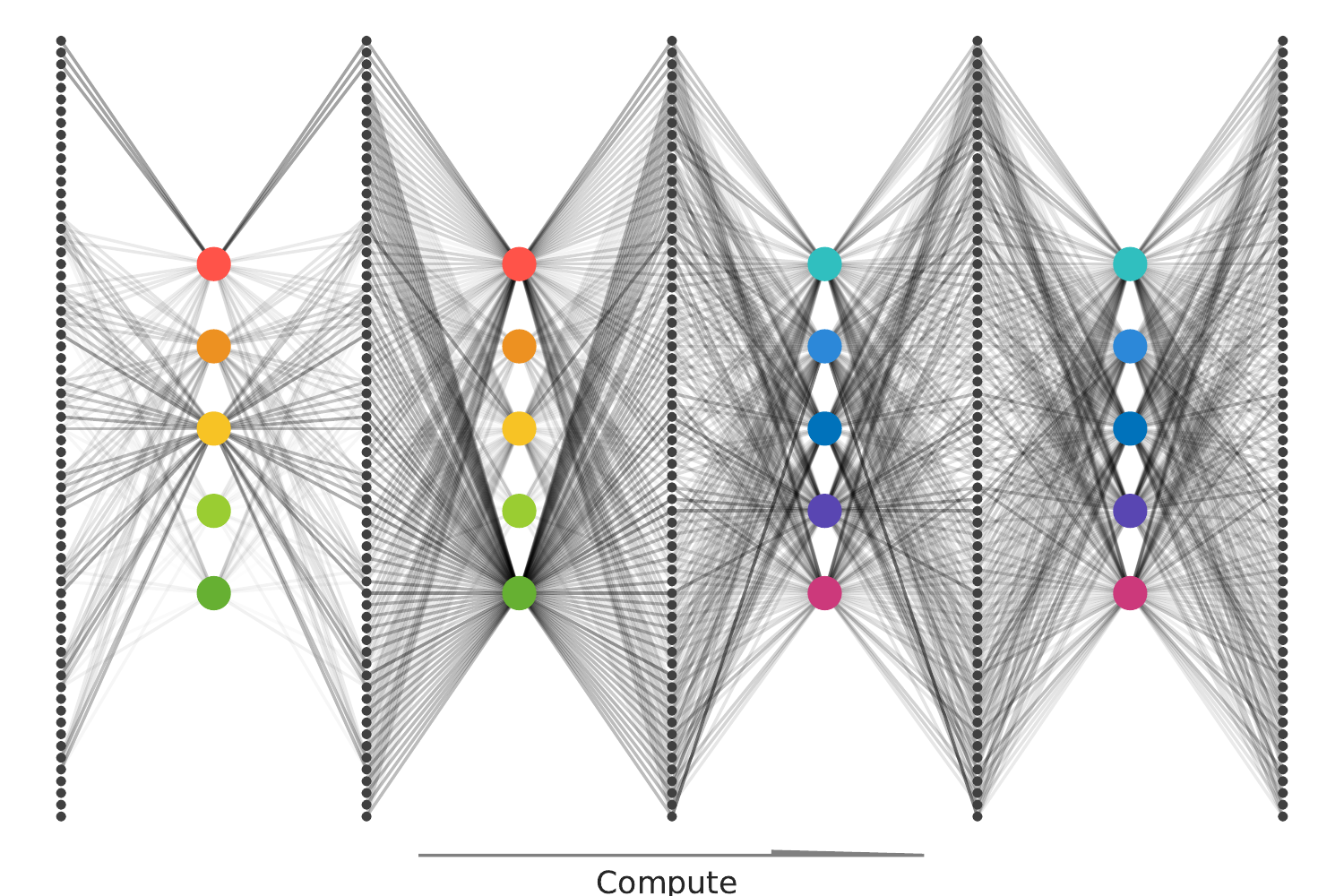}
\end{subfigure} 
\caption{
\looseness=-1
\small Visualization of computational paths taken by input set elements corresponding to three different samples as they progress through a Neural Interpreter. Colored dots identify functions (same color implies shared parameters), and the weight of the lines denote their compatibility with set elements. \textbf{Gist:} There are variations (but also similarities) between samples in how their constituent set elements are routed through the network.} \label{fig:digits_compgraphs}
\vspace{-13pt}
\end{figure}
\begin{figure}[t]
\begin{subfigure}[t]{0.24\textwidth} 
\centering
\includegraphics[width=1\textwidth]{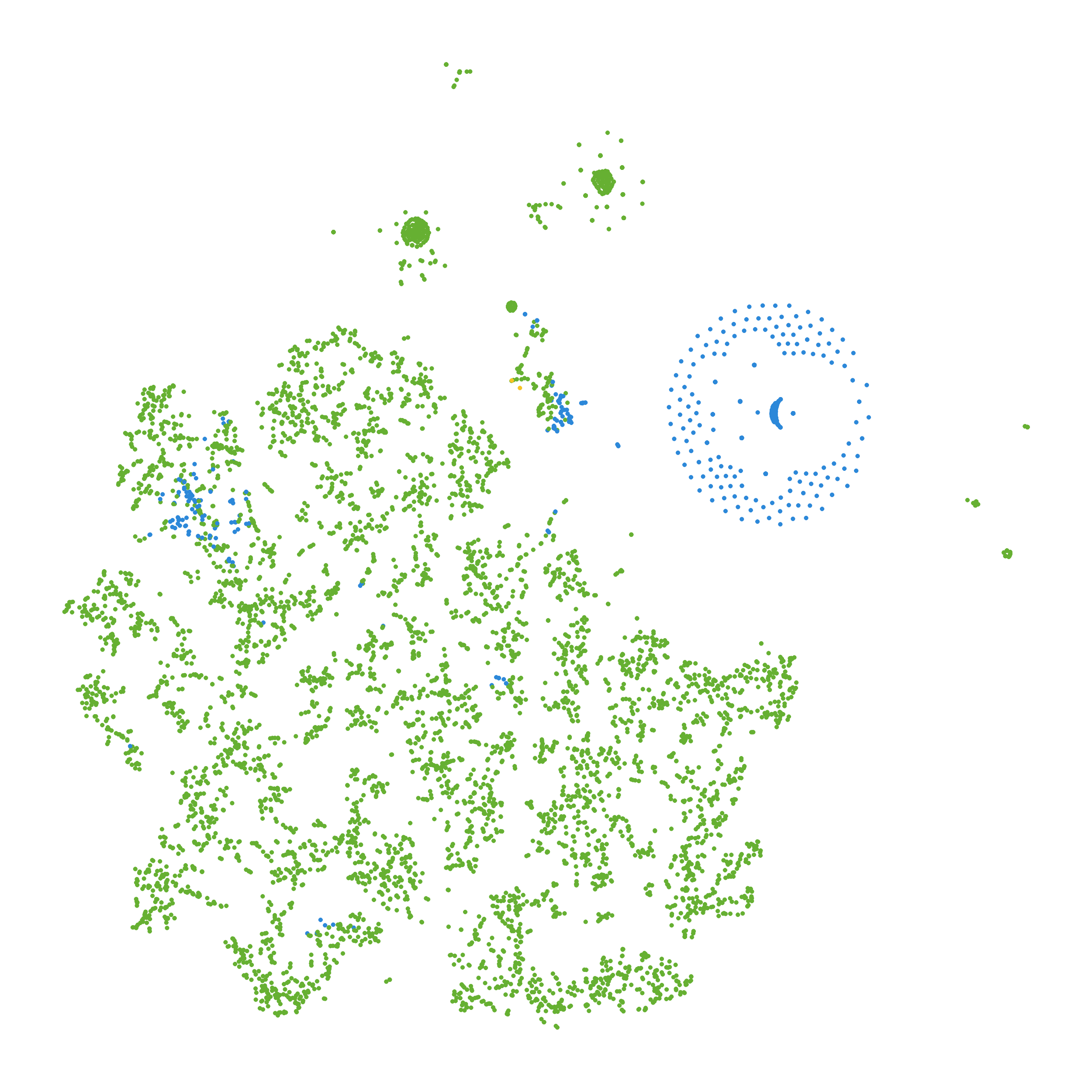}
\end{subfigure}
\hfill
\begin{subfigure}[t]{0.24\textwidth}
\centering
\includegraphics[width=1\textwidth]{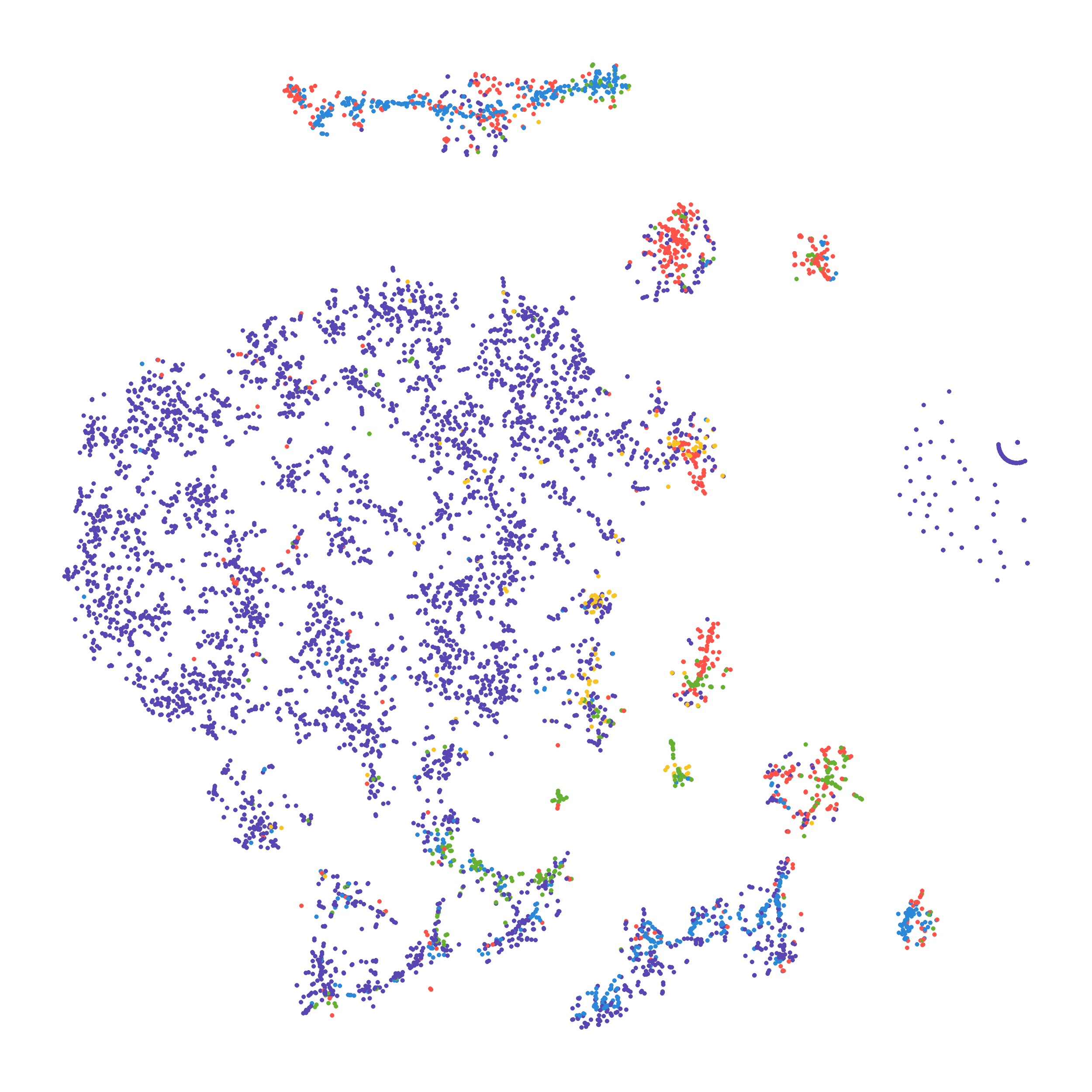}
\end{subfigure} 
\hfill
\begin{subfigure}[t]{0.24\textwidth}
\centering
\includegraphics[width=1\textwidth]{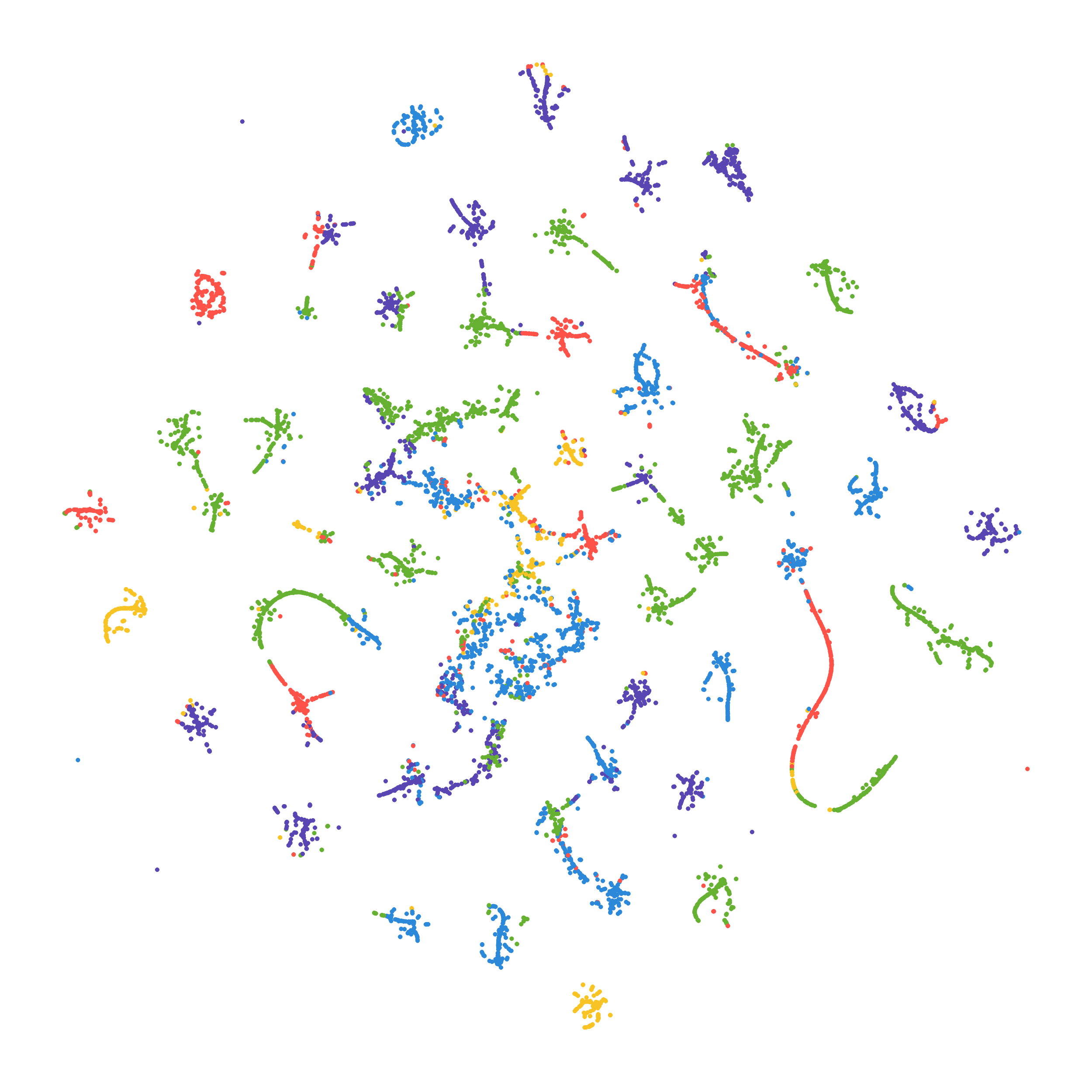}
\end{subfigure} 
\hfill
\begin{subfigure}[t]{0.24\textwidth}
\centering
\includegraphics[width=1\textwidth]{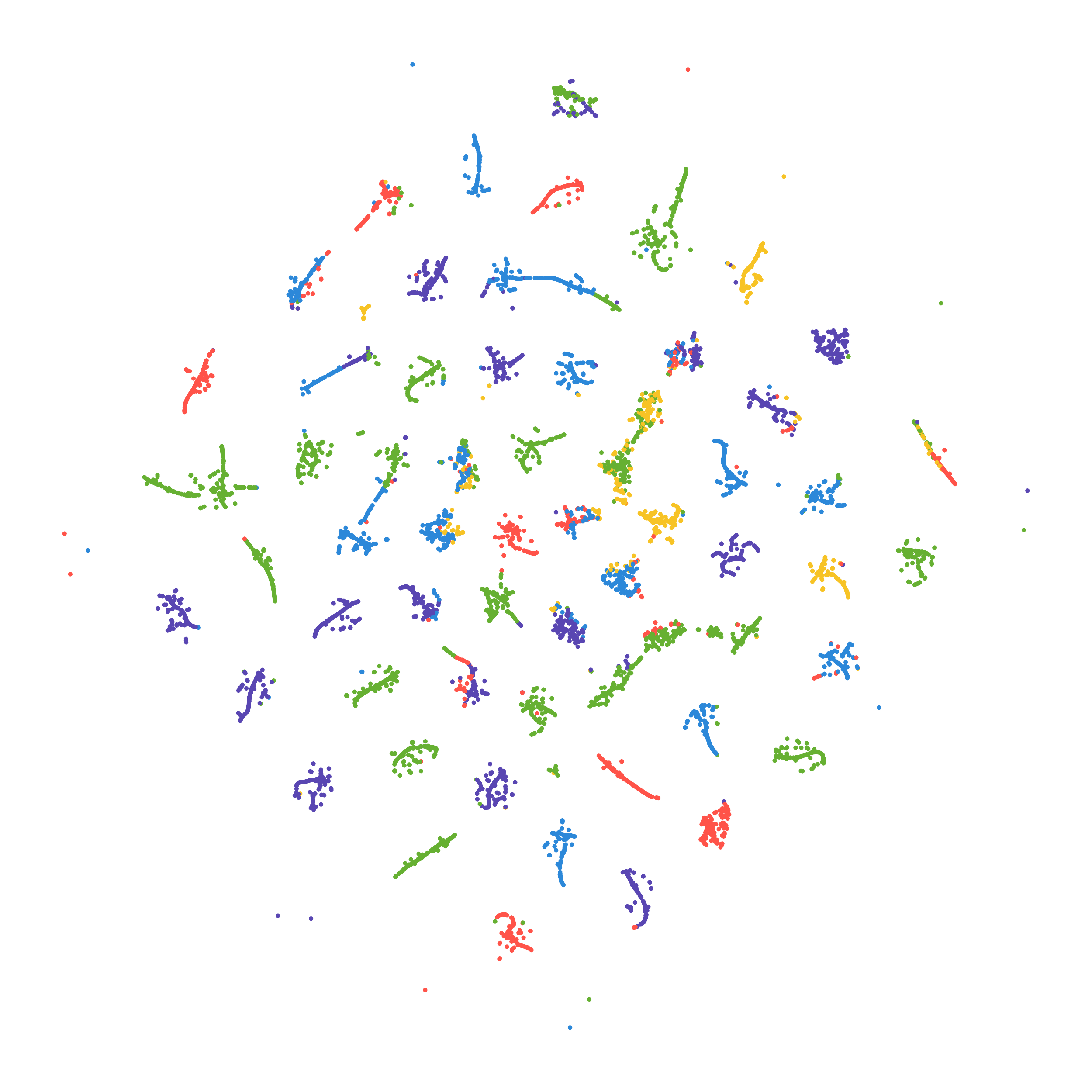}
\end{subfigure} 
\caption{
\small t-SNE embeddings of the inferred types of set elements as they progress through a Neural Interpreter with two scripts with two function iterations each. The color identifies the closest function in type space, and the progression from left to right is over the function iterations. \textbf{Gist:} Types are more clustered in the later function iterations, suggesting that the input set elements gradually \textit{develop} a type as they progress through the network.} \label{fig:digits_tsne}
\vspace{-15pt}
\end{figure}

\vspace{-3pt}
\subsection{Multi-Task Image Classification} \label{sec:digits}

In this section, we evaluate Neural Interpreters in a multi-task image classification setting. Our goals are (a) to determine whether the inductive bias helps with fast adaptation, (b) to investigate whether the interpreter can indeed function as a general instruction executor, as intuited in Section~\ref{sec:ni}, and (c) to demonstrate that the proposed architecture produces modules that can function autonomously (without additional training objectives that encourage this behaviour). Additional results in Appendix~\ref{app:digits} analyze the effects of varying hyper-parameters. 

\textbf{Task Definition.} We consider three related datasets sharing the same label semantics, namely SVHN \citep{netzer2011reading}, MNISTM \citep{ganin2016domain} and MNIST \citep{lecun2010mnist}. The images therein are upsampled to shape $32 \times 32$ (if required), and the resulting image is augmented with RandAugment \citep{cubuk2019randaugment}. Subsequently, the augmented images are split \citep{cordonnier2019relationship} to 64 patches of shape $4 \times 4$. 
In addition, we also use unaugmented samples from the K-MNIST dataset \citep{clanuwat2018deep} of Hiragana characters to probe fast-adaptation to new data. 

\textbf{Method.} We train Neural Interpreters for 100 epochs on the combined \emph{digits} dataset described above. The input set contains 67 vector valued elements -- the first 64 corresponding to linear embeddings of the $4 \times 4$ patches, and the remaining 3 to learnable CLS tokens (vectors), one for each dataset. For each CLS token, a linear classification head is attached to the corresponding output; given an input sample from a certain dataset, the respective classification head is trained to predict the correct class.
To inject position information into the model, we use a variant of the relative positional encoding scheme described in \citep{cordonnier2019relationship}, but applied only to the first 64 input elements (corresponding to the image patches). Having pre-trained on the digits dataset, we finetune the model on K-MNIST with varying numbers of samples for 10 epochs. For additional details, please refer to Appendix~\ref{app:digits}.

\textbf{Baseline.} Vision Transformers (ViT) \citep{cordonnier2019relationship, dosovitskiy2020image} make the natural baseline for Neural Interpreters, given the fact that the former is a special case of the latter. The set-up with CLS tokens and classification heads is identical to that of Neural Interpreters, as is the training protocol. We use a light-weight 8-layer deep model with approximately 1.8M parameters, but ensure that the considered Neural Interpreter model is even lighter, with roughly 0.6M parameters. 

\begin{figure*}
\begin{subfigure}[t]{0.32\textwidth} 
\centering
\includegraphics[width=1\textwidth]{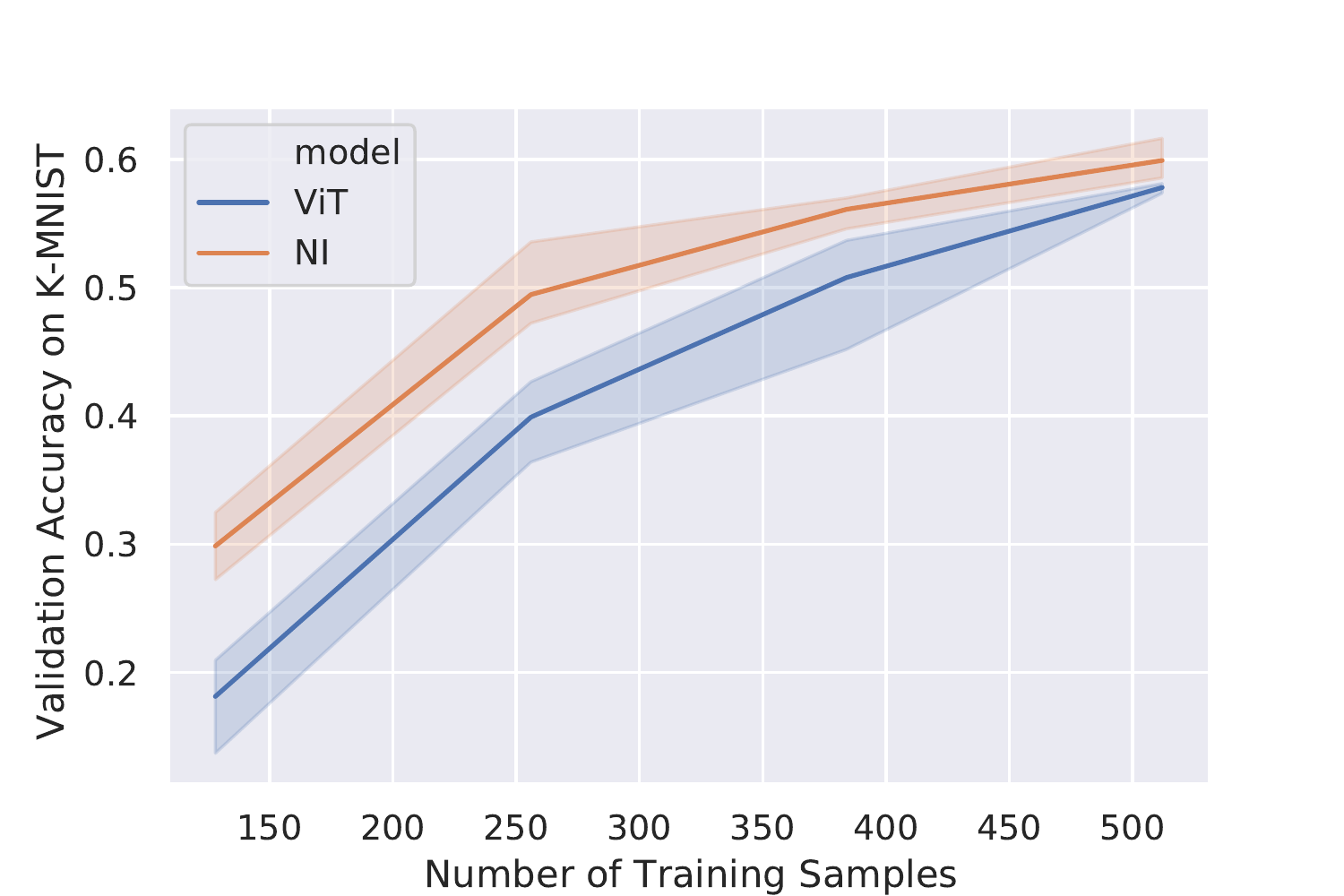}
\caption{\small Fast adaptation} \label{fig:digits_fastadapt}
\end{subfigure}
\hfill
\begin{subfigure}[t]{0.32\textwidth}
\centering
\includegraphics[width=1\textwidth]{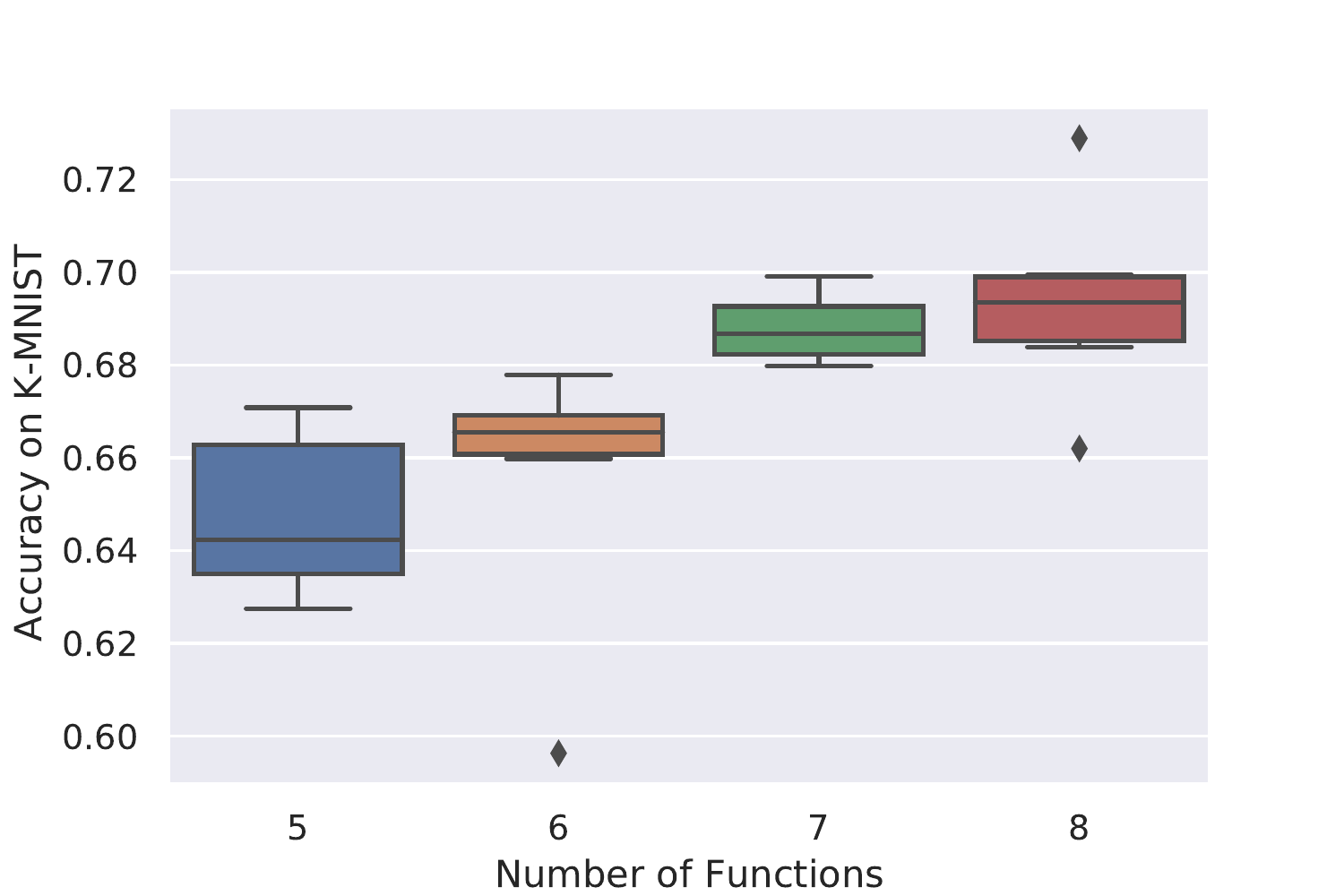}
\caption{\small Capacity extension} \label{fig:digits_capexp}
\end{subfigure} 
\hfill
\begin{subfigure}[t]{0.32\textwidth}
\centering
\includegraphics[width=0.88\textwidth]{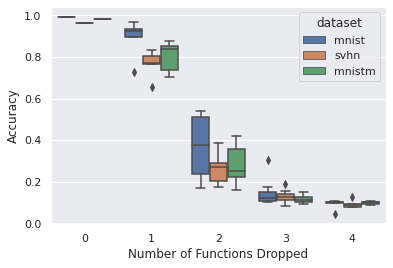}
\caption{\small Dropped functions} \label{fig:digits_fndrop}
\end{subfigure}
\caption{\small \textit{\textbf{Figure \ref{fig:digits_fastadapt}:}} Performance of Neural Interpreters (NI) compared to that of a Vision Transformer (ViT) that performs equally well on the validation set (y-axis), plotted against the number of training samples presented (x-axis). We see Neural Interpreters can adapt faster in the low-data regime. \textit{\textbf{Figure \ref{fig:digits_capexp}:}} Validation performance (y-axis) of a Neural Interpreter trained with 5 functions but finetuned with various number of functions (x-axis).  We see that performance increases with increasing functions, showing that the model does not \textit{overfit} to the number of functions it was trained with. \textbf{\textit{Figure \ref{fig:digits_fndrop}:}} Accuracy on the digits dataset as a function of the number of dropped functions. We find that the performance degrades gracefully as functions are dropped, suggesting that learned functions are autonomous, i.e. they can operate in the absence of other functions.} \label{fig:digits}
\vspace{-20pt}
\end{figure*}

\begin{wrapfigure}{r}{0.37\textwidth}
\vspace{-14pt}
\centering
\includegraphics[width=\linewidth]{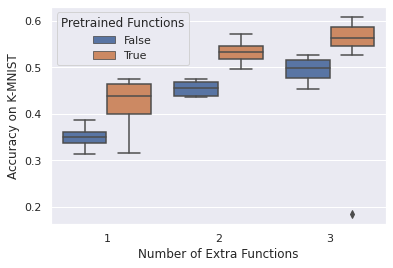}
\caption{
\small Performance on K-MNIST after adaptation. In one case (orange), functions pretrained from the digits dataset are kept intact. In another case (blue), function codes and signatures are randomly sampled before adaptation. We find that the former performs better. \textbf{Gist:} Knowledge acquired during pretraining is indeed reused during adaptation. This is apparent when comparing with a baseline where this knowledge is destroyed, leading to decreased performance after adaptation. \label{fig:digits_pretrainedvsnot}}
\vspace{-10pt}
\end{wrapfigure}
\looseness=-1
\textbf{Hypotheses.} \textbf{(1)} It has been suggested that a model that appropriately modularizes the learning problem should excel at fast-adaptation to a new but related data distribution \citep{bengio2019metatransfer}. If Neural Interpreters obtain a useful modularization of the problem, we should expect it to be able to adapt faster to the K-MNIST dataset than a non-modular baseline like the ViT. This is because some of the visual primitives present in the digit datasets (e.g., motifs, strokes or shapes) might reoccur in the K-MNIST dataset, which Neural Interpreters should be able to leverage. 
\textbf{(2)} Recall that in Section~\ref{sec:ni}, we positioned the interpreter as a general purpose instruction executor, capable of being programmed by a code vector. If this is indeed the case, and if the interpreter does not overfit to the functions it was trained with, we should expect the capacity (and therefore performance) to increase as we add and train new functions (i.e., tuples of code and signatures), while keeping all other parameters fixed. \textbf{(3)} If functions (modules) are indeed autonomous, we should expect that as some of them are removed at test time, the others are still able to maintain performance. Consequently, the accuracy of the Neural Interpreter (trained with all functions) should only degrade gracefully as functions are (randomly) removed at test time. 

\looseness=-1
\textbf{Results.} Figure~\ref{fig:digits_fastadapt} compares the fast-adaptation performance of a Vision Transformer to that of a Neural Interpreter. While both models achieve almost identical performance on the validation set for all datasets (the difference being less than 0.2\% in favor of Vision Transformers), we find that Neural Interpreters significantly outperform Vision Transformers at fast-adaptation in the low-data regime, and the performance gap is only gradually closed as more data becomes available. This observation supports hypothesis \textbf{(1)}. 
Further, in Figure~\ref{fig:digits_pretrainedvsnot}, we establish a baseline where the functions pretrained on the digits dataset are destroyed prior to adaptation. We find that keeping the functions acquired from pretraining intact leads to significantly improved adaptation performance, suggesting that knowledge acquired during pretraining is indeed being reused during adaptation. This lends further support to hypothesis \textbf{(1)}. 
Figure~\ref{fig:digits_capexp} shows the validation performance of a Neural Interpreter that was pre-trained with 5 functions on the digits dataset, but tested on the K-MNIST dataset with varying numbers of functions, having finetuned just the function signatures and codes. We find that the performance improves as new functions are added at adaptation-time, supporting hypothesis \textbf{(2)}. Figure \ref{fig:digits_fndrop} shows the effect of randomly removing functions on the model accuracy. We observe that the performance on all datasets degrades gracefully as more functions are removed, thereby supporting hypothesis \textbf{(3)}.
In addition, Figure~\ref{fig:digits_compgraphs} visualizes the computational path taken by input set elements as they progress through the network, verifying that there is diversity in how samples are routed through the network, and shows that the routing mechanism discriminates between samples. Figure~\ref{fig:digits_tsne} shows that the input set elements gradually \textit{develop} a type as they progress through the model. 
Additional figures in Appendix~\ref{app:digits} analyze the effect of varying the number of scripts, number of function iterations, number of functions, kernel truncation parameter, dimension of type space $\mathcal{T}$ and freezing the function codes / signatures on the validation performance. The key finding is that, on the one hand, a wide range of hyper-parameter settings leads to performant models; on the other hand, there are patterns in what hyper-parameters perform consistently well.

\vspace{-3pt}
\subsection{Abstract Reasoning} \label{sec:pgm}
\vspace{-2pt}
In this section, our goal is to (a) use visual abstract reasoning tasks to evaluate whether Neural Interpreters are capable of systematic (compositional) generalization, and (b) characterize how Neural Interpreters maintain performance when the amount of compute is reduced at test time.

\begin{wrapfigure}{r}{0.26\textwidth}
\vspace{-12pt}
\centering
\includegraphics[width=\linewidth]{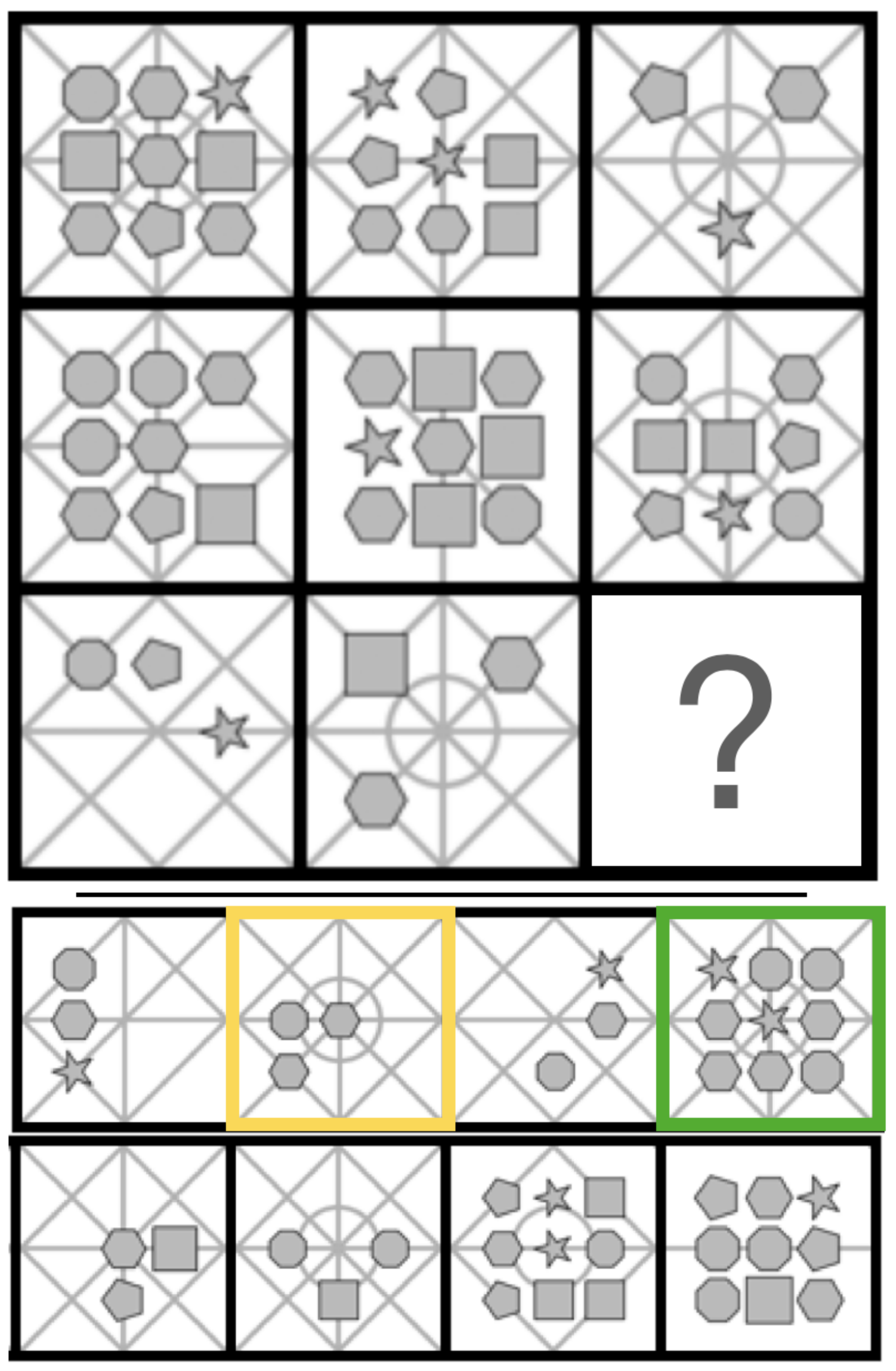}
\caption{\small A PGM task. \textbf{Top:} Context panels. \textbf{Bottom:} Candidate panels. WReN and ViT predict the wrong answer (yellow), whereas NI predicts the correct one (green).  \label{fig:pgm-sample}}
\vspace{-10pt}
\end{wrapfigure}
\looseness=-1
\textbf{Task Definition.} 
Progressively Generated Matrices (PGM) \citep{barrett2018measuring} is a procedurally generated series of datasets based on Raven Progressive Matrices (RPMs) \citep{raven1998raven}, a popular IQ test designed for humans. 
The datasets, each comprising around 1.2M samples, aim to assess abstract and analogical reasoning capabilities of machine learning models beyond surface statistical regularities \citep{jo2017measuring, rahaman2019spectral}. 
Each sample in a dataset (Figure~\ref{fig:pgm-sample}) comprises 8 context panels arranged in an incomplete $3 \times 3$ matrix, and 8 candidate answer panels (a panel being an $80 \times 80$ image). 
Given access to the context panels, the model must predict which of the candidate images is the most appropriate choice to complete the matrix. The content of the panels in a matrix are related by one or more \textit{triples}, where a triple comprises a logical rule ($\in$ progression, XOR, OR, AND, consistent union) applied to an attribute ($\in$ size, type, color, position, number) of an object of a type ($\in$ shapes, lines). 
There are 8 datasets in the series, of which 7 measure systematic generalization in different ways, i.e., the test datasets contain samples where the panels are related in ways that are not encountered during training. 
In this work, we consider 6 datasets that probe the compositional reasoning ability of the model, namely: Interpolation, Extrapolation, Held-out (H.O.) triples, H.O. pairs of triples, H.O. Attribute Pairs, and Neutral. We omit H.O. line-type and H.O. shape-type datasets, since they stress the convolutional feature extraction instead of compositional reasoning component of the model. Please refer to Appendix~\ref{app:pgm} and \citep{barrett2018measuring} for additional details. 

\textbf{Method.} Each panel (context and choice) is embedded with a shallow convolutional neural network to obtain an embedding vector. The input to the model is a set of 10 elements, comprising the embeddings of 8 context panels, that of a single candidate panel, and a CLS token (learnable vector). The model output corresponding to the CLS token is fed as input to the prediction head, which is trained to output a score measuring the compatibility of the candidate panel to the context panels. The final prediction is obtained by applying a softmax over the scores of all candidate panels. We note that this set-up resembles the one proposed in \citep{barrett2018measuring} for WReNs, and refer to Appendix~\ref{app:pgm} for details.

\textbf{Baselines.} We train Vision Transformers \citep{cordonnier2019relationship,dosovitskiy2020image} and Neural Interpreters with the same architectural scaffolding (described above). Additional baselines include Wild Relational Networks (WReN) \citep{barrett2018measuring}, Multiplex Graph Networks (MXGNet) \citep{wang2020abstract} and WReNs atop disentangled representations (VAE-WReN) \citep{steenbrugge2018improving}. Regarding the latter, we note that disentangled representations only improves on the feature extraction component of the WReN model; as such, it is entirely complementary to our contribution, which is the abstract reasoning component. Future work may explore replacing the convolutional embeddings with disentangled representations.

\begin{wrapfigure}{r}{0.38\textwidth}
\vspace{-14pt}
\centering
\includegraphics[width=\linewidth]{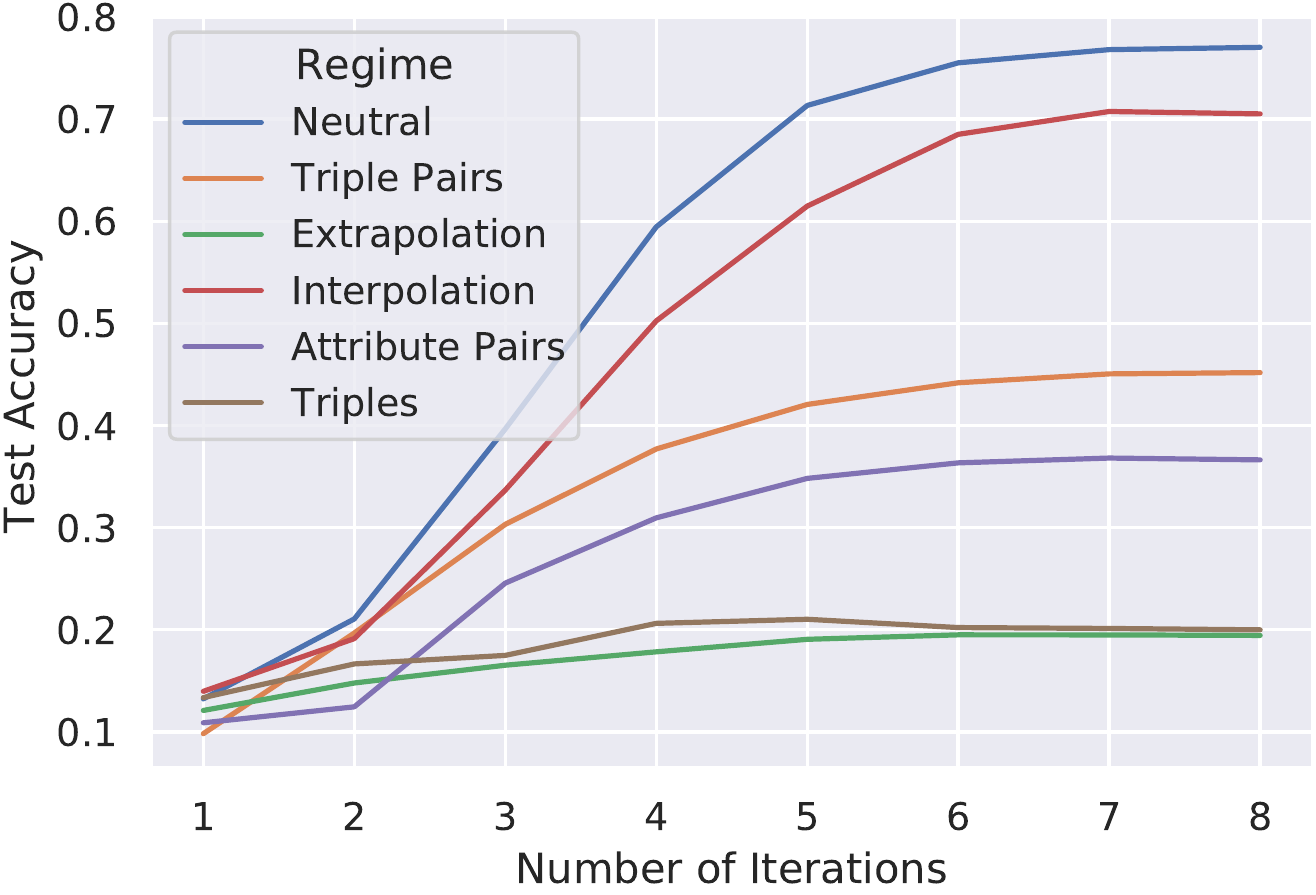}
\caption{
\small Test performance of a Neural Interpreter as a function of the number of function iterations (i.e. the amount of inference-time compute). \textbf{Gist:} Neural Interpreters are capable of trading-off performance with compute. \label{fig:pgm-anytime}}
\vspace{-10pt}
\end{wrapfigure}
\textbf{Hypothesis.} Recall that the test sets of (all but one) PGM datasets contain samples where the panels are not related in a way that is encountered during training. A model that is able to infer and reason about the underlying compositional structures is therefore more likely to be successful. 
If the inductive biases do indeed enable Neural Interpreters to factorize knowledge in to recomposable primitives, we might expect them to be able to dynamically recombine these primitives in order to generalize to problems it has not encountered during training and thereby excel at this task.

\linepenalty=2000
\textbf{Results.} Table \ref{table:pgm-main-table} tabulates the validation (in-distribution) and test (out-of-distribution) accuracies obtained by all models for the various considered datasets. We make two observations: 
(1) Neural Interpreters are competitive in terms of test accuracy, outperforming both Vision Transformers and the prior state-of-the art in 4 of 6 datasets. 
(2) Vision Transformers are competitive in-distribution: they outperform the state of the art (excluding Neural Interpreters) in 5 of 6 datasets in terms of validation accuracy. However, they are outperformed in terms of test accuracy by either Neural Interpreters or other baselines in all datasets. In addition, Figure~\ref{fig:pgm-anytime} shows how the test-set performance varies with the number of function iterations, for a model that was trained with a fixed number of function iterations (8). We note that the number of function iterations is proportional to the amount of computation used by the model. In all cases, the model maintains at least 70\% of its original performance while using only half the amount of compute; in most cases, at least 80\% of the original performance is maintained. This suggests that Neural Interpreters can function as anytime predictors \citep{zilberstein1996using}, which are models that can trade-off performance with compute.

\begin{table*}[h]
\centering
\vspace{-5pt}
\caption{Performance (prediction accuracy) of all models on different generalization regimes of PGM datasets. Note that the \emph{Val.} performance measures in-distribution generalization, and the \emph{Test} performance measures out-of-distribution generalization on the corresponding datasets (except for Neutral). \emph{Extra.} refers to Extrapolation, \textit{Attribute P.} to Attribute Pairs and \textit{Triple P.} to Triple Pairs.} 
\label{table:pgm-main-table}
\begin{tabular}{c@{\hspace{0.15cm}}|@{\hspace{0.15cm}}c@{\hspace{0.3cm}}c@{\hspace{0.15cm}}|@{\hspace{0.15cm}}c@{\hspace{0.3cm}}c@{\hspace{0.15cm}}|@{\hspace{0.15cm}}c@{\hspace{0.3cm}}c@{\hspace{0.15cm}}|@{\hspace{0.15cm}}c@{\hspace{0.3cm}}c@{\hspace{0.15cm}}|@{\hspace{0.15cm}}c@{\hspace{0.3cm}}c@{\hspace{0.15cm}}|@{\hspace{0.15cm}}c@{\hspace{0.3cm}}c}
\toprule
\multicolumn{1}{c}{\textbf{Regime}} & \multicolumn{2}{c}{Neutral } & \multicolumn{2}{c}{Interpolation } & \multicolumn{2}{c}{Attribute P.} & \multicolumn{2}{c}{Triple P.} & \multicolumn{2}{c}{Triples} & \multicolumn{2}{c}{Extra.}\\

\textbf{Model} & \textbf{Val.} & \textbf{Test} & \textbf{Val.} & \textbf{Test} & \textbf{Val.} & \textbf{Test} & \textbf{Val.} & \textbf{Test} & \textbf{Val.} & \textbf{Test} & \textbf{Val.} & \textbf{Test}\\
\midrule
WReN \citep{barrett2018measuring}  & 63.0     &  62.6 & 79.0     &  64.4 & 46.7     &  27.2 & 63.9     &  41.9 &  63.4     &  19.0 &  69.3     &  17.2 \\
VAE-WReN \citep{steenbrugge2018improving} & 64.8 &  64.2 & - & - & \textbf{70.1} &  \textbf{36.8} & 64.6 & 43.6 & 59.5 & \textbf{24.6} & - & - \\
MXGNet \citep{wang2020abstract}  & 67.1     &  66.7 & 74.2     &  65.4 & 68.3     &  33.6 & 67.1     &  43.3 &  63.7     &  19.9 &  69.1    &  18.9 \\
ViT   & 73.3     &  72.7 & \textbf{89.9}     &  67.7 & 69.4     &  34.1 & 67.6     &  44.1 &  73.8     &  15.9 &  \textbf{92.2}     &  16.4 \\
\midrule
NI (ours)   & \textbf{77.3}     &  \textbf{77.0} & 87.9     &  \textbf{70.5} & 69.5 & 36.6 & \textbf{68.6}     &  \textbf{45.2} &  \textbf{79.9} &  20.0 &  91.8     &  \textbf{19.4} \\
\bottomrule
\end{tabular}
\end{table*}
\vspace{-8pt}
\section{Conclusion}
\vspace{-5pt}
We have presented Neural Interpreters, a self-attention based architecture capable of learning a system of recomposable computational primitives. Neural Interpreters relax the rigidity in how computation is reused in current neural architectures, and our experiments show that the modular inductive biases it incorporates facilitate systematic generalization and sample-efficient adaptation to new tasks. 

There are multiple exciting avenues of future research. 
One line of work could leverage the capacity extension capability of Neural Interpreters in a continual learning setting, thereby enabling \textit{continuous integration} of knowledge in a neural model. 
Another promising direction entails using Neural Interpreters as a backbone for learning world models, where systematic generalization and out-of-distribution robustness are of paramount importance.

\bibliographystyle{plain}
\bibliography{neurips_2021}

\newpage
\appendix
\section{General Details about the Architecture}

\subsection{Hyperparameters and How to Set Them}

Neural Interpeters introduce a number of components that are not present in Vision Transformers, and accordingly, it introduces additional hyperparameters. While we found a large range of hyperparameters can work well in practice, there are patterns that warrant discussion. In this section, we provide a detailed discussion of these hyperparameters, and how these are set in this work. We also remark that there might be other hyperparameter settings that work well for different settings, and the insights in this section should merely function as a guide. 

\subsubsection{Partitioning Depth between LOCs, Function Iterations and Script} \label{app:general_hparams_depth}
There are three distinct ways of increasing the depth of Neural Interpreters. 

\begin{enumerate}
    \item \textbf{Increasing the number of Function Iterations.} Increasing $n_i$, the number of function iterations, is a natural way of increasing the depth of Neural Interpreters \textit{without} increasing the number of parameters. Large $n_i$ encourages the recursive and iterative reuse of computation \citep{dehghani2019universal}, but might result in a scarcity of parameters. 
    \item \textbf{Increasing the number of Scripts.} Increasing $n_s$, the number of scripts, is a way of increasing the depth of Neural Interpreters \textit{while} increasing the number of parameters. Larger $n_s$ tends to result in models that are easier to train, potentially due to a larger number of good solutions in the parameter space \citep{choromanska2015loss} (owing to the larger number of parameters). However, if $n_s$ is increased at the price of decreasing $n_i$, one might forego some inductive bias towards iterative reuse of computation. 
    \item \textbf{Increasing the number of LOCs.} Increasing $n_l$, the number of LOCs, also increases the depth \textit{while} increasing the number of parameters. However, unlike increasing $n_s$, increasing $n_l$ results in a deeper block of layers being recursively applied. Increasing $n_l$ might come at the price of decreasing $n_i$, in which case some recursive inductive bias is foregone; or it might come at the price of decreasing $n_s$, which might result in models that are less consistent. 
\end{enumerate}

\textbf{Recommendation.} If training is less stable or in-distribution performance is important, one should consider increasing the number of scripts $n_s$. If the training is stable but out-of-distribution generalization or fast-adaptation performance is important for the application, one should consider increasing the number of function iterations $n_i$. If there is additional budget for hyper-parameter search, one could consider tuning the number of LOCs (starting with $n_l \in \{1, 2\}$). 

\subsubsection{Increasing the Number of Functions} \label{app:general_hparams_numfuncs}

Increasing $n_f$, the number of functions, is a parameter efficient way of increasing the width of the network in a model-parallelizable way. This is especially apparent from Equation~\ref{eq:loc}, where the index over functions ($u$) can be effectively folded in to the batch-axis. Further, we found benefits in increasing the number of functions (also in terms of in-distribution performance), suggesting that the distribution of parameters between interpreter and the codes (as described in Equation~\ref{eq:modlin}) is scalable.

\textbf{Recommendation.} The number of functions can be safely increased to match available resource capacity. 

\subsubsection{Kernel Truncation Parameter and Dimension of Type Space}

These hyperparameters (inherited from \citep{rahaman2020s2rms}) have to do with routing of information through the network. The truncation parameter $\tau \in [0, 2)$ controls the hardness of the routing -- if $\tau$ is small, functions are only granted access to set elements whose types lie in the immediate vicinity of their signatures. For larger $\tau$, functions may be granted access to set elements whose types are less similar to their signatures in type-space, albeit the said elements are down-weighted by the kernel. The type space dimension $d_{\text{type}}$ controls the amount of flexibility afforded to the routing mechanism. Intuitively, larger $d_{\text{type}}$ implies that there are more ways to how the signature and type vectors can be positioned in the type space $\mathcal{T}$ (a hypersphere of dimension $d_{\text{type}}$) relative to each other. 

\textbf{Recommendation.} These hyperparameters may vary with the problem at hand. If sparsity is desired, one should consider lower values for $\tau$. If training is less stable, larger values of $\tau$ might mitigate the issue. We find $\tau \in [1.2, 1.7]$ to be a reasonable range for hyperparameter sweeps. As for $d_{\text{type}}$, we find all values between $20$ and $50$ to work well in our experiments. 

\subsubsection{Learning Function Signatures and Code} \label{app:general_hparams_ffs}

When pre-training the model, one decision that must be made is whether or not the function signatures and codes should be trained. Note that freezing these parameters at the pre-training stage does not necessarily constrain the model in a significant way -- if the function signatures are fixed, the type-inference MLP can adapt (Equation~\ref{eq:type_inference}); likewise, if function codes are frozen, the weight matrices $W_c$ (Equation~\ref{eq:loc}) can adapt. Note that this applies in the pre-training phase, where the type inference MLP and the interpreter parameters are allowed to adapt. 

\textbf{Recommendation.} While we did not find a large difference, runs with frozen function codes were slightly less consistent than the ones with learned function codes. At the same time, runs with frozen function signatures tended to perform at least as well as the ones that learned function signatures, if not slightly better.\footnote{This is less surprising in light of the fact that the type-inference MLP has a larger number of parameters that can be adapted during training.} 

\subsubsection{Choice of an Optimizer and Scheduler}

Like for most self-attention based models (including the transformer \citep{vaswani2017attention}), the choice of an optimizer and learning rate schedule plays an important role. A common practice is to use Adam with a linear learning rate warm-up and cosine annealing (once per optimization step). However, learning rate warm-up is known to be a heuristic to control the variance of Adam learning rate in the early stages of training, a problem that Rectified Adam \citep{liu2019variance} (RAdam) solves in a more principled way while eliminating a sensitive hyperparameter (the number of warm-up steps). Further, for certain adaptation tasks where the loss-landscape can potentially be challenging, we found Shampoo \citep{gupta2018shampoo} to work particularly well. 

\textbf{Recommendation.} For pre-training Neural Interpreters, we can recommend the RAdam optimizer with a cosine annealing schedule (without warm-up). We anneal the learning rate by $\sim$two orders of magnitude over 80-90\% of the training steps, and keep the learning rate at the minimum for the remainder of the steps. While we found RAdam to also work well for most finetuning experiments, Shampoo \citep{gupta2018shampoo} with appropriately tuned learning rate can serve as a reasonable alternative in the event that RAdam does not perform as expected. 

\subsection{Limitations and Future Work}

\textbf{Lack of Top-Down Processing.} Like most self-attention based models, Neural Interpreters process their inputs in a bottom-up manner where the input is encountered only once, which is at the first (input) layer. However, top-down processing of information is known to be a useful prior \citep{locatello2020objectcentric, goyal2019recurrent, jaegle2021perceiver}, and future work may explore incorporating this in the proposed architecture. 

\textbf{Higher Order Functions.} While Neural Interpreters are a step towards models that can flexibly compose computational primitives, they are (in their current form) missing certain notions from functional programming that could potentially serve as useful inductive biases. One of these notions is that of higher-order functions, i.e., functions that can manipulate other functions depending on the context.\footnote{In python, these are like decorators.} Support for higher-order functions can enable the model to no longer rely on a discrete set of pre-learned functions; instead, the model can learn to create new functions on the fly (i.e. at test time). 
We nevertheless note that Neural Interpreters already posses some features that can facilitate this inductive bias, the most important one being explicit representation of function codes (which can be manipulated by functions, just like other set elements). 
 
\textbf{Lazy Function Execution.} Another potentially powerful notion (borrowed from functional programming) that is yet to be incorporated is that of \textit{lazy} or \textit{deferred} function executions. When implemented, this can enable models that support functional representations, or \textit{abstractions} as they are known in the language of lambda calculus. 

\subsection{Broader Impact}
Neural Interpreters is a general neural back-bone that can be used in a variety of applications that we may not yet foresee. Nevertheless, the presented work provides an architecture that is computationally scalable, implying that a user might be tempted to experiment with more compute than is strictly necessary for obtaining good results; if not powered by nuclear or other renewable sources of energy, this might result in a larger carbon footprint.

\section{Learning Fuzzy Boolean Expressions} \label{app:fuzzy-boolean}

\subsection{Sampling Fuzzy Boolean Functions}

\begin{figure}[htp]
\centering
\begin{subfigure}{\textwidth}
  \centering
  \includegraphics[width=1\linewidth]{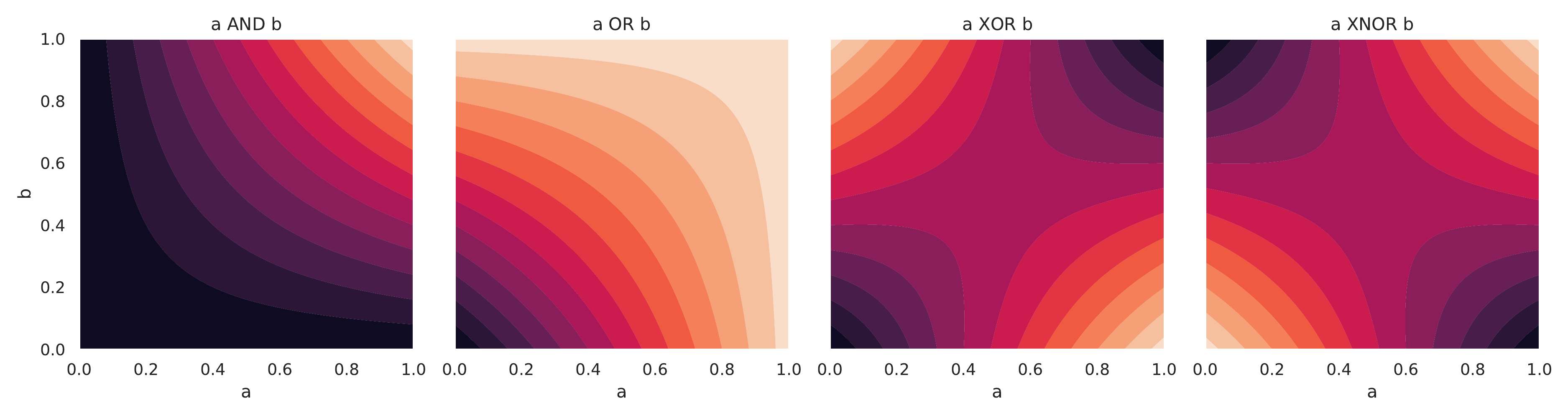}
\end{subfigure}
\caption{Visualization of fuzzy relaxations of binary operations mapping $a \in [0, 1]$ and $b \in [0, 1]$ to a value in $[0, 1]$. From left to right: \texttt{and}, \texttt{or}, \texttt{xor} and \texttt{xnor}.}
\label{fig:fuzzybool_ops}
\end{figure}

In what follows, we define a family of smooth functions mapping from the unit hyper-cube $[0, 1]^N$ to $[0, 1]$. To this end, consider again the primitives defined in Equation~\ref{eq:fuzzbool}. Where $x_i, x_j \in [0, 1]$, we define: 
\begin{align}
\texttt{and}(x_i, x_j) &= x_i x_j \label{eq:fuzzbool_and} \\ 
\texttt{not}(x_i) &= \bar{x}_i = 1 - x_i \label{eq:fuzzbool_not} \\ 
\texttt{or}(x_i, x_j) &= x_i \oplus x_j = 1 - (1 - x_i)(1 - x_j) \label{eq:fuzzbool_or}
\end{align}
Observe that if $x_i, x_j \in \{0, 1\}$, these operations reduce to their Boolean namesakes, and Equation~\ref{eq:fuzzbool_or} is consistent with de Morgan's law. In this sense, the primitives described above induce a \textit{relaxation} of Boolean logic to real numbers on the compact interval $[0, 1]$. We note that this relaxation, called product fuzzy logic, is not unique: there exist other definitions of the \texttt{and} and \texttt{not} operations that define other logics (examples being Łukasiewicz and G\"odel-Dummett logics). 

Given these primitives, it is now possible to construct functions that resemble boolean functions in the cannonical disjunctive normal form (i.e., in the sum-of-products form). As an example, consider a vector $\mathbf{x} \in [0, 1]^5$, whose components we call $a, b, c, d, e \in [0, 1]$. One may now define a function: 
\begin{align} \label{eq:fuzzbool_eg}
f(\mathbf{x}): [0, 1]^5 \to [0, 1], \; (a, b, c, d, e) \mapsto \bar{a}bcd\bar{e} \oplus a\bar{b}c\bar{d}e \oplus ab\bar{c}d\bar{e}
\end{align}
If $a, b, c, d, e$ were to be boolean (i.e., $\in \{0, 1\}$), the function $f$ would have a truth table where $f = 1$ only if $a = 0, b = 1, c = 1, d = 1, e = 0$, or $a = 1, b = 0, c = 1, d = 0, e = 1$, or $a = 1, b = 1, c = 0, d = 1, e = 0$. Conversely, given this truth table, it is possible to reconstruct $f$ in the sum-of-products form described above. 

The above fact makes randomly sampling a fuzzy boolean function like sampling from the Bernoulli distribution: for all combinations of possible values of $a, b, c, d, e \in \{0, 1\}^5$, we sample the value of a boolean function $f(a, b, c, d, e) \sim \text{Bernoulli}(0.5)$ in order to populate the truth-table of $f$. Given the randomly sampled truth table, we construct the expression for $f$ in the sum-of-product form. Finally, we interpret the boolean expression (mapping from $\{0, 1\}^5 \to \{0, 1\}$) as a fuzzy boolean expression mapping from $[0, 1]^5 \to [0, 1]$ using the corresponding primitives defined in Equation~\ref{eq:fuzzbool_and} et seq.

\subsection{Hyperparameters}
Please refer to Table~\ref{table:logic_hparams}.

\begin{table}
\centering
\caption{Hyperparameters for results in Table~\ref{tab:logic} (Learning Fuzzy Boolean Expressions).} \label{table:logic_hparams}
\begin{tabular}{l  l}
  \toprule
  \textbf{Parameters} & \textbf{Values}\\
  \midrule 
  Batch size & 128\\
  Pretraining epochs & 20 \\
  Finetuning epochs & 3 \\
  \midrule
  Dimension of code vector ($\mathbf{c}$) & 128 \\
  Dimension of intermediate features & 128 \\
  Number of scripts ($n_s$) & 2 \\
  Number of function Iterations ($n_i$) & 2 \\
  Number of LOCs ($n_l$) & 1 \\
  Number of functions ($n_f$) & 4 \\
  Number of heads per LOC & 1 \\
  Number of features per LOC head & 32 \\
  Type Inference MLP Depth & 2 \\
  Type Inference MLP Width & 128 \\
  Frozen Function Signatures & \texttt{False} \\
  Frozen Function Codes & \texttt{False} \\
  Truncation Parameter ($\tau$) & 1.6 \\
  Type Space Dimension ($d_{\text{type}}$) & 24 \\
  \midrule
  Optimizer & RAdam \citep{liu2019variance} \\
  Adam: learning rate (pre-training) & 0.006 \\
  Adam: learning rate (finetuning) & 0.05 \\
  Adam: $\beta_1$ & 0.9\\
  Adam: $\beta_2$ & 0.999\\
  Adam: $\epsilon$ & 1e-8\\
  Learning rate scheduler & None \\
  \bottomrule
\end{tabular}
\end{table}%

\section{Multi-Task Image Classification} \label{app:digits}
\subsection{The Digits Dataset}

\begin{figure}[htp]
\centering
\begin{subfigure}{\textwidth}
  \centering
  \includegraphics[width=1\linewidth]{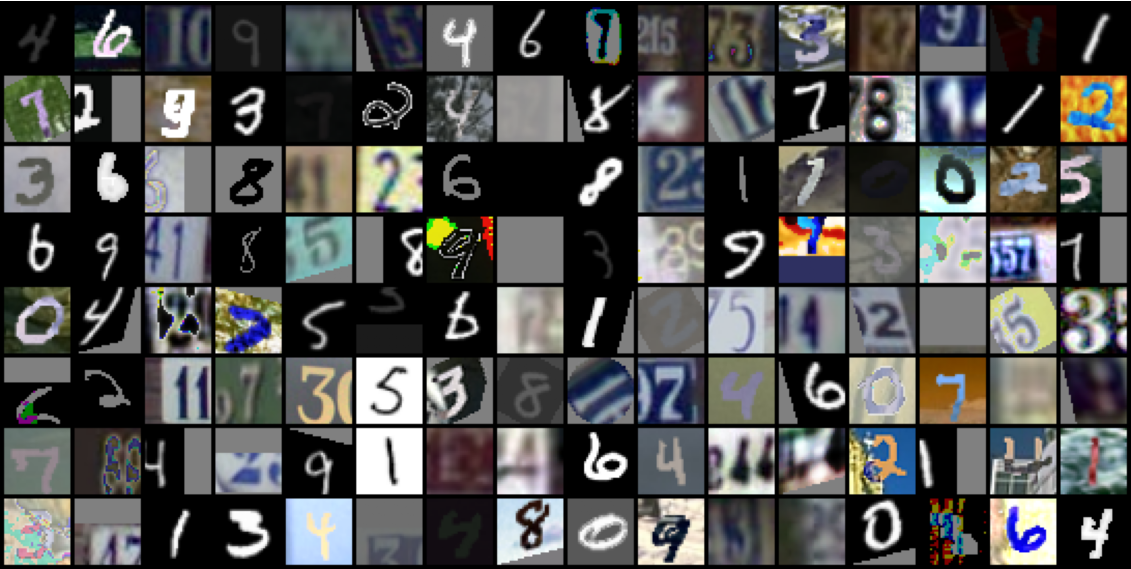}
\end{subfigure}
\caption{Augmented Samples from the Digits Dataset.}
\label{fig:digits_samples}
\end{figure}

The Digits dataset is a concatenation of three datasets of labelled images of digits: SVHN \citep{netzer2011reading}, MNISTM \citep{ganin2016domain}, and MNIST \citep{lecun2010mnist}. All images are up-sampled to RGB images of size $32 \times 32$, and the combined training set has 193257 samples, whereas the validation set has 46032 samples. In addition to the images and labels, we also preserve information about which of the constituent datasets a sampled image originates from. 

We use RandAugment \citep{cubuk2019randaugment} to augment the input images before feeding them to the model, and use the implementation from Pytorch Image Models \citep{rw2019timm}. Figure~\ref{fig:digits_samples} shows augmented samples from the dataset. 

\subsection{Hyperparameters}

\subsubsection{Pre-training}
Table~\ref{table:digits_ni_hparams} shows the hyperparameters used for pre-training the Neural Interpreter model considered in Figure~\ref{fig:digits}. Table~\ref{table:digits_vit_hparams} shows the same, but for the Vision Transformer model. 

\begin{table}
\centering
\caption{Hyperparameters for pre-training the Neural Interpreter Model used in Figure~\ref{fig:digits}.} \label{table:digits_ni_hparams}
\begin{tabular}{l  l}
  \toprule
  \textbf{Parameters} & \textbf{Values}\\
  \midrule 
  Batch size & 128\\
  Pre-training epochs & 100 \\
  \midrule
  Dimension of code vector ($\mathbf{c}$) & 192 \\
  Dimension of intermediate features & 192 \\
  Number of scripts ($n_s$) & 1 \\
  Number of function iterations ($n_i$) & 8 \\
  Number of LOCs ($n_l$) & 1 \\
  Number of functions ($n_f$) & 5 \\
  Number of heads per LOC & 4 \\
  Number of features per LOC head & 128 \\
  Type Inference MLP Depth & 2 \\
  Type Inference MLP Width & 192 \\
  Frozen Function Signatures & \texttt{True} \\
  Frozen Function Codes & \texttt{False} \\
  Truncation Parameter ($\tau$) & 1.4 \\
  Type Space Dimension ($d_{\text{type}}$) & 24 \\
  \midrule
  Optimizer & RAdam \citep{liu2019variance} \\
  Adam: $\beta_1$ & 0.9\\
  Adam: $\beta_2$ & 0.999\\
  Adam: $\epsilon$ & 1e-8\\
  Learning rate scheduler & Cosine (no warm-up) \\
  Scheduler: $\eta_{\text{max}}$ (Max LR) & 0.0008 \\
  Scheduler: $\eta_{\text{min}}$ (Min LR) & 0.000001 \\
  Scheduler: Number of decay steps & 120000 \\
  \midrule
  Number of parameters & $6.43 \times 10^{5}$ \\
  \midrule
  Accuracy on SVHN & 96.2 \% \\
  Accuracy on MNISTM & 98.4 \% \\
  Accuracy on MNIST & 99.4 \% \\
  \bottomrule
\end{tabular}
\end{table}

\begin{table}
\centering
\caption{Hyperparameters for pre-training the ViT Model used in Figure~\ref{fig:digits}.} \label{table:digits_vit_hparams}
\begin{tabular}{l  l}
  \toprule
  \textbf{Parameters} & \textbf{Values}\\
  \midrule 
  Batch size & 128\\
  Pretraining epochs & 100 \\
  \midrule
  Dimension of intermediate features & 192 \\
  Number of MLP Features & 192 \\
  Depth & 8 \\
  Number of heads & 3 \\
  Number of features per head & 64 \\
  \midrule
  Optimizer & RAdam \citep{liu2019variance} \\
  Adam: $\beta_1$ & 0.9\\
  Adam: $\beta_2$ & 0.999\\
  Adam: $\epsilon$ & 1e-8\\
  Learning rate scheduler & Cosine (no warm-up) \\
  Scheduler: $\eta_{\text{max}}$ (Max LR) & 0.0008 \\
  Scheduler: $\eta_{\text{min}}$ (Min LR) & 0.000001 \\
  Scheduler: Number of decay steps & 120000 \\
  \midrule
  Number of parameters & $1.80 \times 10^{6}$ \\
  \midrule
  Accuracy on SVHN & 96.3 \% \\
  Accuracy on MNISTM & 98.3 \% \\
  Accuracy on MNIST & 99.6 \% \\
  \bottomrule
\end{tabular}
\end{table}

\subsubsection{Finetuning}

Both models shown in Figure~\ref{fig:digits} (top) were fine-tuned for 10 epochs with varying number of samples. We used the same batch-size as in pre-training (128). The error bands are with respect to 6 random seeds, where the random seed also determines the subset of K-MNIST that was used. We used RAdam optimizer with a constant learning rate, which was found with a grid search (0.03 for ViT and 0.05 for NI). 

For the results shown in Figure~\ref{fig:digits} (bottom), the function codes and signatures were trained for 10 epochs on 8192 samples with Shampoo \citep{gupta2018shampoo}. We again used 6 random seeds, and for each set of trainable parameters, we grid-searched the learning rate.  We did not see good performance with RAdam in this particular setting, suggesting that the loss landscape might necessitate the pre-conditioning that is present in Shampoo (but not in RAdam).

\subsubsection{Positional Encoding}

We use a variant of the relative positional encoding scheme presented in \citep{cordonnier2019relationship}, which we now describe. Consider an array $\mathbf{X}$ of shape $C \times H \times W$, where $C$ is the number of channels, and $H$ and $W$ can be interpreted as height and width. Note that the array $\mathbf{X}$ need not be an image; it could (for instance) be a collection of embedding vectors of patches, i.e., $\mathbf{X}_{\cdot, ij}$ could be the embedding vector (of dimension $C$) of the patch that is $i$-th from top and $j$-th from left. 

We now denote with $\mathbf{e}_{\text{row}}[i_2 - i_1]$ a vector that is a learned embedding of the difference of row-indices $i_2$ and $i_1$. Likewise, we let $\mathbf{e}_{\text{col}}[j_2 - j_1]$ be a learned embedding of the difference of column-indices $j_2$ and $j_1$. Where $h$ indexes attention heads, we define: 
\begin{align}
e_{\text{row}}^h[i_2 - i_1] &= \mathbf{w}^h_{\text{row}} \cdot \mathbf{e}_{\text{row}}[i_2 - i_1] \\
e_{\text{col}}^h[j_2 - j_1] &= \mathbf{w}^h_{\text{col}} \cdot \mathbf{e}_{\text{col}}[j_2 - j_1]
\end{align}
Here, $\mathbf{w}^h_{\text{row}}$ and $\mathbf{w}^h_{\text{row}}$ are learned weight vectors, and $e_{\text{row}}^h[i_2 - i_1]$ and $e_{\text{col}}^h[i_2 - i_1]$ are scalars, one per attention head. This set-up allows each attention head to develop a positional bias independently from other heads, a feature we inherit from \citep{cordonnier2019relationship}. In the context of Neural Interpreters, we additionally allow functions to have their own positional bias, conditioned on its code $\mathbf{c}_u$. We have: 
\begin{align}
p_{\text{row}}^{uh}[i_2 - i_1] &= \text{ModLin}_{\text{row}}^h(\mathbf{e}_{\text{row}}[i_2 - i_1]; \mathbf{c}_u) \\
p_{\text{col}}^{uh}[j_2 - j_1] &= \text{ModLin}_{\text{col}}^h(\mathbf{e}_{\text{col}}[j_2 - j_1]; \mathbf{c}_u)
\end{align}
Here, $p_{\text{row}}^{uh}[i_2 - i_1]$ and $p_{\text{col}}^{uh}[j_2 - j_1]$ are scalars specific to function $u$ and attention head $h$. Finally, the overall positional bias is given as following, where broadcasting operations are implied: 
\begin{align}
b^{uh}[i_2 - i_1, j_2 - j_1] = (p_{\text{row}}^{uh}[i_2 - i_1] + e_{\text{row}}^h[i_2 - i_1]) + (p_{\text{col}}^{uh}[j_2 - j_1] + e_{\text{col}}^h[j_2 - j_1])
\end{align}
Here, $b^{uh}[i_2 - i_1, j_2 - j_1]$ is the positional bias that is added to the pre-softmax dot-product attention weights coupling the embedding vectors $\mathbf{X}_{\cdot, i_1 j_1}$ and $\mathbf{X}_{\cdot, i_2 j_2}$ at function $f_u$ and attention head at index $h$. We remark that this scheme only differs from \citep{cordonnier2019relationship} in that we allow each function to develop its own positional bias. 

\subsection{Additional Results and Ablations}

In order to understand the effect of various hyperparameters, we analyze the results of a random sweep over 100 runs on the Digits dataset. The distributions over sweep parameters are presented in Table~\ref{table:digits_sweep}. 

\begin{table}[htp]
\centering
\caption{Distribution over hyperparameters used in the sweep. $\mathcal{U}$ denotes the uniform distribution.} \label{table:digits_sweep}
\begin{tabular}{l  l}
  \toprule
  \textbf{Parameters} & \textbf{Distribution}\\
  \midrule 
  Truncation parameter ($\tau$) & $\mathcal{U}([0.7, 1.7])$ \\
  Dimension of type space ($d_{\text{type}}$) & $\mathcal{U}(\{4 * i \; | \; i \in \{2, 3, ..., 12\} \})$ \\
  Number of functions ($n_f$) & $\mathcal{U}(\{1, 2, 3, 4, 5\})$ \\
  Num. of scripts, function iterations, and LOCs $(n_s, n_i, n_l)$ & $\mathcal{U}(\{(2, 2, 2), (2, 4, 1), (4, 2, 1), (1, 8, 1)\})$ \\
  Frozen function signatures & $\mathcal{U}(\{\texttt{True}, \texttt{False}\})$ \\
  Frozen function codes & $\mathcal{U}(\{\texttt{True}, \texttt{False}\})$ \\
  Frozen patch embeddings \citep{chen2021empirical, touvron2021going} & $\mathcal{U}(\{\texttt{True}, \texttt{False}\})$ \\
  \bottomrule
\end{tabular}
\end{table}

\begin{figure}[htp]
  \begin{subfigure}[b]{0.33\textwidth}
    \centering
    \includegraphics[width=\textwidth]{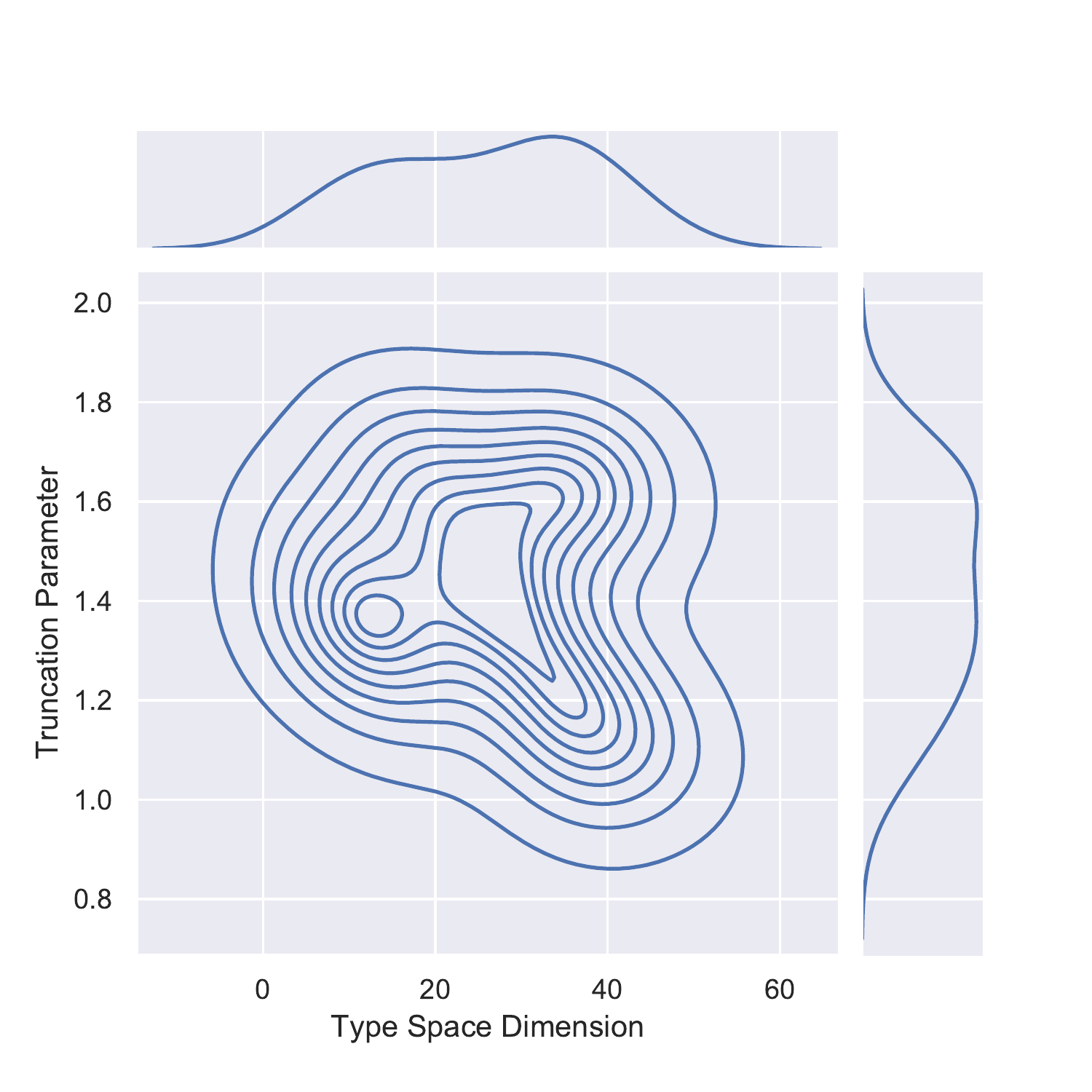}
    \caption[]%
    {SVHN}
  \end{subfigure}%
  \hfill
  \begin{subfigure}[b]{0.33\textwidth}
    \centering
    \includegraphics[width=\textwidth]{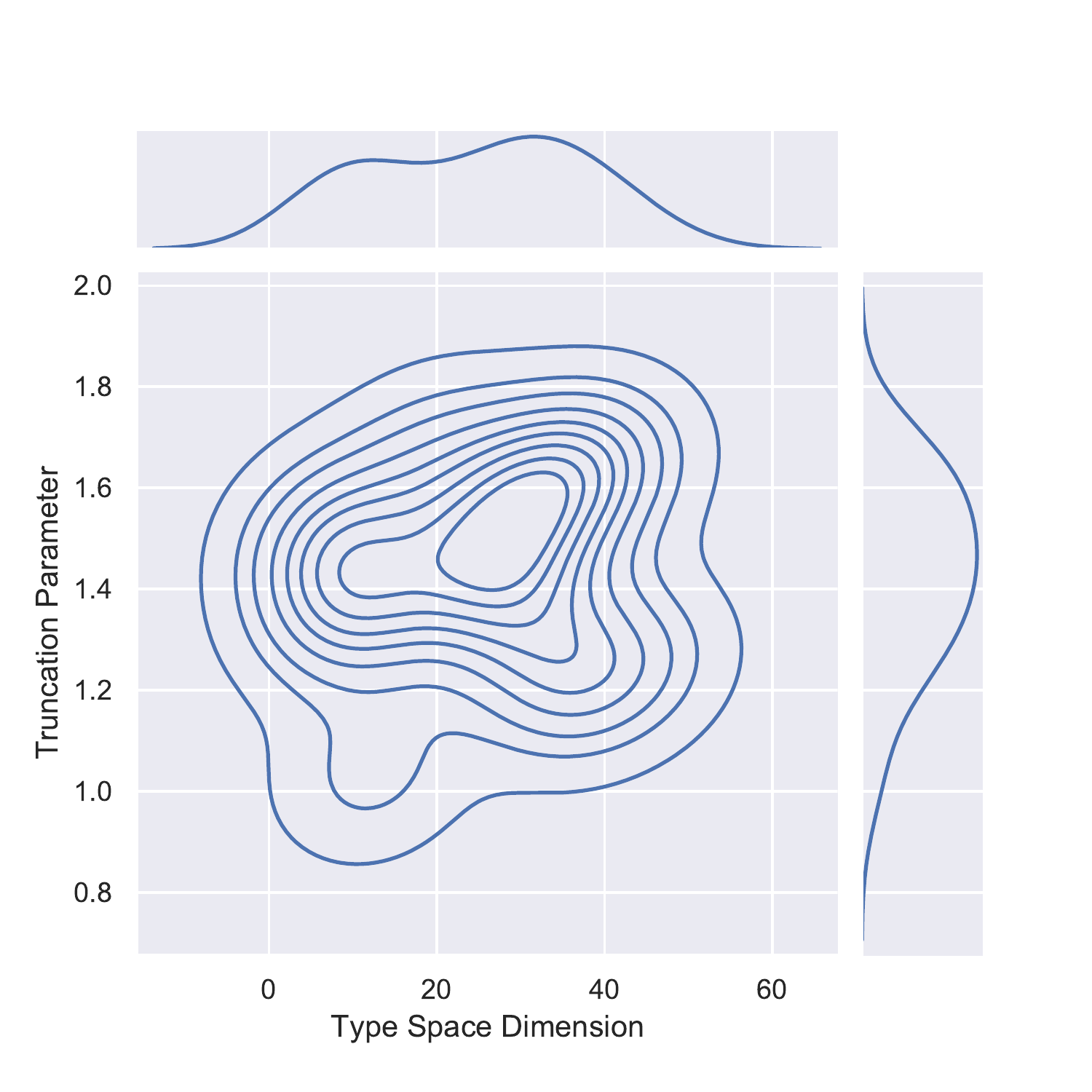}
    \caption[]%
    {MNIST-M}
  \end{subfigure}
   \hfill
  \begin{subfigure}[b]{0.33\textwidth}
    \centering
    \includegraphics[width=\textwidth]{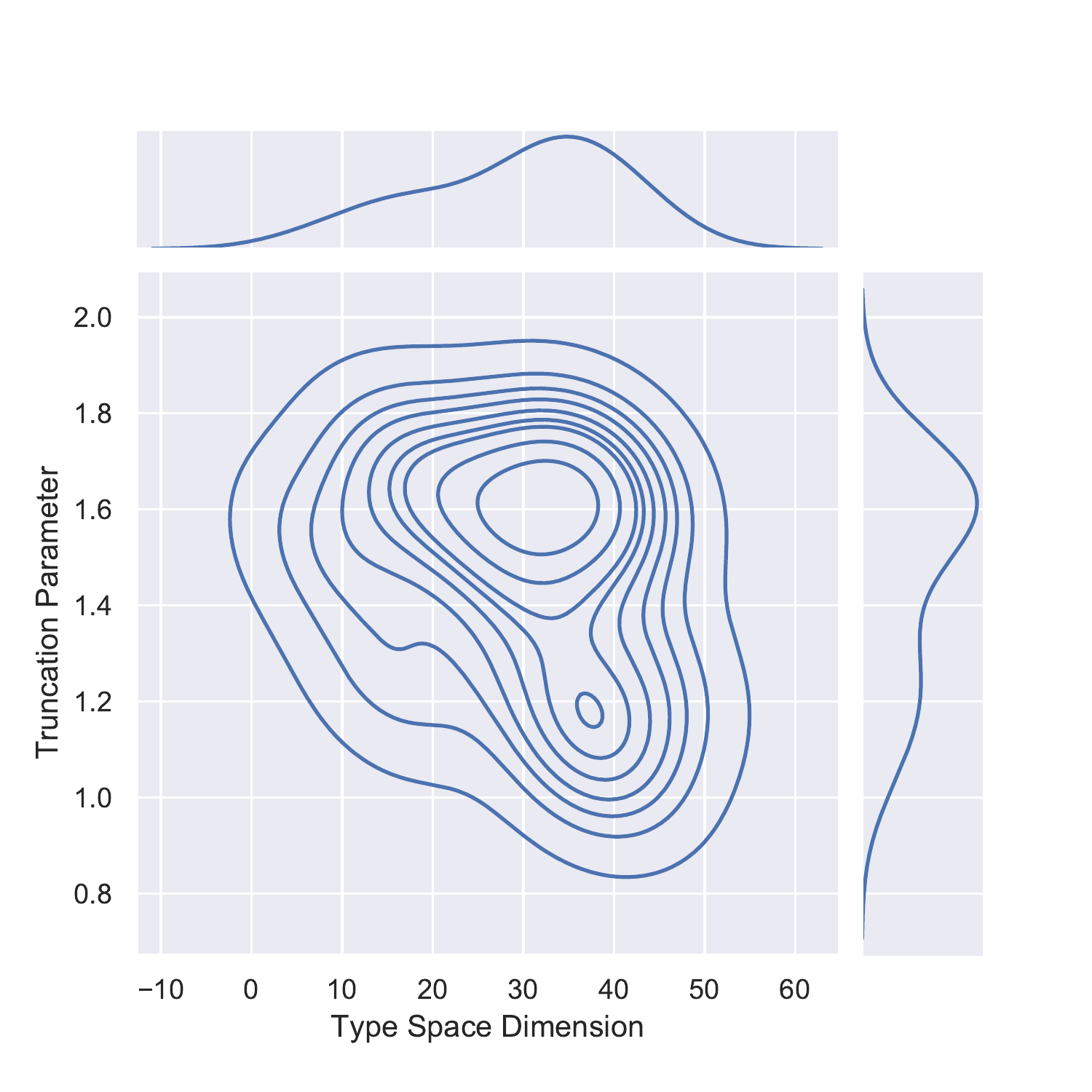}
    \caption[]%
    {MNIST}
  \end{subfigure}
  \caption[ Global caption ]
  {Kernel Density Estimates of truncation parameters $\tau$ and type space dimensions $d_{\text{type}}$ of the top 10\% of runs for each dataset.}
  \label{fig:digits_ablation_tsd_vs_ktp}
\end{figure}

\textbf{Kernel Truncation and Dimension of Type Space.} In Figure~\ref{fig:digits_ablation_tsd_vs_ktp}, we select for each dataset the top 10\% of all runs (w.r.t. validation performance), and plot a Kernel Density Estimate of their type space dimensions ($d_{\text{type}}$) and truncation parameters ($\tau$). We find that while the optimal $\tau$ and $d_{\text{type}}$ only somewhat depend on each other, there are minor variations between the SVHN and MNIST-M vs. MNIST. We speculate that this is due to SVHN and MNIST-M having cluttered backgrounds; the flexibility afforded by a larger type space is less desirable when the model must learn to suppress background clutter by routing noisy patches through similar functions. 

\begin{figure}[htp]
  \begin{subfigure}[b]{0.33\textwidth}
    \centering
    \includegraphics[width=\textwidth]{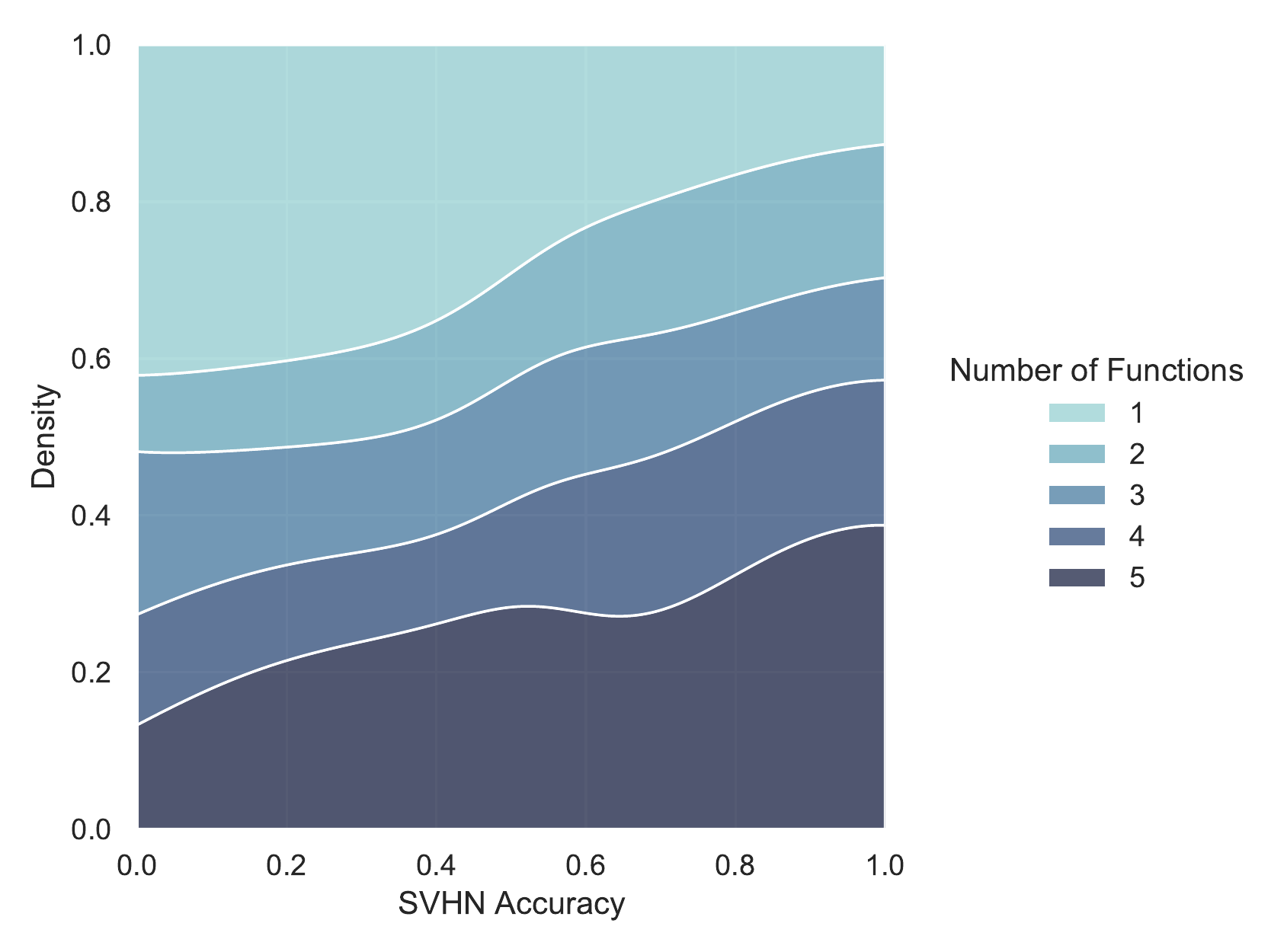}
    \caption[]%
    {SVHN}
  \end{subfigure}%
  \hfill
  \begin{subfigure}[b]{0.33\textwidth}
    \centering
    \includegraphics[width=\textwidth]{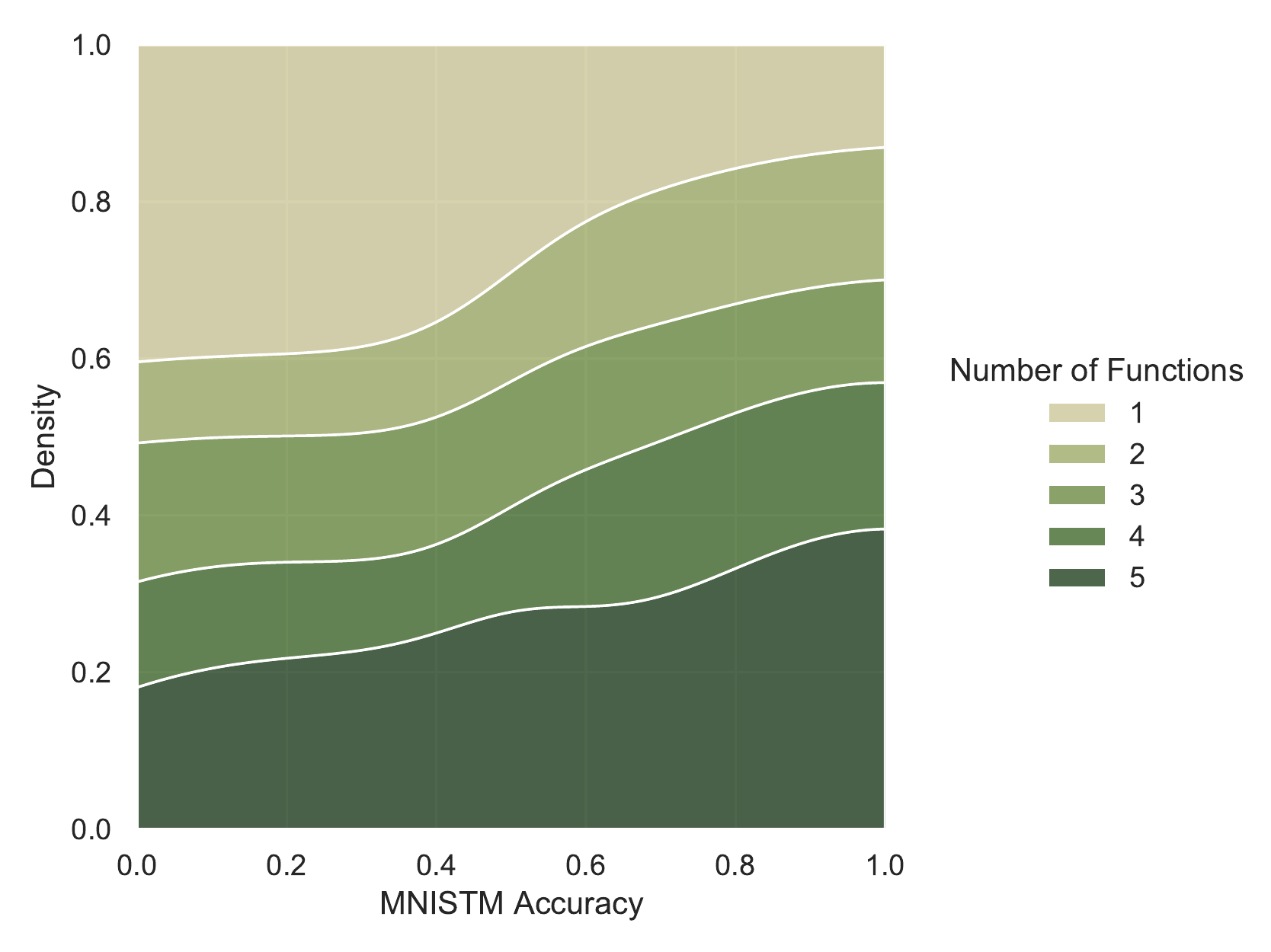}
    \caption[]%
    {MNIST-M}
  \end{subfigure}
   \hfill
  \begin{subfigure}[b]{0.33\textwidth}
    \centering
    \includegraphics[width=\textwidth]{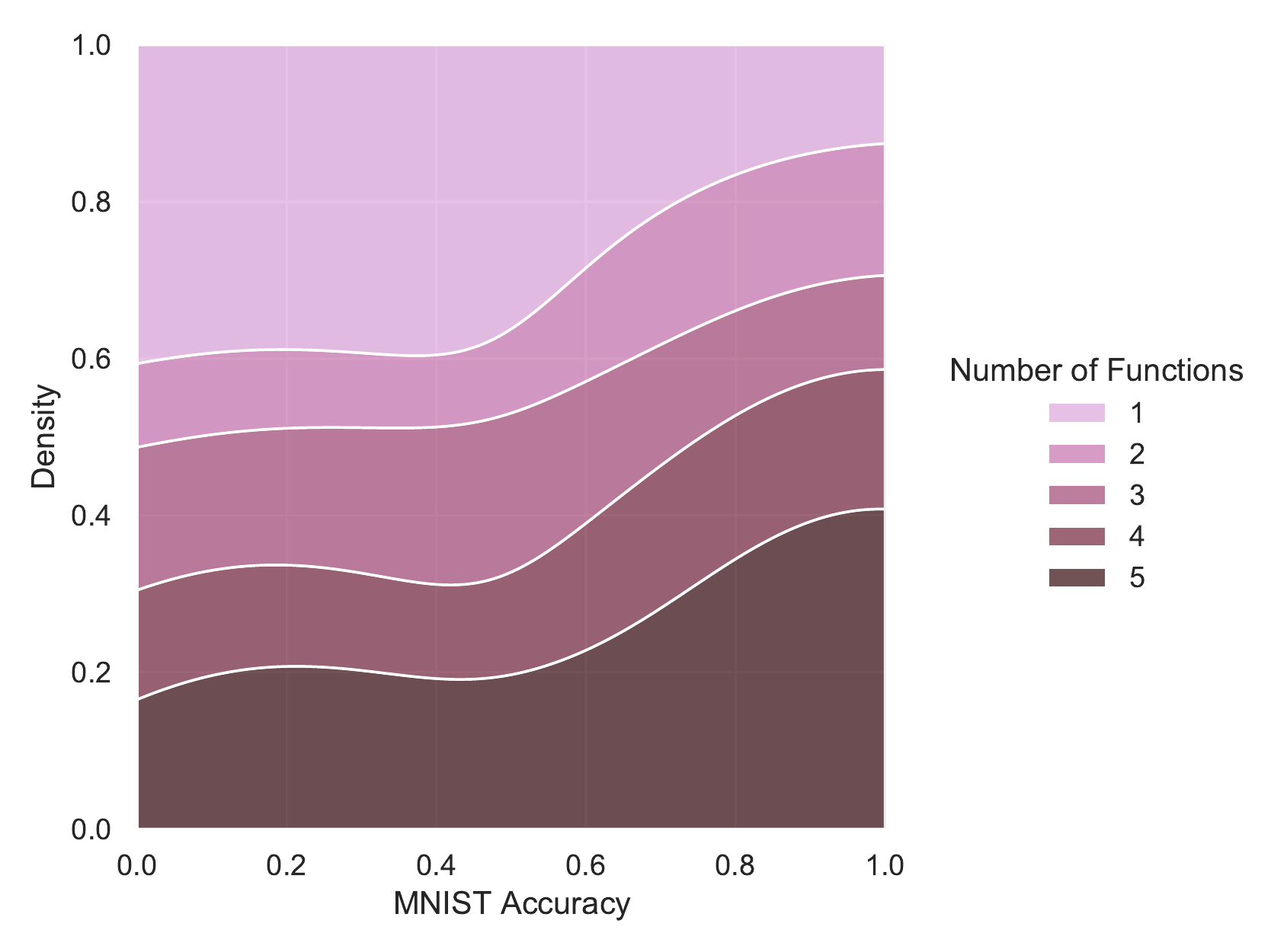}
    \caption[]%
    {MNIST}
  \end{subfigure}
  \caption[ Global caption ]
  {Conditional Kernel Density Estimates of validation performance, conditioned on the number of functions $n_f$.}
  \label{fig:digits_ablation_nf}
\end{figure}

\textbf{Number of Functions.} Figure~\ref{fig:digits_ablation_nf} shows the kernel density estimate of the validation performance of \textit{all} 100 runs, conditioned on the number of functions. We read that on the one hand, runs with 5 functions perform consistently well; on the other hand, there exist runs with a single function that perform well, but most of these runs fail. This is consistent with the recommendation in Section~\ref{app:general_hparams_numfuncs}.

\begin{figure}[htp]
  \begin{subfigure}[b]{0.33\textwidth}
    \centering
    \includegraphics[width=\textwidth]{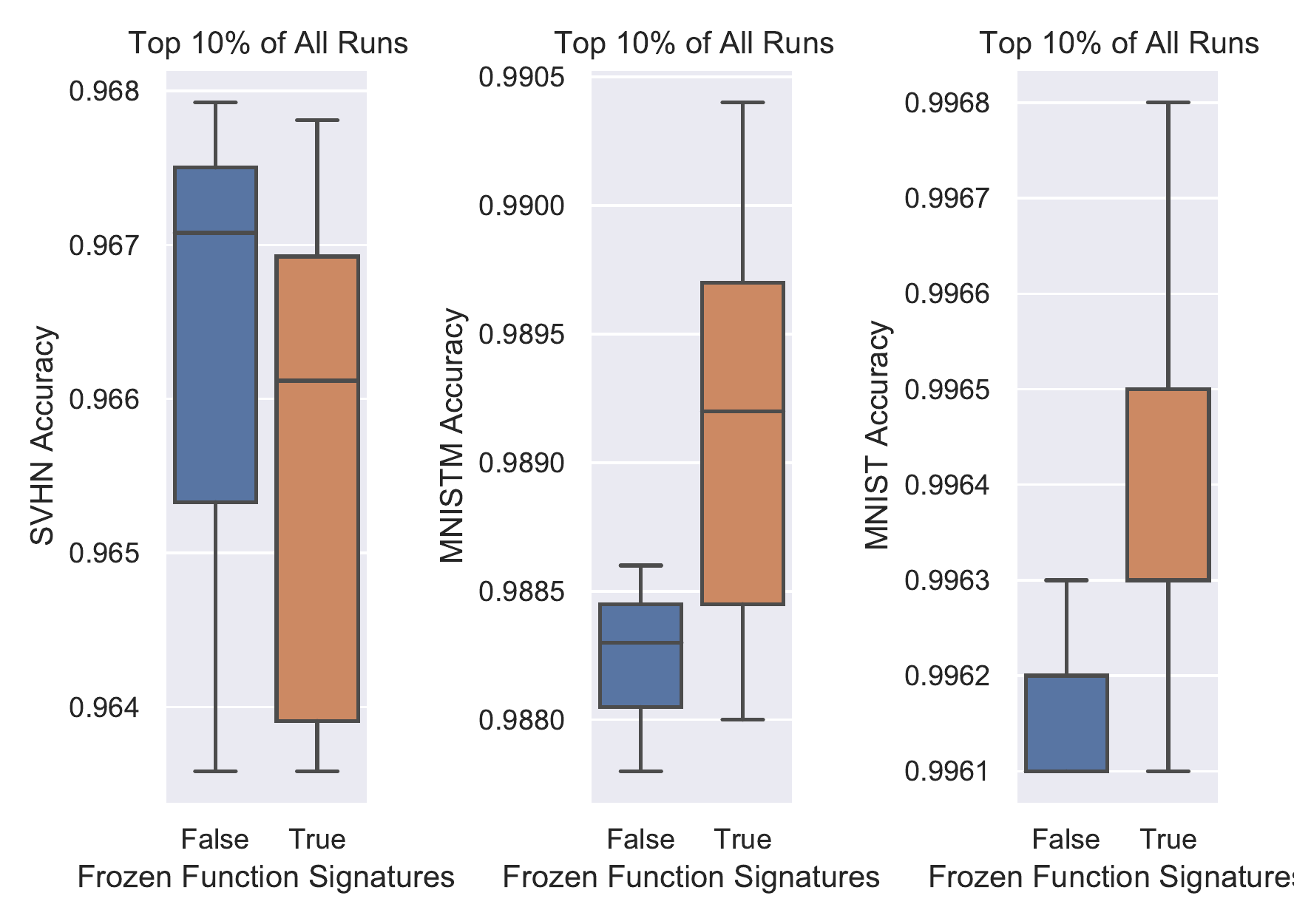}
    \caption[]%
    {Frozen Function Signatures}
  \end{subfigure}%
  \hfill
  \begin{subfigure}[b]{0.33\textwidth}
    \centering
    \includegraphics[width=\textwidth]{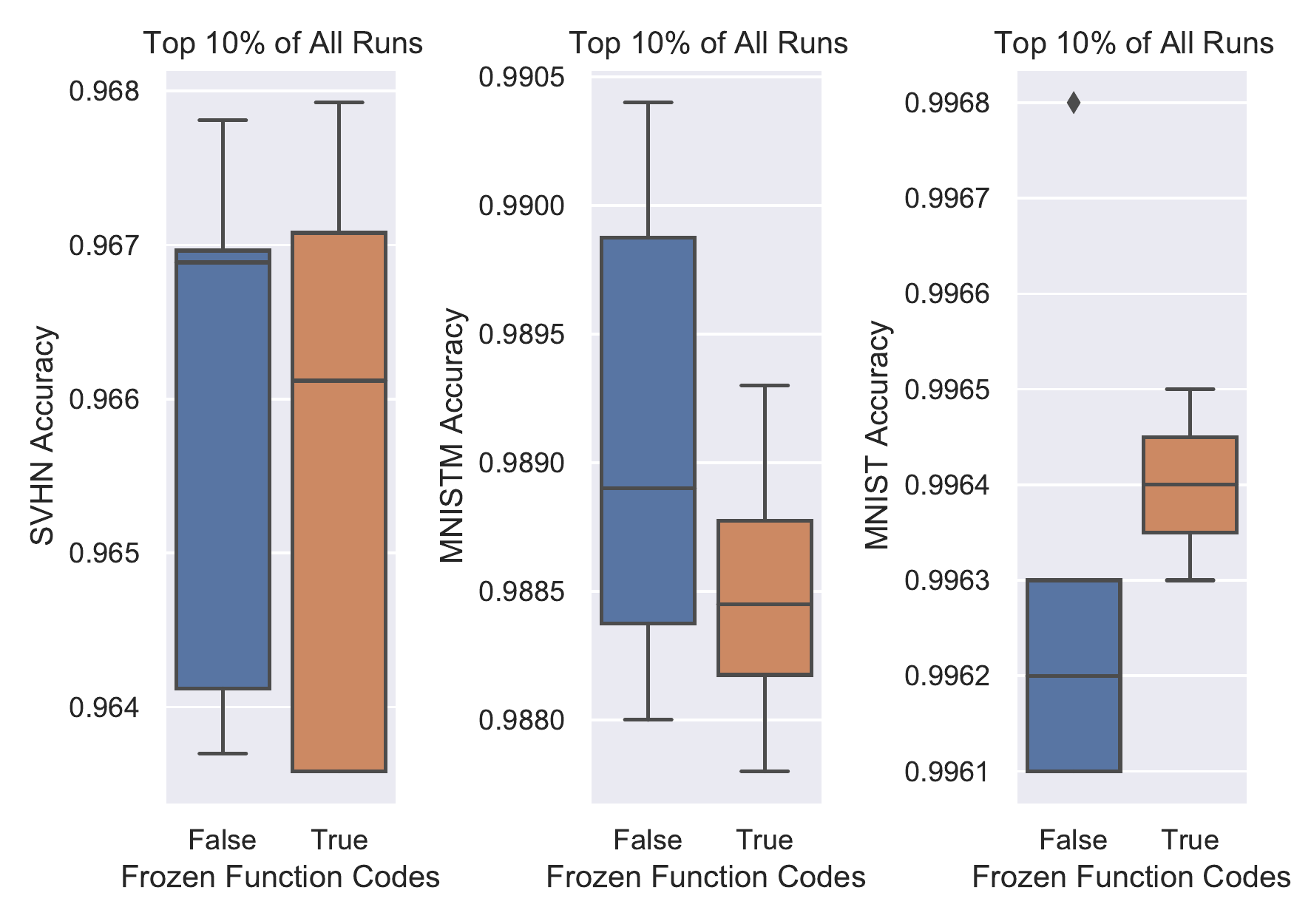}
    \caption[]%
    {Frozen Function Codes}
  \end{subfigure}
   \hfill
  \begin{subfigure}[b]{0.33\textwidth}
    \centering
    \includegraphics[width=\textwidth]{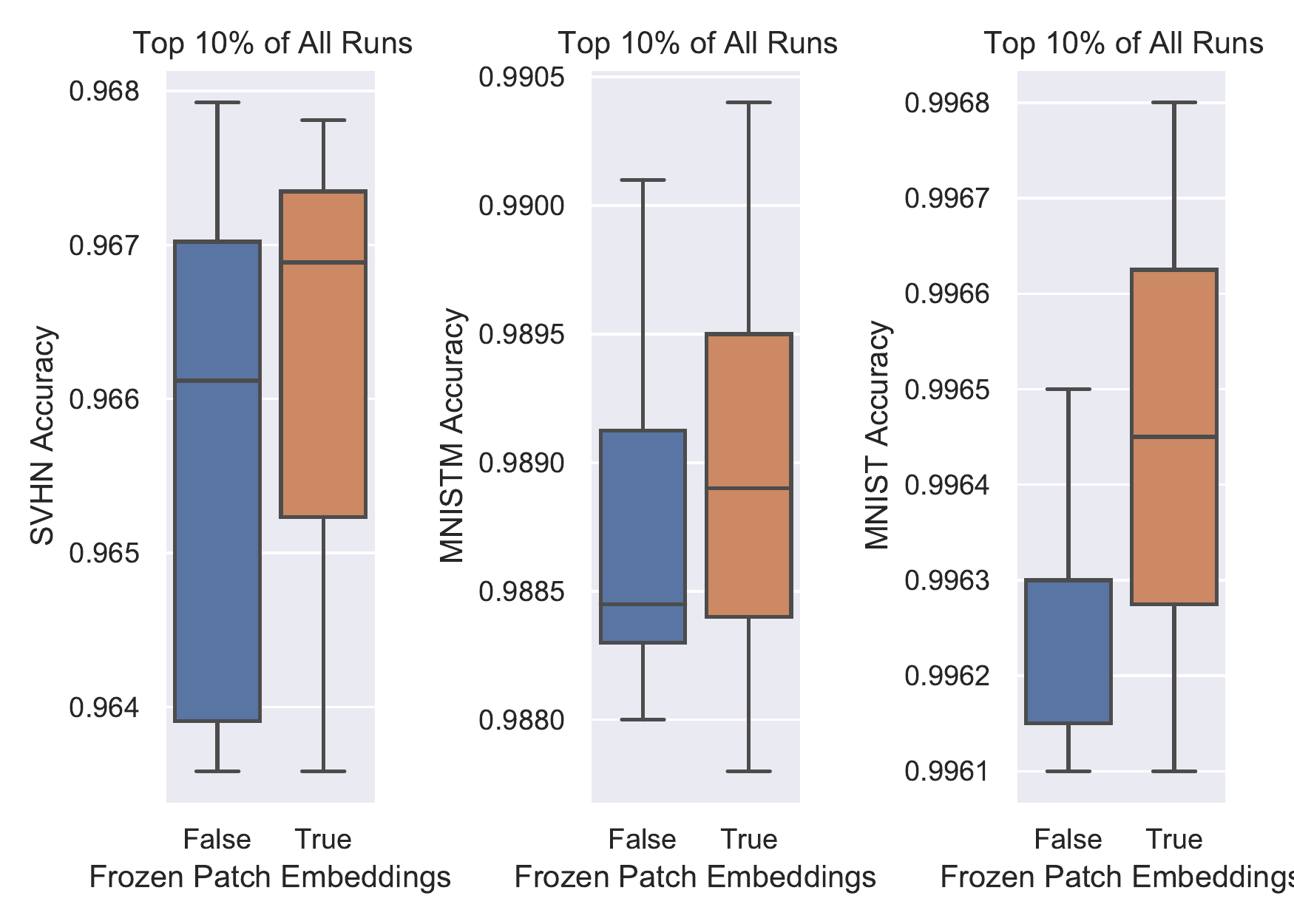}
    \caption[]%
    {Frozen Patch Embeddings}
  \end{subfigure}
  \caption[ Global caption ]
  {Box plots of validation performance of top 10\% of all runs.}
  \label{fig:digits_ablation_ffs}
\end{figure}

\textbf{Frozen Patch Embeddings, Function Signatures and Codes.} Figure~\ref{fig:digits_ablation_ffs} shows the validation performance of top 10\% of runs (for the respective dataset), with or without frozen patch embeddings, function signatures and codes. As elaborated in Section~\ref{app:general_hparams_ffs}, it is not surprising that Neural Interpreters can work well even when function codes and signatures remain frozen during training. We find that freezing function signatures can be marginally beneficial, but freezing function codes less so. We also experimented with freezing the patch embeddings, as recommended in \citep{touvron2021going, chen2021empirical}, and find that it slightly improves performance.

\textbf{Number of Scripts, Function Iterations and LOCs.} Figure~\ref{fig:digits_ablation_ns} shows the conditional kernel density estimates of validation performance, conditioned on the number of scripts. We again find that all evaluated configurations can work well. A larger number of scripts can stabilize training and lead to consistent in-distribution performance, as expected from Section~\ref{app:general_hparams_depth}. 
\begin{figure}[htp]
  \begin{subfigure}[b]{0.33\textwidth}
    \centering
    \includegraphics[width=\textwidth]{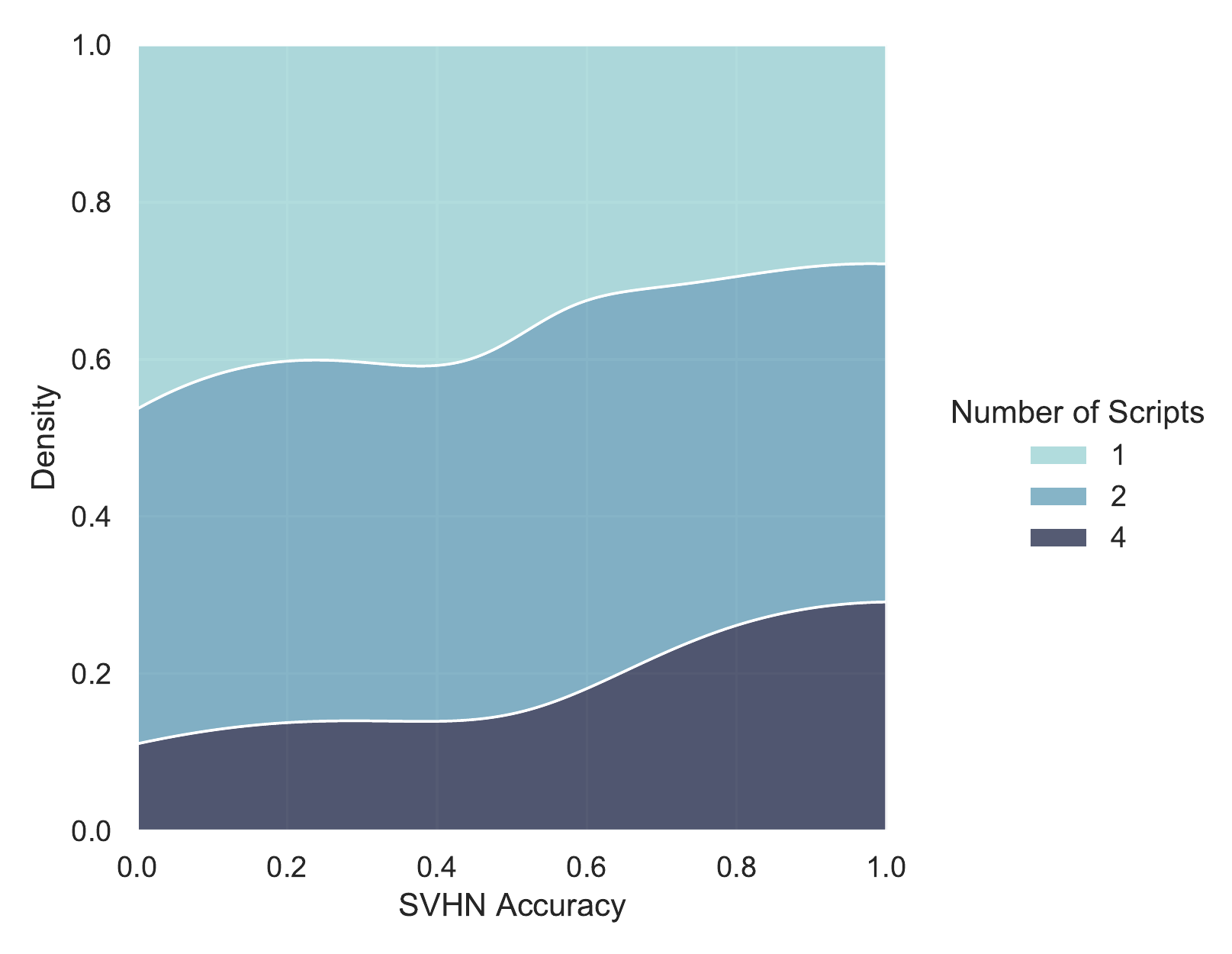}
    \caption[]%
    {SVHN}
  \end{subfigure}%
  \hfill
  \begin{subfigure}[b]{0.33\textwidth}
    \centering
    \includegraphics[width=\textwidth]{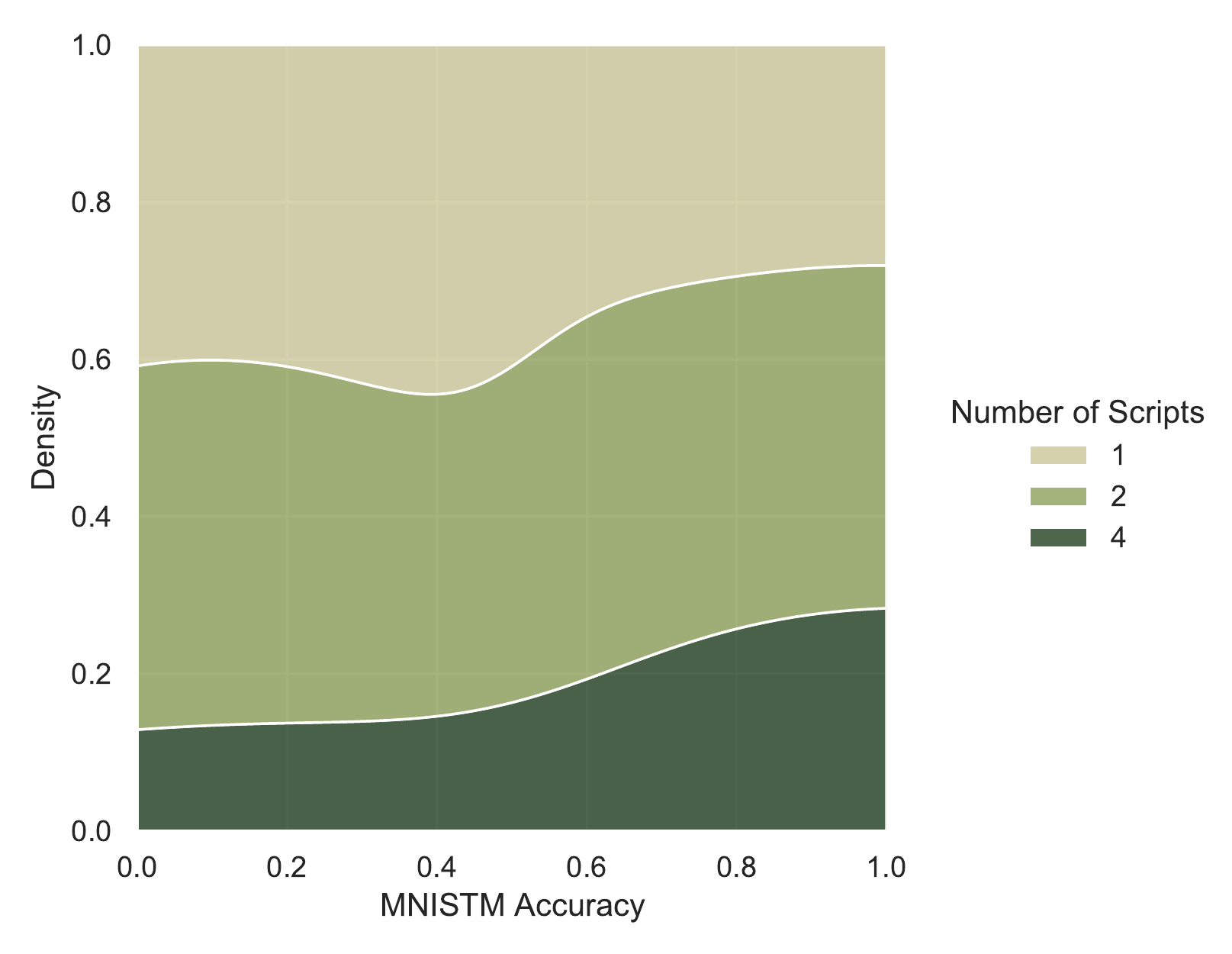}
    \caption[]%
    {MNIST-M}
  \end{subfigure}
   \hfill
  \begin{subfigure}[b]{0.33\textwidth}
    \centering
    \includegraphics[width=\textwidth]{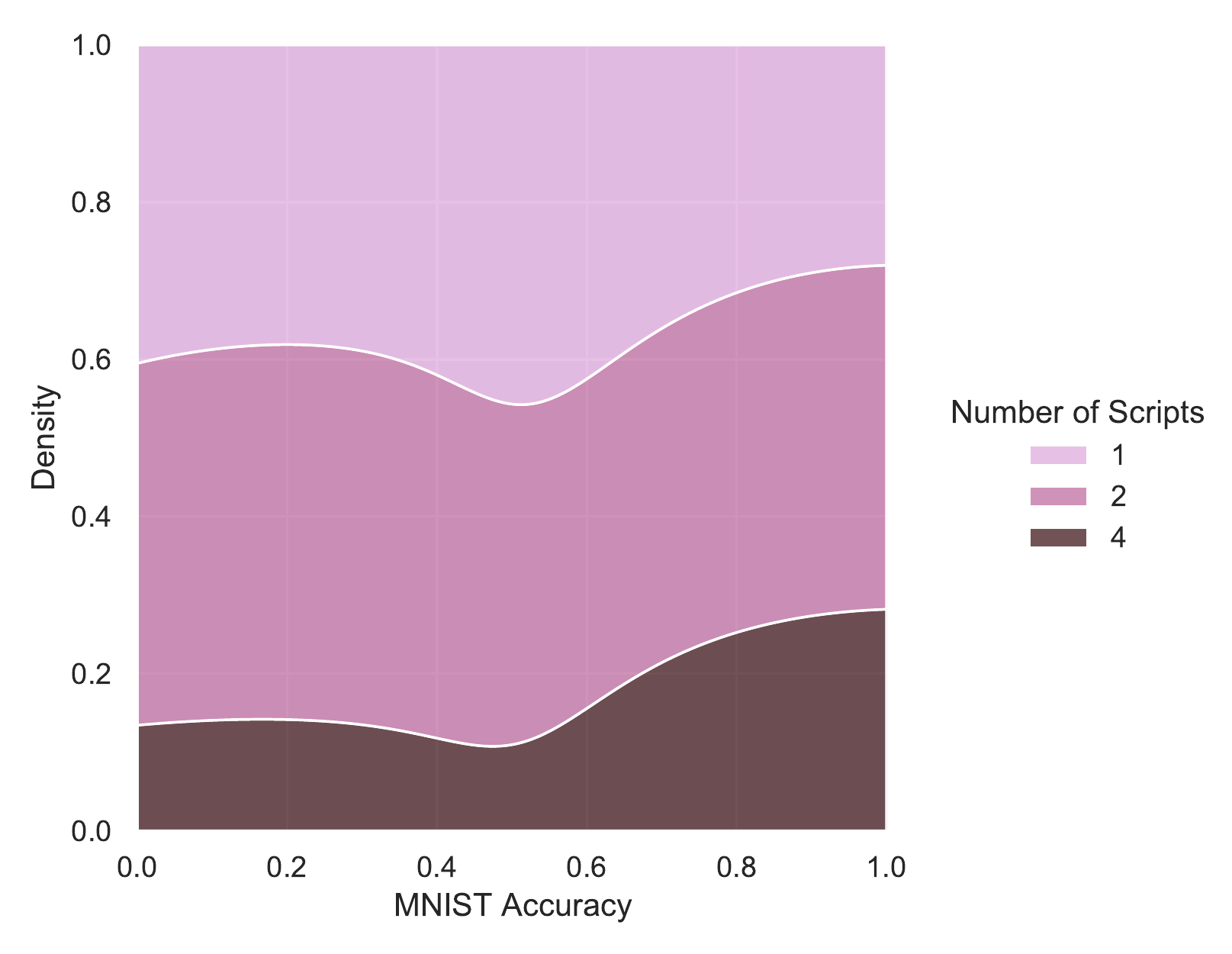}
    \caption[]%
    {MNIST}
  \end{subfigure}
  \caption[ Global caption ]
  {Conditional Kernel Density Estimates of validation performance, conditioned on the number of scripts $n_s$.}
  \label{fig:digits_ablation_ns}
\end{figure}
\begin{figure}[htp]
  \begin{subfigure}[b]{0.33\textwidth}
    \centering
    \includegraphics[width=\textwidth]{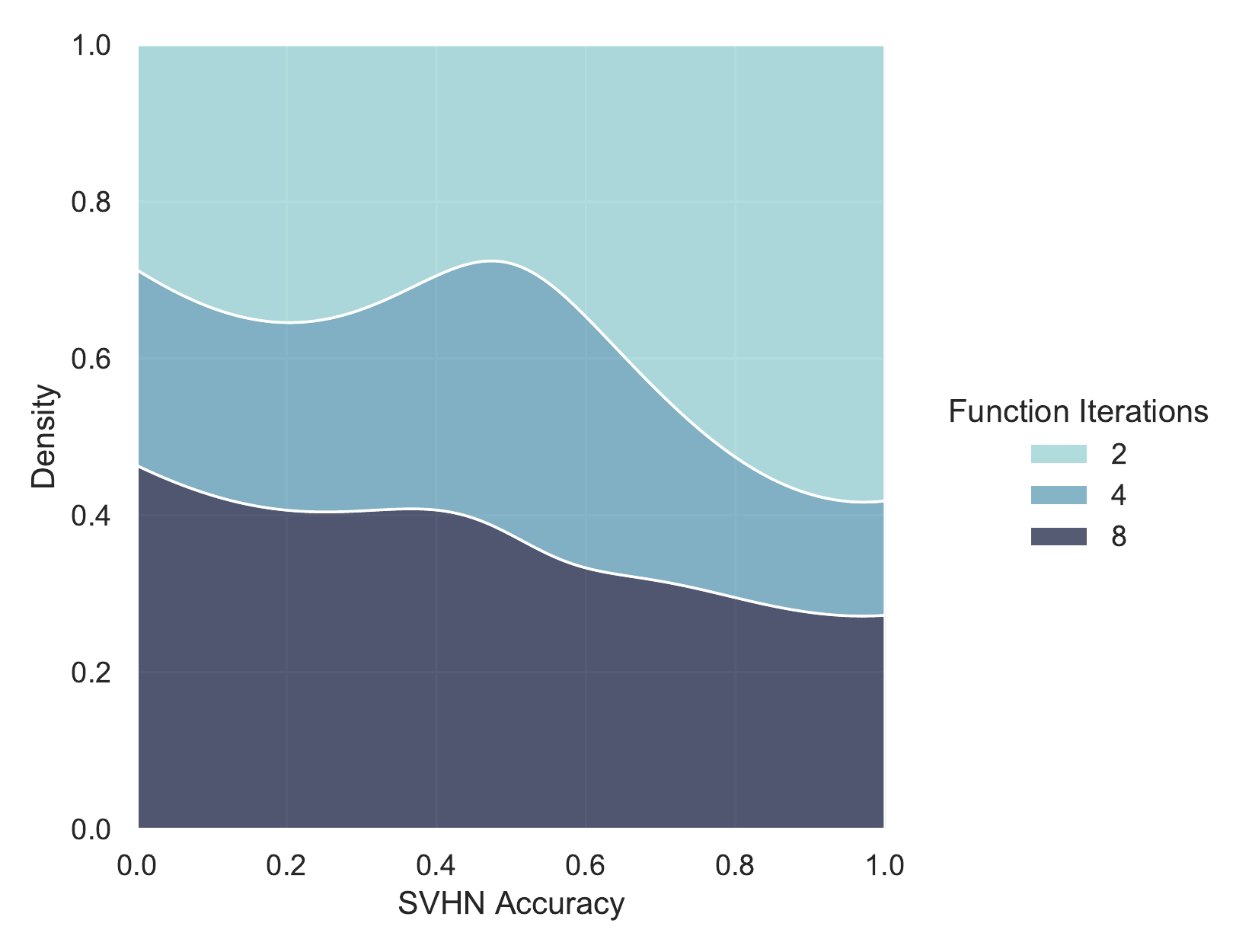}
    \caption[]%
    {SVHN}
  \end{subfigure}%
  \hfill
  \begin{subfigure}[b]{0.33\textwidth}
    \centering
    \includegraphics[width=\textwidth]{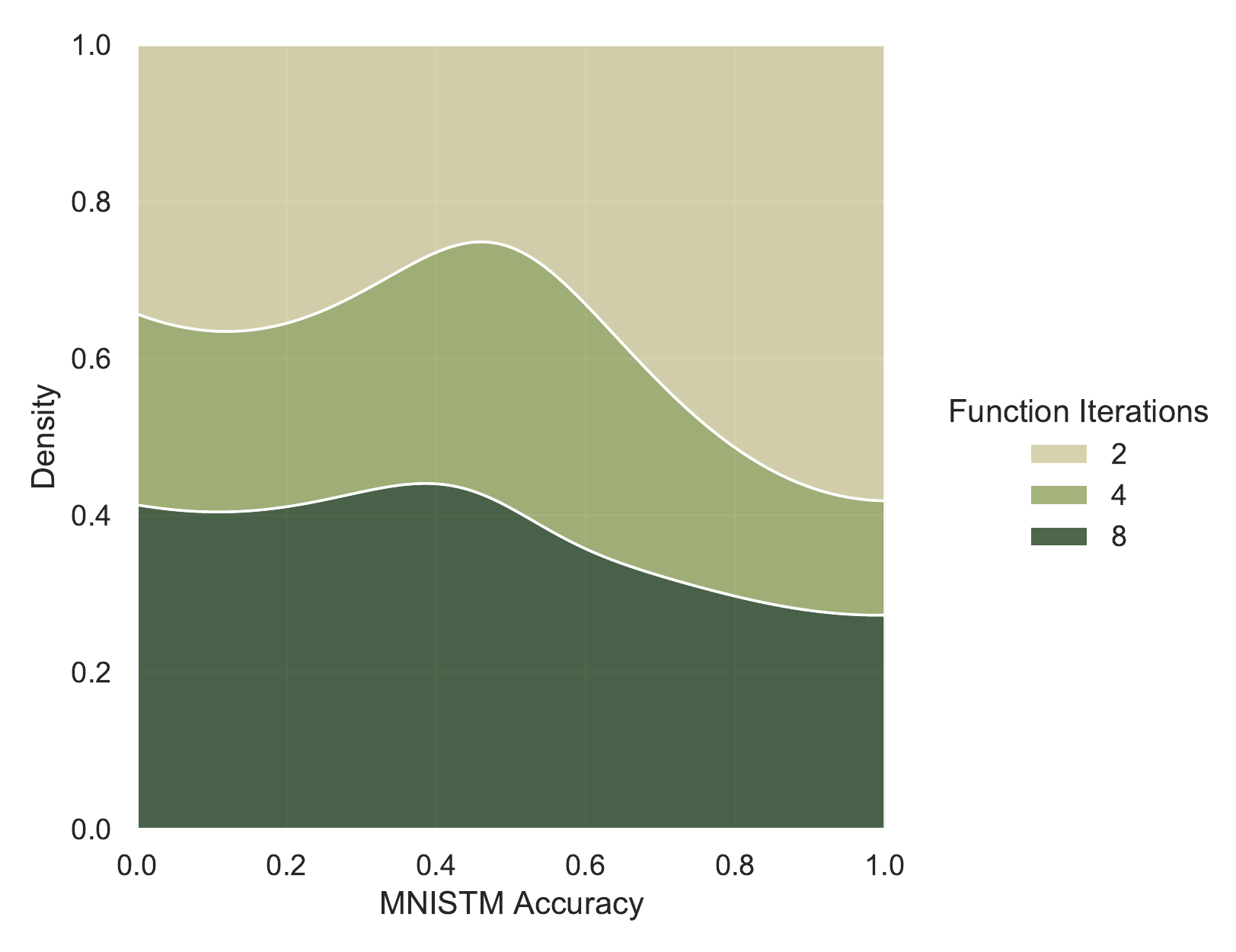}
    \caption[]%
    {MNIST-M}
  \end{subfigure}
   \hfill
  \begin{subfigure}[b]{0.33\textwidth}
    \centering
    \includegraphics[width=\textwidth]{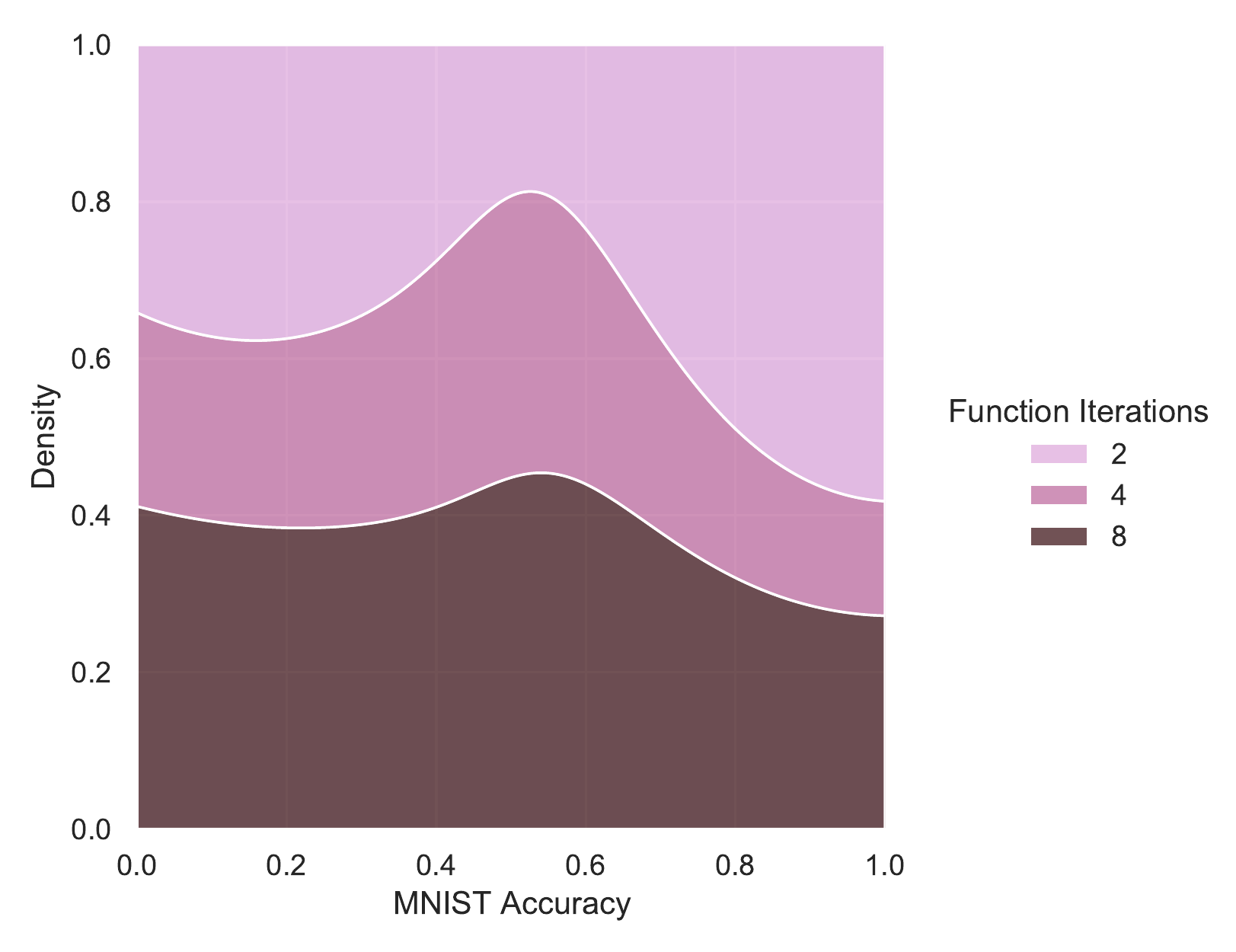}
    \caption[]%
    {MNIST}
  \end{subfigure}
  \caption[ Global caption ]
  {Conditional Kernel Density Estimates of validation performance, conditioned on the number of function iterations $n_i$.}
  \label{fig:digits_ablation_ns}
\end{figure}
\begin{figure}[htp]
  \begin{subfigure}[b]{0.33\textwidth}
    \centering
    \includegraphics[width=\textwidth]{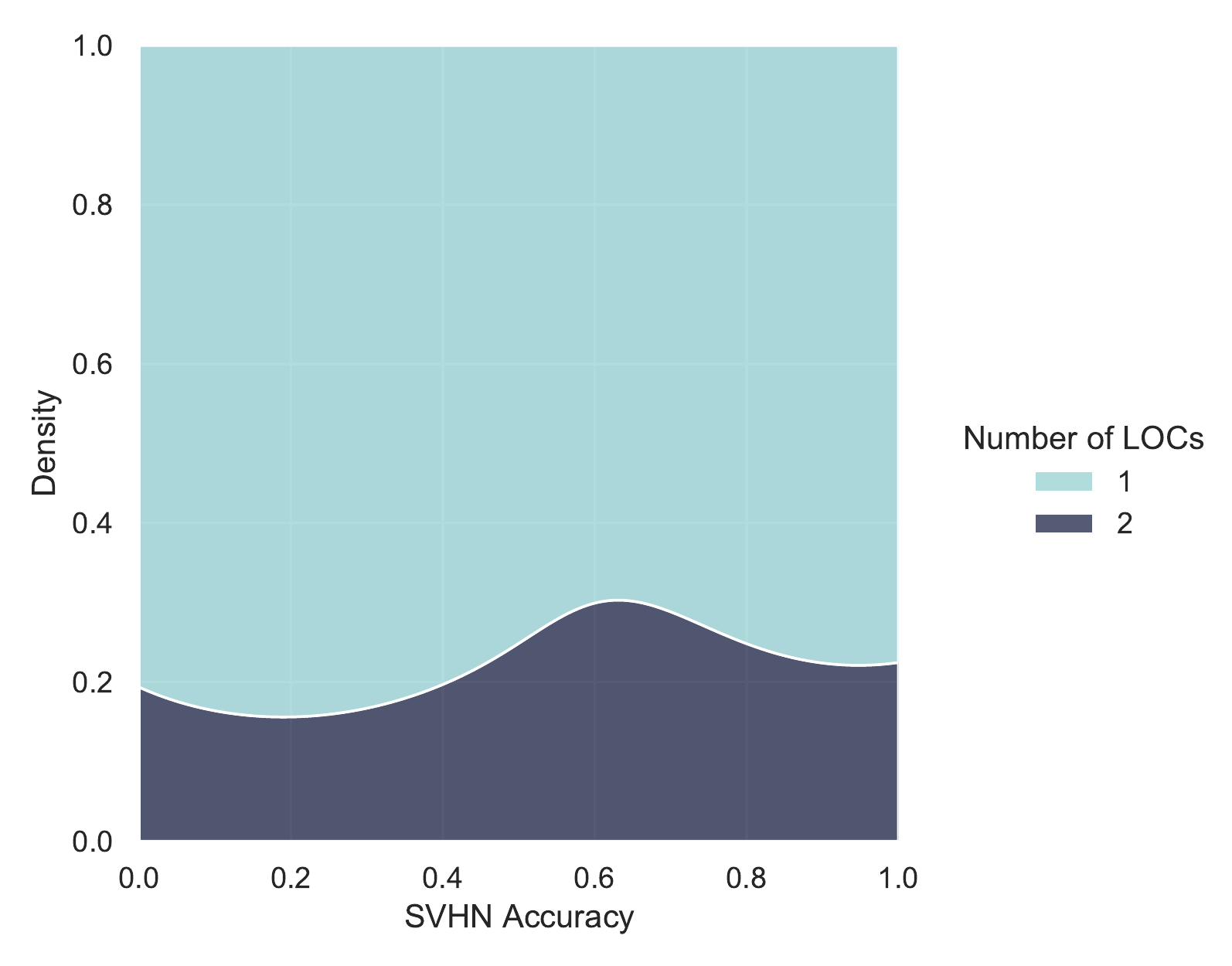}
    \caption[]%
    {SVHN}
  \end{subfigure}%
  \hfill
  \begin{subfigure}[b]{0.33\textwidth}
    \centering
    \includegraphics[width=\textwidth]{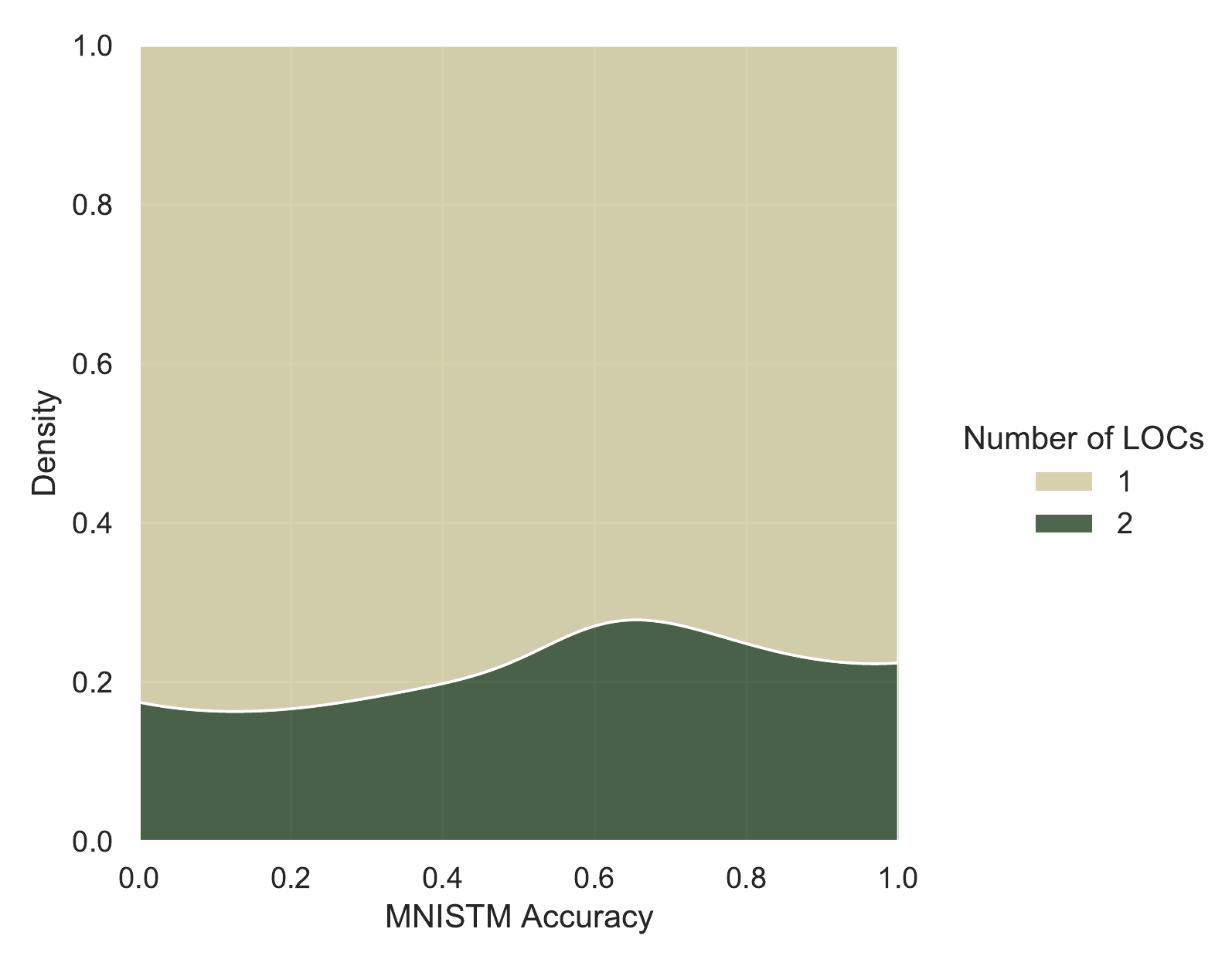}
    \caption[]%
    {MNIST-M}
  \end{subfigure}
   \hfill
  \begin{subfigure}[b]{0.33\textwidth}
    \centering
    \includegraphics[width=\textwidth]{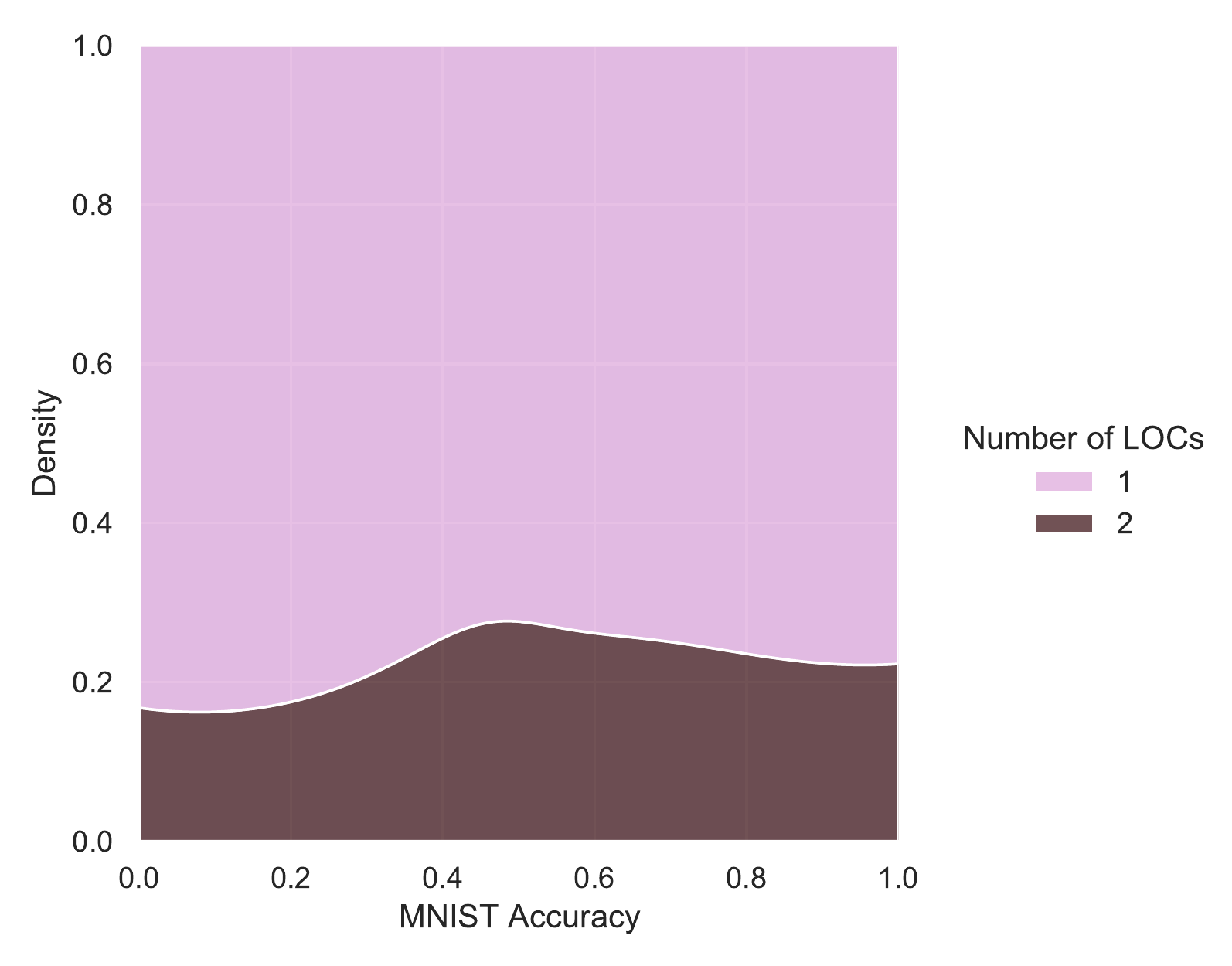}
    \caption[]%
    {MNIST}
  \end{subfigure}
  \caption[ Global caption ]
  {Conditional Kernel Density Estimates of validation performance, conditioned on the number of LOCs $n_l$.}
  \label{fig:digits_ablation_ns}
\end{figure}

\section{Abstract Reasoning with PGMs}\label{app:pgm}
Progressively Generated Matrices (PGMs) \citep{barrett2018measuring} have been used as a diagnostic dataset to study the compositional generalization capability of machine learning models \citep{wang2020abstract,steenbrugge2018improving}. The dataset consists of complex visual analogical reasoning tasks that require relational reasoning between attributes of different objects. The \lq object types' include shape and line comprising of the \lq attribute types': size, type, color, position, and number, where each attribute can take one of a finite number of discrete values. The \lq relationship types' consist of progression, XOR, OR, AND, and consistent union. The structure of a PGM task is governed by \emph{triples}, which are defined by applying a certain relationship type to the attributes of the objects. On an average, one to four relationships are used per task.

To study the various aspects of generalization in models, \citep{barrett2018measuring} introduced 8 different sub-datasets, corresponding to different generalization regimes of compositional reasoning. Except for the \textit{Neutral} regime, the test dataset in each regime measures the out-of-distribution generalization i.e., the test and the training datasets are different in a clearly defined manner. We use 6 of such regimes in this work, namely: Interpolation, Extrapolation, Held-out (H.O.) triples, H.O. pairs of triples, H.O. Attribute Pairs, and Neutral. The details on these regimes are provided below.

\subsubsection{Generalization Regimes}

\textbf{Neutral:} The neutral regime measures the in-distribution generalization, i.e., the training and the test sets consist of any triples.

\textbf{Interpolation:} In the training dataset of interpolation regime, the discrete values of the attributes are restricted to even numbers whereas the test set consists of odd-valued attributes.

\textbf{Extrapolation:} For the training dataset of extrapolation regime, the attribute values were restricted to the lower half of the discrete set whereas the test set consists of values sampled from the upper half of the discrete set.

\textbf{Held-out Triples:} The PGM dataset contains 29 unique triples. In the test set of held-out triples, 7 of such triples were held-out, while the rest of the triples are used to create the training dataset.

\textbf{Held-out Pairs of Triples:} All tasks contain at least two triples, leading to 400 viable triple pairs \citep{barrett2018measuring}. In \textit{Held-out Pairs of Triples}, 360 such pairs are randomly allocated to the training dataset and rest to the test dataset.

\textbf{Held-out Attribute Pairs:} Here, each task consists of at least two triples, where there are 20 viable pairs of attributes. Of these 20 pairs, 16 have been used to create the training set while the remaining 4 are used in the test set.

\subsection{Details of PGM Experiments}\label{sec:pgm-experiments-details}
For each PGM sub-dataset, we train multiple models for both Vision Transformers and Neural Interpreters. Each model is trained for 30 epochs and the model selection is done by evaluating its performance on validation datasets. The reported test accuracy in Table ~\ref{table:pgm-main-table} corresponds to the best validation performance. We perform hyper-parameter sweeps to find the best configuration of Neural Interpreters in each regime. 

\subsubsection{Hyperparameter Settings}
We perform random sweeps to find the optimal hyperparameters for each PGM regime. Due to the huge computational overload and the massive size of the datasets, the number of experiments in each sweep is limited to 35. We carried forward the knowledge that we learned from the digits experiments (Section \ref{sec:digits}), and perturbed only those hyperparameters that had significant influence on the model's performance. Apart from changing these selected hyperparameters, the models are identical in all aspects. Table ~\ref{table:pgm_hparams_constants} provides the hyperparameters that are kept the same in all the models, whereas Table ~\ref{table:pgm_hparams_ranges} shows the hyperparameters that we perturb and the ranges from which their values are randomly sampled.

\begin{table}
\centering
\caption{Hyperparameters and their values that are kept the same in all the PGM experiments.} \label{table:pgm_hparams_constants}
\begin{tabular}{l  l}
  \toprule
  \textbf{Parameters} & \textbf{Values}\\
  \midrule 
  Batch size & $72$\\
  Epochs & 30 \\
  \midrule
  
  Dimension of code vector ($\mathbf{c}$) & 192 \\
  Dimension of intermediate features & 192 \\
  Number of scripts ($n_s$) & 2 \\
  Number of function iterations ($n_i$) & 8 \\
  Number of LOCs ($n_l$) & 1 \\
  Number of functions ($n_f$) & 5 \\
  Number of heads per LOC & 4 \\
  Number of features per LOC head & 32 \\
  Type Inference MLP Depth & 2 \\
  Type Inference MLP Width & 192 \\
  Variable Features dimensions & 192 \\
  Frozen Function Codes & \texttt{False} \\
  \midrule
  Optimizer & RAdam \\
  Adam: beta1 & 0.9\\
  Adam: beta2 & 0.999\\
  Adam: epsilon & 1e-8\\
  Adam: learning rate & 0.0004 \\
  Learning Rate Scheduler & Cosine \\ 
  Cosine Scheduler Eta Max & 0.0004 \\
  Cosine Scheduler Eta Min & 0.0001 \\
  \midrule
  Number of parameters & 1.6M \\
  \bottomrule
\end{tabular}
\end{table}%

\begin{table} 
\centering
\caption{Hyperparameters whose values are randomly sampled from the given ranges for each experiment.} \label{table:pgm_hparams_ranges}
    \begin{tabular}{l  l}
      \toprule
      \textbf{Parameters} & \textbf{Ranges}\\
      \midrule 
      Kernel truncation parameter ($\tau$) & $[1.3, 1.7]$\\
      Type features dimensions & $[20, 24, 28, 32, 36, 40]$ \\
      Detach Function Signatures & [\texttt{True}, \texttt{False}] \\
      Number of scripts ($n_s$) & $[1, 2, 4]$ \\
      Function Iterations & $[4, 8, 16]$ \\
      \bottomrule
    \end{tabular}
\end{table}
For the sake of consistency, we make sure that the number of computational steps remain the same in all the experiments. For PGM sweeps, the number of computational steps are set to be 16. We varied the numbers of scripts $n_s$ and function iterations $n_i$ such that their product comes out to be 16.

\subsubsection{Optimal Neural Interpreters Configuration for PGMs} After running the set of experiments, we found that the one configuration that outperformed all other configurations was with 2 scripts i.e., $n_s = 2$ and 8 function iterations. There are small fluctuations in the selection of kernel truncation parameter $\tau$ and dimensions of type space $d_{\text{type}}$ that we detail below in Table ~\ref{table:pgm-selected-hparams}. 

\begin{table*}[h!]
\centering
\vspace{-5pt}
\caption{Hyperparameters of Neural Interpreters for the considered PGM datasets.} 
\label{table:pgm-selected-hparams}
\begin{tabular}{l@{\hspace{0.15cm}}|@{\hspace{0.15cm}}c@{\hspace{0.3cm}}|@{\hspace{0.15cm}}c@{\hspace{0.3cm}}|@{\hspace{0.15cm}}c@{\hspace{0.3cm}}|@{\hspace{0.15cm}}c@{\hspace{0.3cm}}|@{\hspace{0.15cm}}c@{\hspace{0.3cm}}|@{\hspace{0.15cm}}c@{\hspace{0.3cm}}}
\toprule
\textbf{Regime} & Neutral & Interpolation  &Attribute P. & Triple P. & Triples & Extra.\\
\midrule
  Number of scripts ($n_s$) & 2 & 2& 2& 2& 2& 2 \\
  Function iterations ($n_i$) & 8 & 8 & 8 & 8 & 8 & 8 \\
  Kernel truncation parameter ($\tau$) & 1.62 & 1.62 & 1.40 & 1.66 & 1.42 & 1.42 \\
  Type space dimensions ($d_{\text{type}}$) & 20 & 20 & 32 & 24 & 24 & 24 \\
  Frozen function signatures & False & False & True & False & True & True \\
\bottomrule
\end{tabular}
\end{table*}

\section{Diversity in Routing Mechanism}
\begin{figure}[htp]
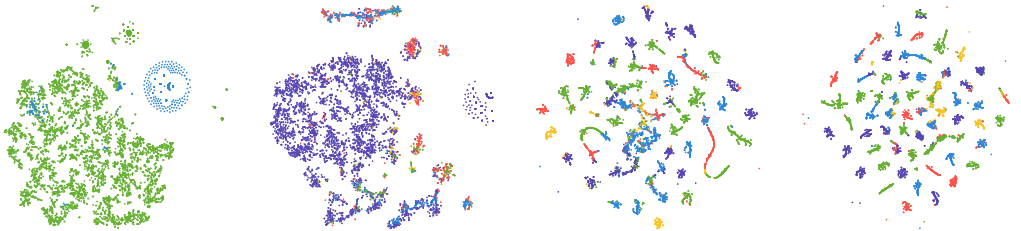

\begin{subfigure}[t]{0.24\textwidth} 
\centering
\includegraphics[width=1\textwidth]{figures/digits/type_tsne_s0_fp0_0b_b11.pdf}
\end{subfigure}
\hfill
\begin{subfigure}[t]{0.24\textwidth}
\centering
\includegraphics[width=1\textwidth]{figures/digits/type_tsne_s0_fp1_0b_b11.pdf}
\end{subfigure} 
\hfill
\begin{subfigure}[t]{0.24\textwidth}
\centering
\includegraphics[width=1\textwidth]{figures/digits/type_tsne_s1_fp0_0b_b11.pdf}
\end{subfigure} 
\hfill
\begin{subfigure}[t]{0.24\textwidth}
\centering
\includegraphics[width=1\textwidth]{figures/digits/type_tsne_s1_fp1_0b_b11.pdf}
\end{subfigure} 
\caption{
\small t-SNE embeddings of the inferred types of set elements as they progress through a Neural Interpreter with two scripts with two function iterations each. The color identifies the closest function in type space, and the progression from left to right is over the function iterations. \textbf{Gist:} Types are more clustered in the later function iterations, suggesting that the input set elements gradually \textit{develop} a type as they progress through the network.} \label{fig:digits_tsne_v2}
\end{figure}

\begin{figure}[htp]
\begin{subfigure}[t]{0.24\textwidth} 
\centering
\includegraphics[width=1\textwidth]{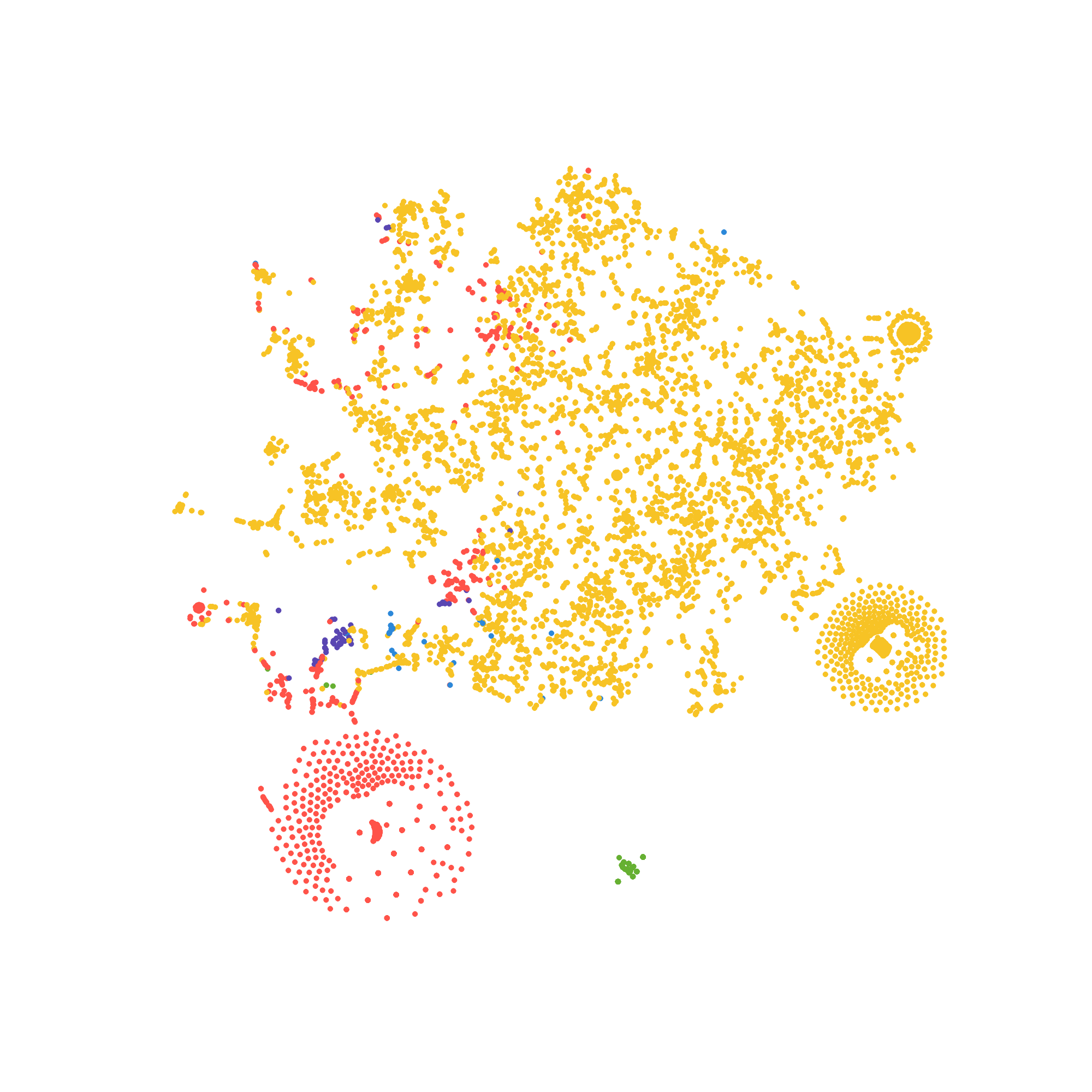}
\end{subfigure}
\hfill
\begin{subfigure}[t]{0.24\textwidth}
\centering
\includegraphics[width=1\textwidth]{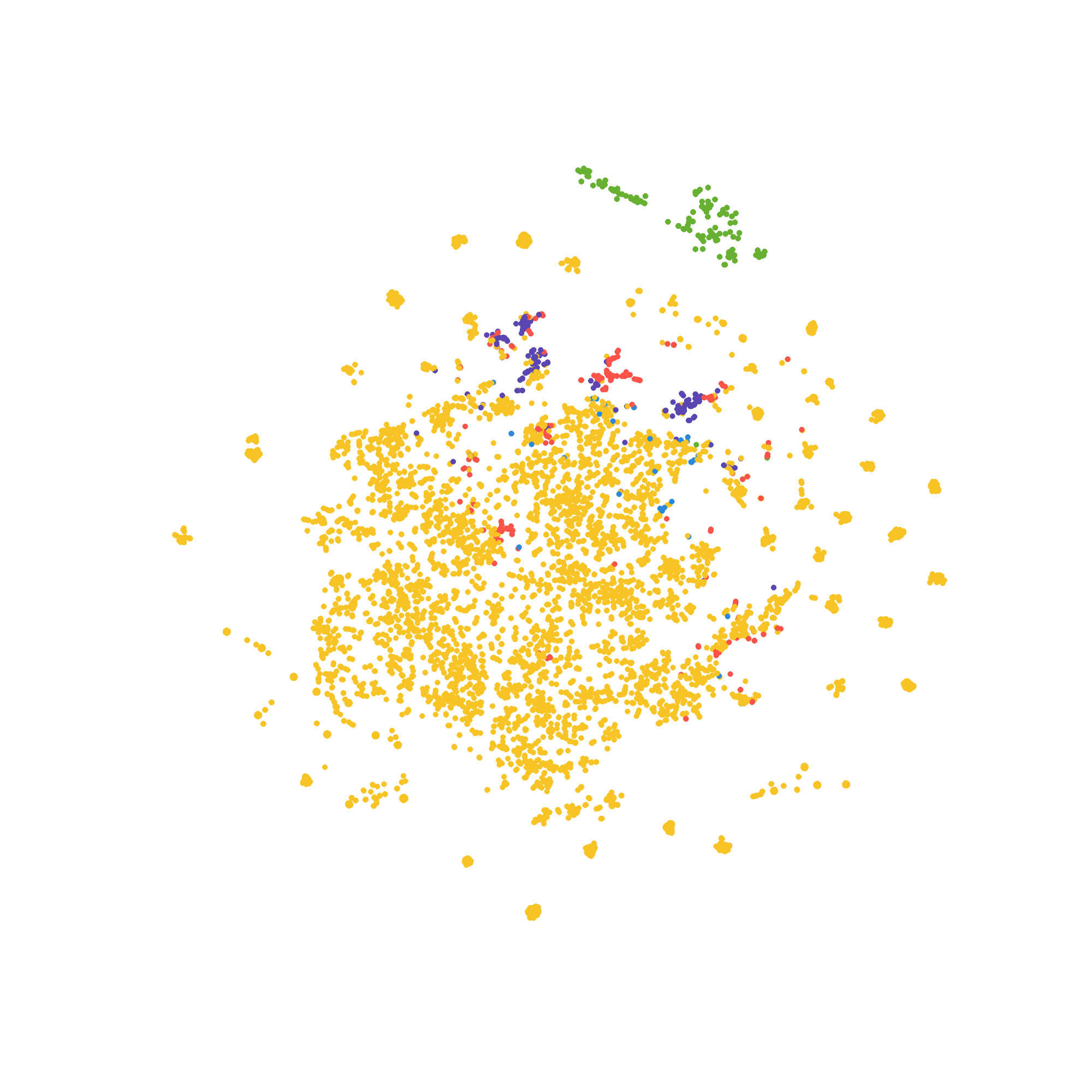}
\end{subfigure} 
\hfill
\begin{subfigure}[t]{0.24\textwidth}
\centering
\includegraphics[width=1\textwidth]{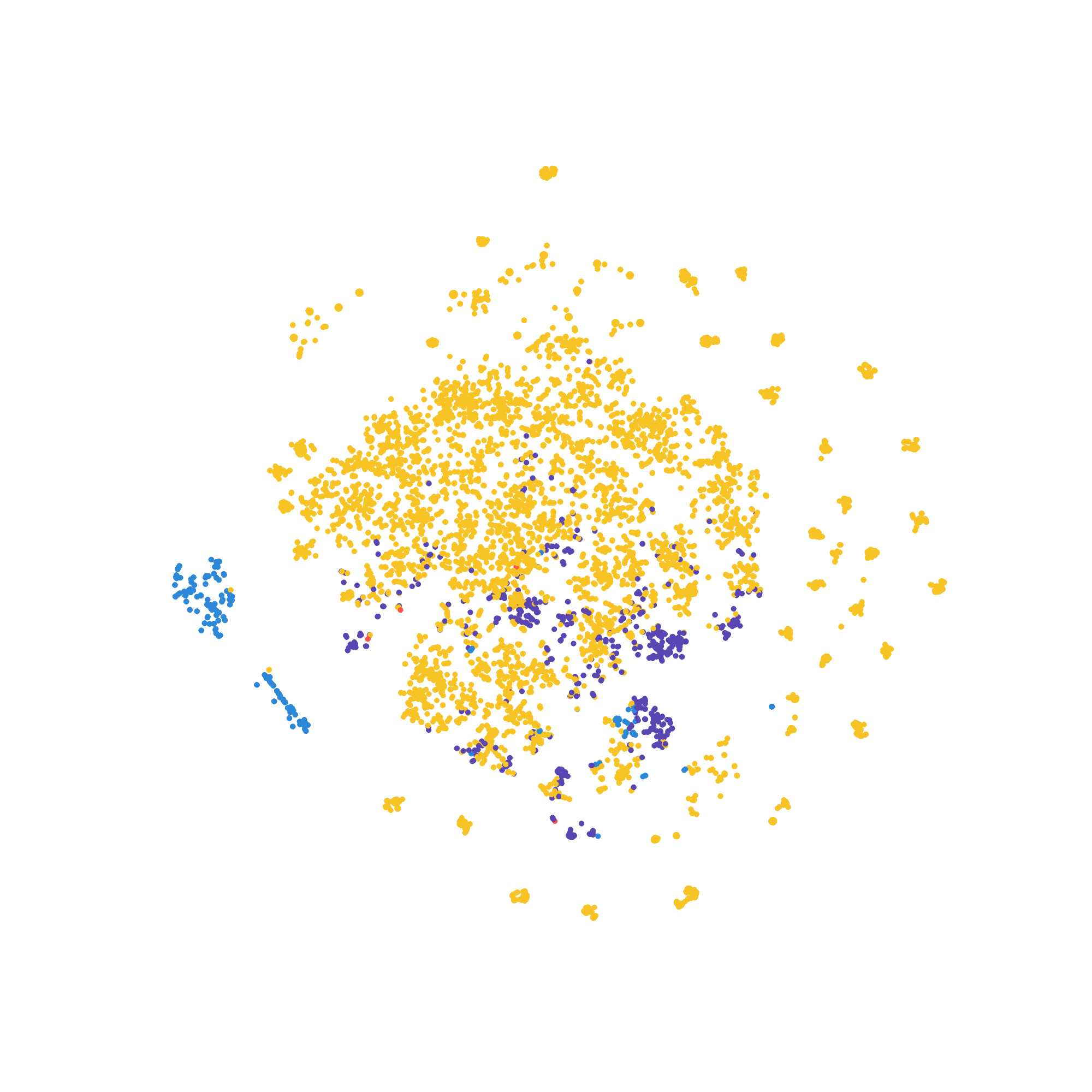}
\end{subfigure} 
\hfill
\begin{subfigure}[t]{0.24\textwidth}
\centering
\includegraphics[width=1\textwidth]{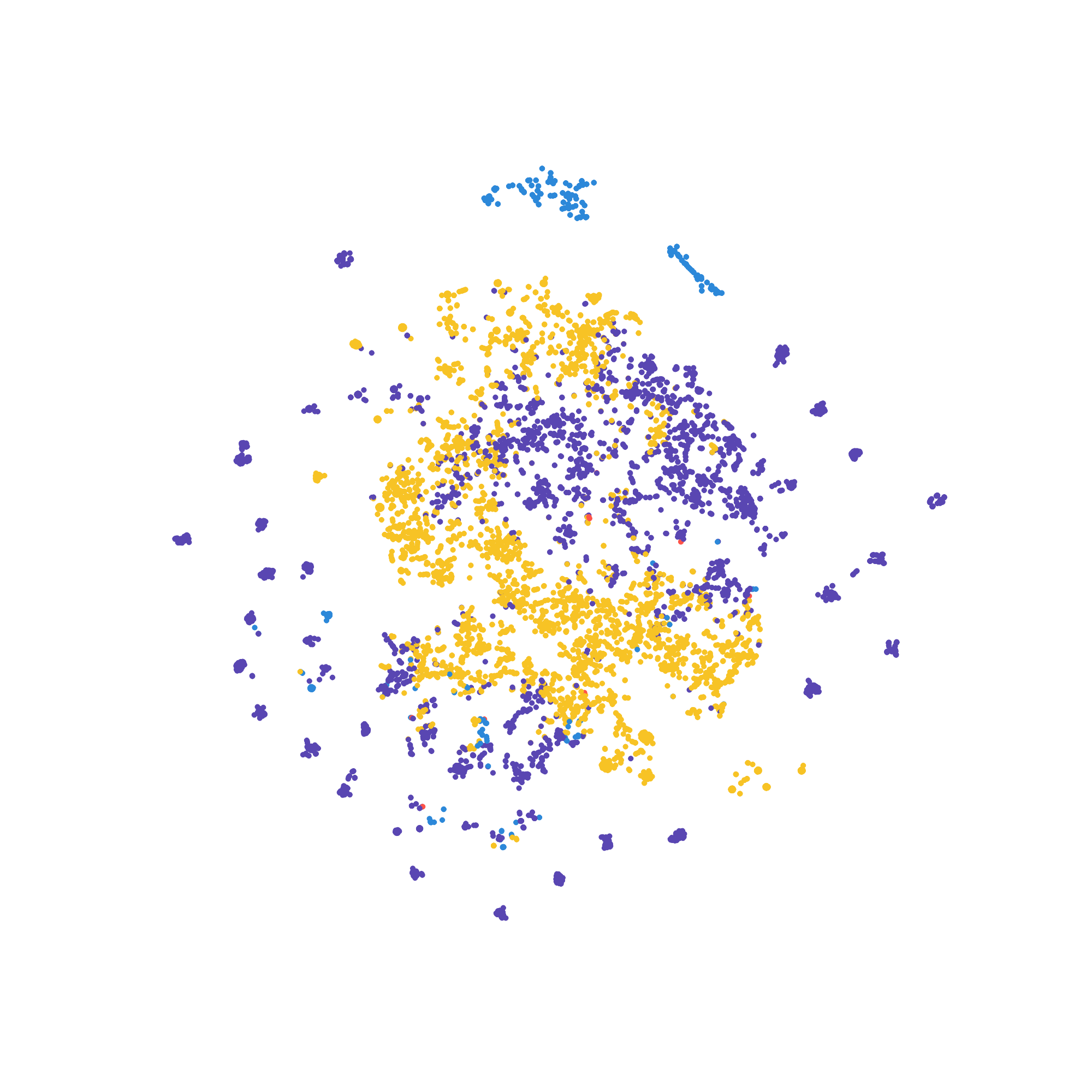}
\end{subfigure} 
\caption{
\small Same plot as above, but now with routing fixed at initialization. \textbf{Gist:} Inferred types of set elements exhibit less structure and diversity at initialization, especially at later function iterations. This suggests that the learning process indeed induces non-trivial patterns in how information is routed through the network.} \label{fig:digits_tsne_v3}
\end{figure}

We further investigate whether the learned routing in neural interpreters is meaningfully diverse i.e. whether certain samples get routed through certain functions? To answer it, we visualize the t-SNE embeddings of the variable types in Figure~\ref{fig:digits_tsne_v2}. The color-codes represent the close affinities between variables and certain functions in type space. We compare it against the case where the routing is fixed at initialization Figure~\ref{fig:digits_tsne_v3}. It can be seen that in the randomly initialized routing the type-function assignments (given by the colors assigned to a dot) exhibit less structure and diversity, especially at the later function iterations. This suggests that the learning process in neural interpreters indeed induces non-trivial patterns in how information is routed between modules.

\end{document}